\documentclass[lettersize,journal]{IEEEtran}
\usepackage{amsmath,amsfonts}
\usepackage{algorithmic}
\usepackage{algorithm}
\usepackage{array}
\usepackage[caption=false]{subfig}
\usepackage{textcomp}
\usepackage{stfloats}
\usepackage{url}
\usepackage{verbatim}
\usepackage{graphicx}
\usepackage{cite}
\begin{document}

\title{GlossyGS: Inverse Rendering of Glossy Objects with 3D Gaussian Splatting}

\author{Shuichang Lai, Letian Huang, Jie Guo, Kai Cheng, Bowen Pan, Xiaoxiao Long, Jiangjing Lyu, Chengfei Lv, Yanwen Guo
\thanks{Shuichang Lai and Letian Huang contributed equally.

Shuichang Lai, Bowen Pan, Jiangjing Lyu and Chengfei Lv are with Alibaba Group (e-mail: laishuichang.lsc@alibaba-inc.com; bowen.pbw@alibaba-inc.com; jiangjing.ljj@alibaba-inc.com; chengfei.lcf@alibaba-inc.com).

Letian Huang, Jie Guo and Yanwen Guo are with the National Key Lab for Novel Software Technology, Nanjing University,
Nanjing 210000, China (e-mail: 
lthuang@smail.nju.edu.cn; guojie@nju.edu.cn; ywguo@nju.edu.cn).

Kai Cheng is with the school of artificial intelligence and data science, University of Science and Technology of China, China (e-mail: chengkai21@mail.ustc.edu.cn).

Xiaoxiao Long is with the Department of Computer Science, The University of Hong Kong, Hong Kong (e-mail: xxlong@connect.hku.hk).

Jie Guo is the corresponding author.
}}

\markboth{SUBMITTED TO IEEE TVCG }%
{Shell \MakeLowercase{\textit{et al.}}: A Sample Article Using IEEEtran.cls for IEEE Journals}


\maketitle

\begin{abstract}
Reconstructing objects from posed images is a crucial and complex task in computer graphics and computer vision. While NeRF-based neural reconstruction methods have exhibited impressive reconstruction ability, they tend to be time-comsuming. Recent strategies have adopted 3D Gaussian Splatting (3D-GS) for inverse rendering, which have led to quick and effective outcomes. However, these techniques generally have difficulty in producing believable geometries and materials for glossy objects, a challenge that stems from the inherent ambiguities of inverse rendering. To address this, we introduce GlossyGS, an innovative 3D-GS-based inverse rendering framework that aims to precisely reconstruct the geometry and materials of glossy objects by integrating material priors. The key idea is the use of micro-facet geometry segmentation prior, which helps to reduce the intrinsic ambiguities and improve the decomposition of geometries and materials. Additionally, we introduce a normal map prefiltering strategy to more accurately simulate the normal distribution of reflective surfaces. These strategies are integrated into a hybrid geometry and material representation that employs both explicit and implicit methods to depict glossy objects. We demonstrate through quantitative analysis and qualitative visualization that the proposed method is effective to reconstruct high-fidelity geometries and materials of glossy objects, and performs favorably against state-of-the-arts. 
\end{abstract}


\begin{figure*}[h]
    \centering
    \includegraphics[width=\textwidth]{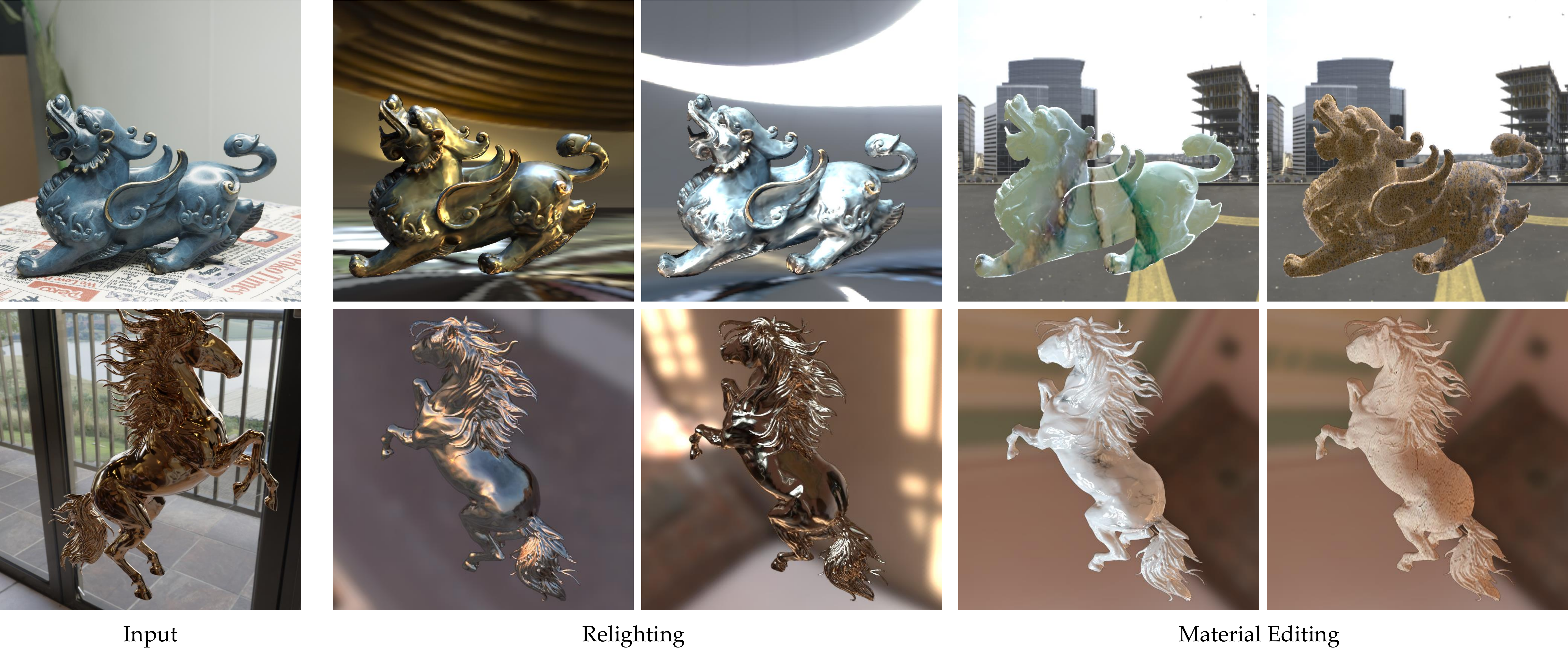}
    \caption{We present an innovative 3D-GS-based inverse rendering framework that aims to precisely reconstruct the geometries and materials of glossy objects by integrating priors from glossy appearance analysis. The well disentangled geometric and material properties allow us to achieve high-quality relighting (the second and third columns) and material editing  (the fourth and fifth columns), especially on highly glossy objects. }
\end{figure*}

\section{introduction}
\IEEEPARstart{R}{econstructing}
 objects with realistic materials using images from multiple viewpoints is a crucial and complex task in computer graphics and computer vision. This challenge is exacerbated when dealing with images taken in uncontrolled settings where the lighting conditions are unknown. Several methods~\cite{zhang2021physg, liu2023nero, karis2013real, jin2023tensoir, srinivasan2021nerv, zeng2023relighting, mai2023neural} build on Neural Radiance Fields (NeRFs)~\cite{mildenhall2021nerf}, using neural networks to depict objects' materials and geometry. However, the required dense sampling for rays undermines the efficiency of these inverse rendering approaches. Recently, 3D Gaussian Splatting (3D-GS)~\cite{kerbl20233d} has emerged as an effective strategy, offering high quality and real-time performance.

Recent studies~\cite{liang2023gs,gao2023relightable,jiang2023gaussianshader} try to build inverse rendering pipelines based on 3D-GS. However, these methods mostly perform well on non-specular objects. Till now, several challenges remain: specular highlights and environmental reflections introduce ambiguities, complicating the differentiation between inherent textures and reflections, thus leading to low-quality geometry reconstruction. Furthermore, microscopic variations in the normals of glossy surfaces can change how they look from distance. This is because tiny folds or protrusions can make microscopic smoothness appear rough. This nonlinear relationship between microscopic normal changes and appearance challenges 3D Gaussian-based forward shading methods, leading to inaccuracies in surface normal estimation.

In this paper, we introduce \textit{GlossyGS}, a novel 3D-GS-based inverse rendering framework designed to accurately identify the physical properties of glossy objects by incorporating material prior. The key idea of \textit{GlossyGS} consists of three aspects. Firstly, we introduce a \textit{normal map prefiltering} strategy that modifies the shading order by blending the normals first. This approach improves the accuracy of simulating the normal distribution of glossy surfaces. Secondly, we propose the \textit{micro-facet geometry segmentation prior} to adjust the micro-facet normal distribution and the geometry of reflective surfaces. This prior significantly reduces the inherent ambiguities in inverse rendering and enhances the decomposition of both geometry and material properties. A micro-facet geometry segmentation model is developed to predict this prior using semantic representations derived from large vision models. Finally, we design a \textit{hybrid geometry and material representation} that utilizes anchors to explicitly represent the macroscopic coarse geometry while implicitly employing feature vectors associated with these anchors to predict multiple neural Gaussians. These neural Gaussians are essential for precisely simulating the normal distribution and material properties of the micro-facet, offering a detailed and realistic representation of glossy objects' surfaces. Extensive experimental results show that our approach can reconstruct high-fidelity geometries and materials of glossy objects, leading to state-of-the-art rendering for relighting. To summarize, our main contributions are three-fold. 
\begin{itemize}
    \item We propose the micro-facet geometry segmentation prior to minimize the inherent ambiguities in inverse rendering, consequently enhancing the decomposition of both geometry and material properties.
    \item We introduce a normal map prefiltering strategy that changes the shading order, enhancing the simulation of the normal distribution for glossy surfaces.
    \item We design a hybrid geometry and material representation tailored for glossy objects, achieving state-of-the-art inverse rendering results.
\end{itemize}



\section{Related Work}
\subsection{3D Gaussian Splatting}

3D Gaussian Splatting (3D-GS)~\cite{kerbl20233d} has achieved high-fidelity real-time neural rendering by splatting the Gaussian  primitives onto the image plane for a fast tile-based rasterization, instead of densely sampling points on the rays like Neural Radiance Fields (NeRF)~\cite{mildenhall2021nerf, barron2021mip} for a slow volume rendering. Due to its real-time performance, some works attempt to integrate the efficient explicit representation into human modeling~\cite{hu2023gauhuman, hu2023gaussianavatar,qian2023gaussianavatars,moreau2023human,kocabas2023hugs,qian20233dgs}, dynamic scenes~\cite{lu20243d, wu20234d, zhou2023drivinggaussian}, large-scale scenes~\cite{lin2024vastgaussian, liu2024citygaussian} and slam~\cite{matsuki2023gaussian, yan2023gs,li2024sgs}, etc. Gaussian Editors~\cite{chen2023gaussianeditor, fang2023gaussianeditor} facilitate editing of the appearance and geometry of 3D-GS through the diffusion model~\cite{diffusion1, diffusion2, diffusion3}. SuGaR~\cite{guedon2023sugar} and 2D-GS~\cite{huang20242d}, analyze the surface representation capability of Gaussians and propose a method for representing Gaussians' normals, facilitating the extraction from Gaussians into traditional meshes. Our method also employs a high-fidelity normal map prefiltering strategy to facilitate the inverse rendering for glossy objects,  which enables photo-realistic relighting and material editing.

Furthermore, there are some works focusing on addressing the current limitations of 3D-GS. Yu et al.~\cite{yu2023mip} utilizes a 3D smoothing filter and a 2D mip filter to reduce aliasing when zooming in and zooming out. Huang et al.~\cite{huang2024error} proposes an optimal projection strategy by minimizing projection errors, resulting in photo-realistic rendering under various focal lengths and various camera models. Radl et al.~\cite{radl2024stopthepop} and Bulò et al.~\cite{bulo2024revising} conduct analysis and improvements based on 3D Gaussians' depth sorting and density control respectively. Scaffold-GS~\cite{lu2023scaffold} employs a hierarchical hybrid representation of scenes through the combination of explicit 3D Gaussians and implicit MLP layers, significantly reducing the storage overhead of 3D-GS. 
Our method leverages the hybrid structure to infer neural Gaussians and neural materials and introduce the similarity of material within latent features to constrain material ambiguities. These designs facilitate the success of inverse rendering of glossy objects.

\subsection{Inverse Rendering of Glossy Objects}

Reconstruction and inverse rendering of a glossy object from mult-view images has been a challenging task due to the complex light interactions and ambiguities~\cite{roth2006specular, han2016mirror, zeng2023mirror,dave2022pandora, kadambi2015polarized, whelan2018reconstructing, aittala2016reflectance, barron2014shape, nimier2019mitsuba, chang2023parameter}. 

NeRO~\cite{liu2023nero} use Integrated
Directional Encoding (IDE) like Ref-NeRF~\cite{verbin2022ref} and employs a two-stage inverse rendering strategy. The first stage uses split-sum approximation~\cite{karis2013real} to accelerate integral calculations, while the second stage employs Monte Carlo sampling to refine materials. 
TensoIR~\cite{jin2023tensoir} employs TensoRF~\cite{chen2022tensorf} as its geometry representation and utilizes ray marching to compute indirect illumination and visibility.
NMF~\cite{mai2023neural} uses a microfacet reflectance model within a volumetric setting by treating each sample along the ray as a (potentially non-opaque) surface.
However, these methods that rely on implicit representations~\cite{zhang2021physg, verbin2022ref, liu2023nero, karis2013real, jin2023tensoir, srinivasan2021nerv, wu2023nerf, ma2023specnerf, zeng2023relighting} involve densely sampled points along rays, leading to a significant decrease in the performance of inverse rendering.  

Some works~\cite{jiang2023gaussianshader, liang2023gs} aim to employ 3D-GS~\cite{kerbl20233d} as an explicit representation of scenes in order to enhance the performance of inverse rendering. GaussianShader~\cite{jiang2023gaussianshader} decomposes the original color attributes into three components including diffuse, specular and residual color for modeling the color of reflective surfaces. 
In GS-IR~\cite{liang2023gs}, 
a technique similar to light probes~\cite{debevec2008median} is employed to obtain the indirect light and ambient occlusion. However, these methods~\cite{jiang2023gaussianshader, liang2023gs} face challenges in representing accurate surfaces and materials using 3D Gaussians without correlation and suffer from the ambiguities between macroscopic normal and microscopic roughness, resulting in lower quality of inverse rendering. In contrast, our method effectively addresses these issues.


\section{Method}

\begin{figure*}[ht]
  \centering
    \includegraphics[width=1\linewidth]{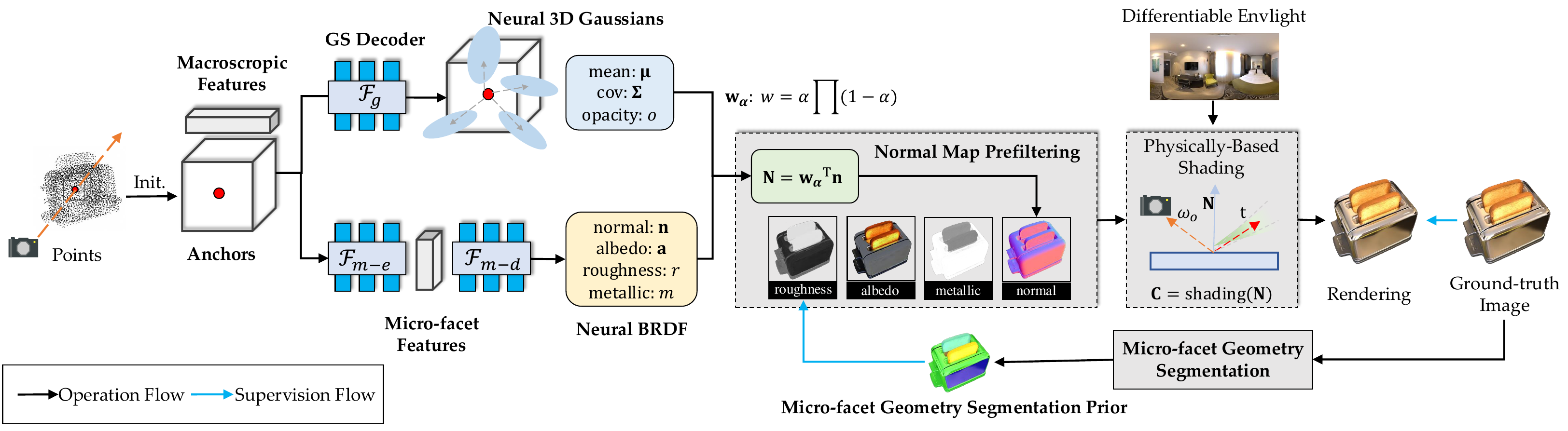}
    \caption{Illustration of the inverse rendering pipeline for our \textit{GlossyGS}. The initial points are used to generate anchors, which carry the macroscopic features. These features are fed to GS decoder and material encoder, generating neural 3D Gaussians and neural materials (BRDF). Normal map prefiltering is used as a prefilter for these neural Gaussians and materials to obtain the corresponding material maps. The final rendering is shading of these maps and differentiable environment lighting, supervised by ground-truth images and micro-facet geometry segmentation prior.}
    \label{fig:pipeline}
\end{figure*}

In this section, we present our approach, \textit{GlossyGS}, tailored for the inverse rendering of glossy objects, as illustrated in Fig.~\ref{fig:pipeline}. To achieve fast and high-quality reconstruction of glossy objects, we employ a hybrid explicit-implicit geometry and material representation to infer neural Gaussians and materials (BRDFs). Our experiments reveal that the shading order of the Gaussians negatively impacts the geometry reconstruction of glossy objects. To mitigate this issue, we propose a normal map prefiltering strategy, which allows us to first generate material maps, subsequently using them to obtain the final rendered images through shading. Another factor affecting the reconstruction of glossy objects is the ambiguities arising from the coupling of microscopic and macroscopic normal distributions. To address this problem, we introduce a micro-facet geometry segmentation prior. Furthermore, we develop a segmentation model that learns this prior based on our custom-built dataset.

\subsection{Preliminaries}

\textit{3D Gaussian Splatting}~\cite{kerbl20233d} 3D Gaussian Splatting takes the sparse point clouds generated during the SFM~\cite{schonberger2016structure} calibration as input. From these points, it constructs a set of 3D Gaussians as graphical primitives explicitly representing the scene. Each graphic primitive is characterized by Gaussian mean $\boldsymbol{\mu}$ and covariance matrix $\boldsymbol{\Sigma}$, as well as opacity $o$ to represent volumetric density, and is characterized by spherical harmonic ($\text{SH}$) coefficients to represent appearance.

The 3D primitives are projected to 2D screen space, and rasterized using $\alpha$-blending. The $\alpha$-blending weights are given as
\begin{equation}
    w_k = \alpha_k\prod_{j=1}^{k-1}\left(1-\alpha_j\right),
\end{equation} 
where $\alpha$ is given by $o \cdot \mathit{g}$. The projected Gaussian function $\mathit{g}$ at pixel $\left(u,v\right)$ is represented by:
\begin{equation}
\mathit{g} \left(u,v\right)=e^{\left(-\frac{1}{2} \left([u,v]^{\top}-\boldsymbol{\mu}^{'}\right)^{\top}{\boldsymbol{\Sigma}^{'}}^{-1}\left([u,v]^{\top}-\boldsymbol{\mu}^{'}\right)\right)}
\end{equation}
where $\boldsymbol{\mu}^{'}=\text{proj}\left(\boldsymbol{\mu}\right)$
and $\boldsymbol{\Sigma}^{'}=\text{proj}\left(\boldsymbol{\Sigma}\right)$
represent the projected 2D mean and covariance matrix, respectively. The pixel color at $\left(u,v\right)$ is calculated using $\alpha$-blending with spherical harmonics ($\text{SH}$) as follows:
\begin{equation}
    \mathit{C}\left(u,v\right)=\sum_{k=1}^{N}{\text{SH}_k\left(u,v\right)w_k},
\end{equation}
where $\text{SH}_k\left(u,v\right)$ represents the color of the $k^{th}$ primitive at pixel $\left(u,v\right)$.
Expanding this method to include attributes like normal and shading gives as:
\begin{equation}
\label{eq:alpha_blending}
\mathit{Y}\left(u,v\right)=\sum_{k=1}^{N}{y_k w_k},
\end{equation} 
a formula commonly used in inverse rendering tasks based on 3D-GS~\cite{gao2023relightable, jiang2023gaussianshader, liang2023gs}.  In this work, we utilize a modified 3D-GS approach for high-quality, real-time inverse rendering.

\textit{Cook-Torrance BRDF}~\cite{cook1982reflectance}
We follow the Cook-Torrance model \cite{cook1982reflectance} and formulate the bidirectional reflectance distribution function (BRDF) as follow:
\begin{equation}\label{eq:brdf1}
f(\omega _i,\omega _o) = \underbrace{(1-m)\frac{\mathbf{a}}{\pi }}_{\text{diffuse}} + \underbrace{\frac{\mathit{DFG}}{4(\omega _i\cdot \mathbf{n} )(\omega _o\cdot \mathbf{n} ) }}_{\text{specular}} ,
\end{equation} 
where $m\in \left [0,1  \right ]$ is the metallic of the point, $\mathbf{a}  \in \left [0,1  \right ]^3 $ is the albedo color of the point, $\mathbf{n} $ denotes the surface normal,  $\omega _i$ denotes the incident direction, $\omega _o$ denotes the outgoing view direction. $\mathit{D}$, $\mathit{F}$, and $\mathit{G}$ refer to the normal distribution function, the Fresnel term, and the geometry term respectively. $\mathit{D}$, $\mathit{F}$, and $\mathit{G}$ are generally determined by the metallic $m$, the roughness $r \in \left [0,1  \right ] $ and the albedo $\mathbf{a}$. To tackle the intractable radiance integral, we adopt image-based lighting (IBL) model and split-sum approximation \cite{karis2013real} following \cite{liu2023nero}. Here, the lighting integral is approximated as:
\begin{gather}
\label{eq:diffuse_color}
    c(\omega _o)=c_{\text{diffuse}}+c_{\text{specular}}(\omega _o), \\
   c_{\text{diffuse}}=\mathbf{a}(1-m) \cdot \underbrace{{\int_{\Omega }\mathit{L}(\omega _i)\frac{\omega _i\cdot\mathbf{n} }{\pi } d\omega _i}}_{\mathit{L}_{d}}, 
\end{gather}
\begin{equation}
\label{eq:specular_color}
\begin{split}
    c_{\text{specular}}(\omega _o)&=\int_{\Omega }\mathit{L}(\omega _i)\frac{\mathit{DFG}}{4(\omega _o\cdot \mathbf{n} )}  d\omega _i\\
    &\approx \underbrace{\int_{\Omega }\frac{\mathit{DFG}}{4(\omega _o\cdot \mathbf{n} )}  d\omega _i}_{\mathit{M}_{s}} \cdot  \underbrace{\int_{\Omega }\mathit{D}(r, \mathbf{t}) \mathit{L}(\omega _i) d\omega _i}_{\mathit{L}_{s}}, 
\end{split} 
\end{equation} 
where $\mathit{L}_s$ denotes the integral of lights on the micro-facet normal distribution function $D(r, \mathbf{t})$, $\mathbf{t}$ is the reflective direction, $\mathit{L}_d$ denotes the diffuse light integral, $\mathit{M}_{s}$ denotes the integral of BRDF. 


\subsection{Hybrid Geometry and Material Representation}
Inspired by Scaffold-GS~\cite{lu2023scaffold}, we adopt a hybrid explicit-implicit geometry representation and employ neural networks to generate neural Gaussians to represent objects. Specifically, the sparse point cloud from COLMAP \cite{schonberger2016structure} is initialized as anchors, with each anchor equipped with a feature vector $\mathbf{f}_{\text{macro}}$, a scaling factor $s$, and learnable offsets. We then employ the GS decoder $\mathcal{F}_g$ to generate neural Gaussians that represent the geometry of objects:
\begin{equation}
    \boldsymbol{\mu}, \boldsymbol{\Sigma}, o = \mathcal{F}_g\left(\mathbf{f}_{\text{macro}},\mathbf{x}_{\text{anchor}}\right)
\end{equation} where $\mathbf{x}_{\text{anchor}}$ denotes positions of the anchors and $\boldsymbol{\mu}, \boldsymbol{\Sigma}, o$ denote attributes of 3D Gaussians. We further extend this hybrid representation to BRDF materials. Since the $\mathbf{f}_{\text{macro}}$ contains complete Gaussian geometry and material information, we use the material encoder $\mathcal{F}_{m-e}$ to generate high-dimensional micro-facet features
\begin{equation}
    \mathbf{f}_{\text{micro}} = \mathcal{F}_{m-e}\left(\mathbf{f}_{\text{macro}}\right), 
\end{equation}
which are then decoded to obtain material attributes
\begin{gather}
    r, m, \mathbf{n}, \mathbf{a} = \mathcal{F}_{m-d}\left(\mathbf{f}_{\text{micro}}\right)
\end{gather} where $r,m$ are the roughness and metallic of the Gaussian respectively. And $\mathbf{n},\mathbf{a}$ are the normal and albedo color of the Gaussian, respectively. 
This representation not only reduces the storage cost of materials but also enhances the correlation between materials in local space for they share a common MLP. For more details about MLP structures, please refer to the supplementary material.

Building on this design, we devise a smooth, low-dimensional manifold for material representation. This contiguous manifold ensures smooth variations in material attributes, thereby preventing the abrupt changes caused by incorrect gradients. This approach not only stabilizes the optimization process but also enhances the robustness of our network, ensuring a more reliable and consistent performance in material representation. In practice, we apply a smoothness regularization on the decoder $\mathcal{F}_{m-d}$,
\begin{equation}
    \mathcal{L}_{s} = \left \| \mathcal{F}_{m-d}(\mathbf{f}_{\text{micro}})- \mathcal{F}_{m-d}(\mathbf{f}_{\text{micro}}+\xi) \right \| _1
\end{equation}
where $\xi$ denotes a random variable sampled from a normal distribution with mean 0 and variance 0.01.



Additionally, since our geometric and material attributes are still discrete, our hybrid representation retains the high performance of the vanilla 3D-GS~\cite{kerbl20233d} by projecting them onto the image plane and performing $\alpha$-blending instead of dense sampling of points along rays.

\subsection{Normal Map Prefiltering}

\begin{figure}[ht]
  \centering
  
  \subfloat[Shading on 3D Gaussian]{
    \includegraphics[width=0.49\linewidth]{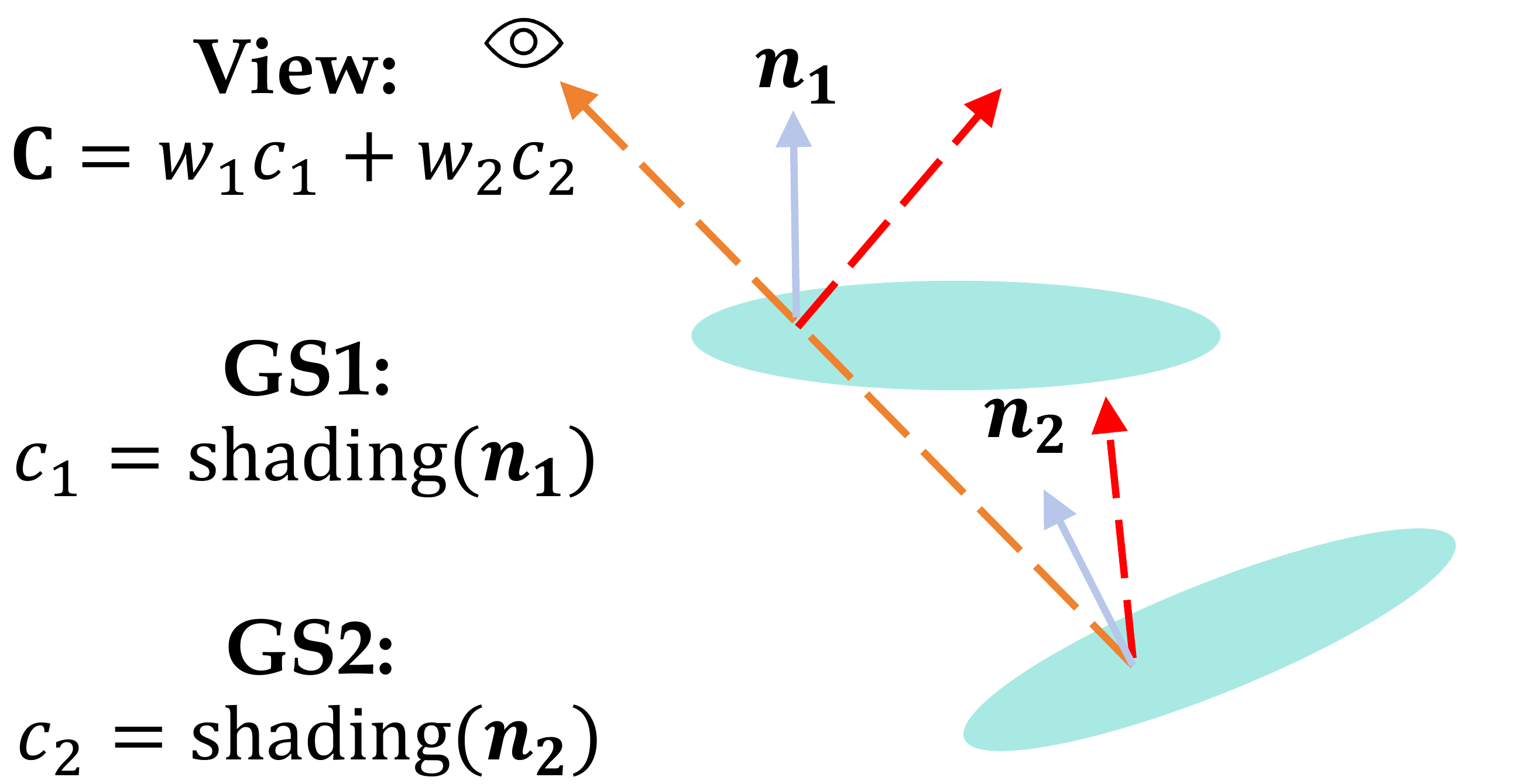}
    \label{fig:motivation_prefilter_gs}}    
  \subfloat[Shading on Surface]{
    \includegraphics[width=0.49\linewidth]{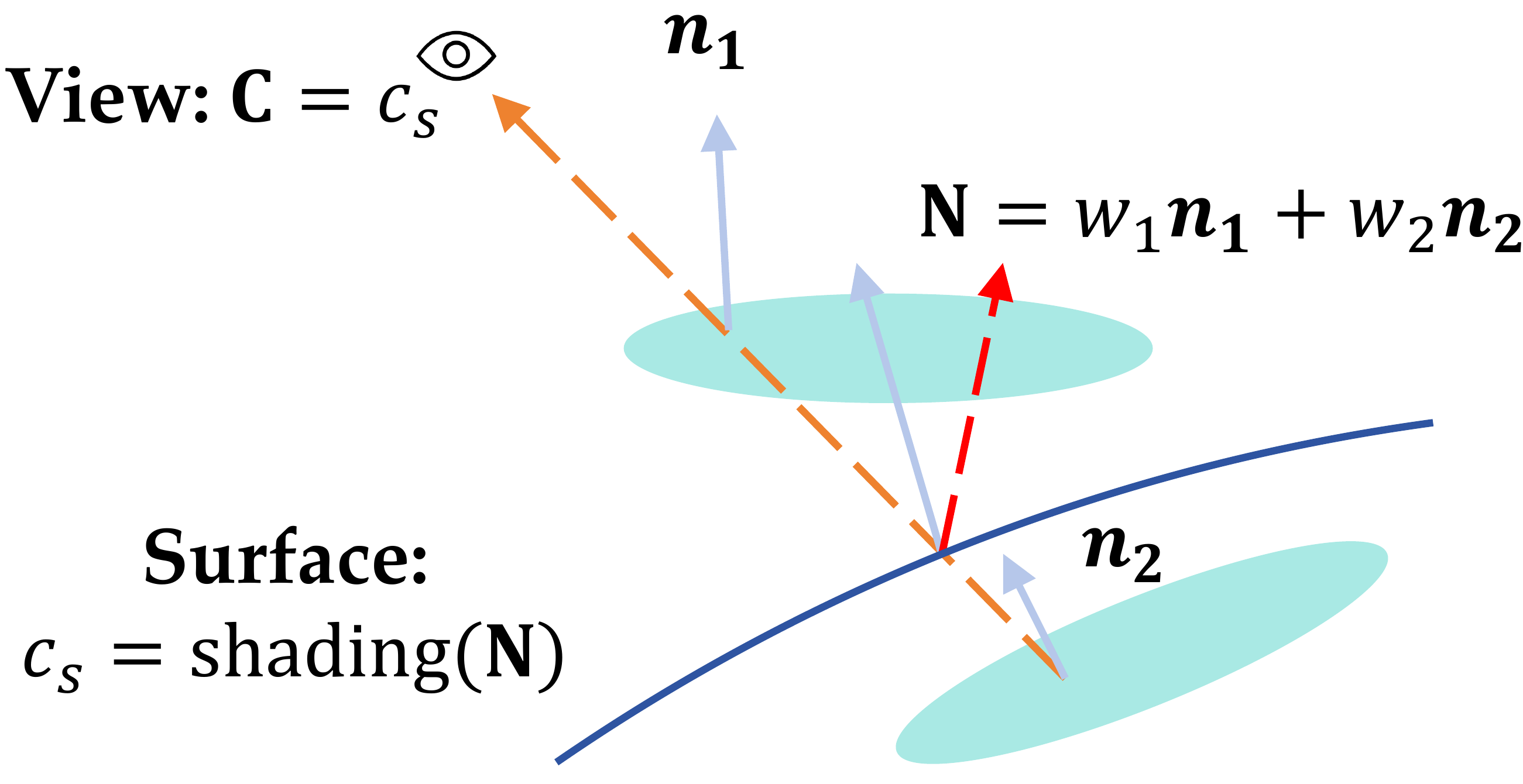}
    \label{fig:motivation_prefilter_sur}}
    \quad

  \subfloat[Impacts of the order of shading and $\alpha$-blending on different materials]{
    \includegraphics[width=0.98\linewidth]{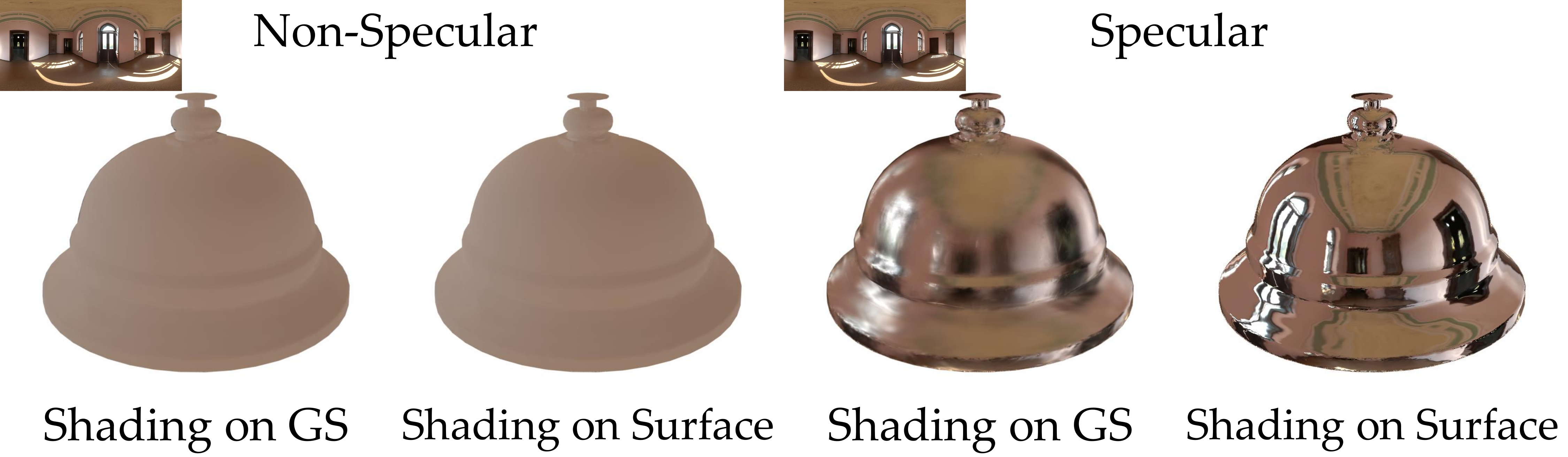}
    \label{fig:motivation_deferdemo}}
  
  \caption{\textbf{Normal Map Prefiltering.} $\mathbf{n_1}, \mathbf{n_2}, \mathbf{N}$ represent the normal of Gaussian 1, Gaussian 2, and the true surface, respectively. $c_1, c_2, c_s$ represent the shading results of Gaussian 1, Gaussian 2, and the true surface, respectively. $\mathbf{C}$ represents the color of the pixel, and $w_1$ and $w_2$ represent the weights for $\alpha$-blending. (a) is the current approach used by 3D-GS-based IR methods~\cite{gao2023relightable,jiang2023gaussianshader}. (b) On the other hand, our normal map prefiltering considers the nonlinearity of normals with respect to specular color (c) which has a significant impact on the rendering results of glossy objects. }\label{fig:motivation_prefilter}
\end{figure}
Accurately predicting normals on reflective surfaces presents a formidable challenge.
Through Eqn.~(\ref{eq:diffuse_color}), it can be inferred that shading of non-specular objects satisfies the following equation with respect to the surface normals $\mathbf{n}$:
\begin{equation}
\begin{split}
    \sum_{k=1}^{N}w_k c_{\text{diffuse}}&=
\sum_{k=1}^{N}w_k{\mathbf{a}(1-m) \cdot \int_{\Omega }L(\omega _i)\frac{\omega _i\cdot\mathbf{n} }{\pi } d\omega _i}\\
&=\mathbf{a}(1-m) \cdot \int_{\Omega }L(\omega _i)\frac{\omega _i\cdot\left(\sum_{k=1}^{N}w_k\mathbf{n}\right) }{\pi } d\omega_i.
\end{split}
\end{equation}
Diffuse color is linear in the normal. Eqn.~(\ref{eq:alpha_blending}) also tells that $\alpha$-blending in 3D-GS~\cite{kerbl20233d} is a linear operation. Therefore, for diffuse objects, the order of $\alpha$-blending and shading has little effect on the final color. However, specular reflection does not satisfy this linear relationship, i.e.,
\begin{equation}
\begin{split}
    \sum_{k=1}^{N}w_k c_{\text{specular}}&=
\sum_{k=1}^{N}w_k{\int_{\Omega }L(\omega _i)\frac{DFG}{4(\omega _o\cdot \mathbf{n} )}  d\omega _i}\\
&\ne {\int_{\Omega }L(\omega _i)\frac{DFG}{4\left(\omega _o\cdot \left(\sum_{k=1}^{N}w_k\mathbf{n}\right) \right)}  d\omega _i}.
\end{split}
\end{equation}
Therefore, the order of shading and $\alpha$-blending is crucial for inverse rendering of glossy objects. Unfortunately, current methods mostly ignore the influence of this order, resulting in unsatisfactory inverse rendering results for glossy objects.

The above-mentioned two different orders are demonstrated in Fig.~\ref{fig:motivation_prefilter}. They differ when the shading function is non-linear in the normal. Most 3D-GS-based inverse rendering methods~\cite{jiang2023gaussianshader, gao2023relightable, liang2023gs} employ the gradient of the depth map as the normal supervision strategy. Since the depth map represents the depth of the true surface, it is necessary to perform shading on the surface instead of shading on 3D Gaussians from this perspective. Furthermore, we demonstrate the visual impact of the shading order for different materials in Fig.~\ref{fig:motivation_deferdemo}. Given the linear relationship between diffuse color and normal, the order of shading does not significantly affect the appearance of non-specular objects. 
However, for specular objects, performing shading first and then applying $\alpha$-blending leads to a blurred appearance caused by the linear blending of the ambient lighting sampled from multiple reflective directions (the red arrows in Fig.~\ref{fig:motivation_prefilter_gs}). This approach fails to accurately capture the environment reflections. In contrast, we reverse the order by blending the normals first, enabling us to maintain a more precise representation of the surface normal. By subsequently sampling the ambient lighting based on a single reflective direction (the red arrow in Fig.~\ref{fig:motivation_prefilter_sur}), we achieve a more realistic rendering of specular objects. These different rendering results stem from the nonlinear relationship between the shading function and the normal for glossy objects. Our proposed normal prefiltering strategy successfully address this issue.

\subsection{Micro-facet Geometry Segmentation Prior}

\begin{figure}[ht]
  \centering
  
  \subfloat[ Large r and low-frequency $\mathbf{n}$]{
    \includegraphics[width=0.49\linewidth]{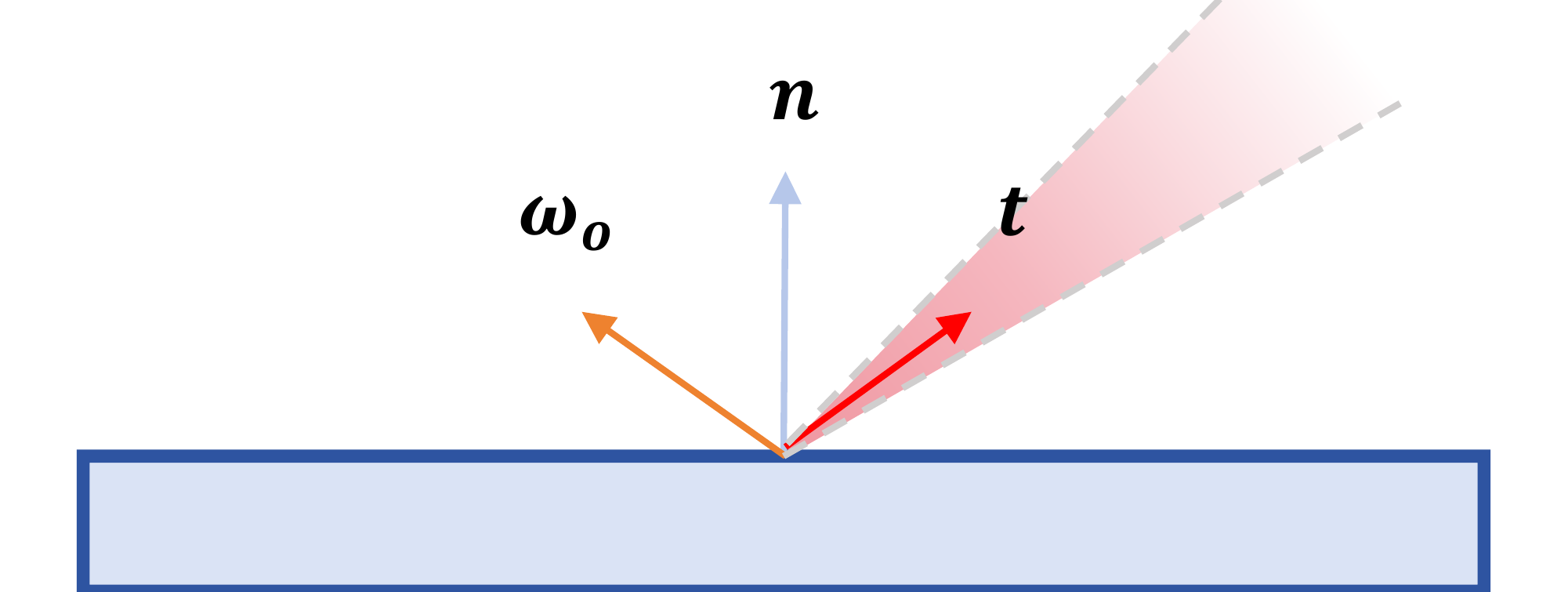}
    \label{fig:motivation_normal}}
  \subfloat[Small r and high-frequency $\mathbf{n}$]{
    \includegraphics[width=0.49\linewidth]{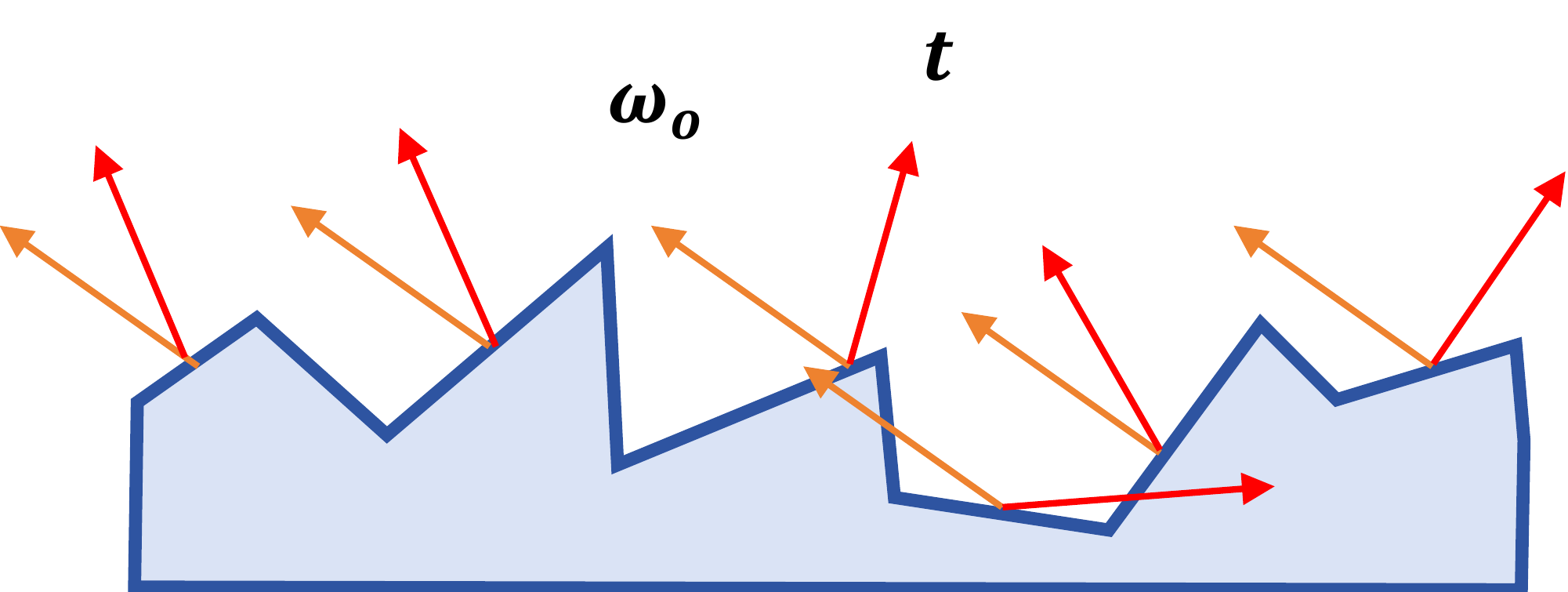}
    \label{fig:motivation_roughness}}
    \quad
  \subfloat[Different roughness-normal pairs corresponding to the GT image]{
    \includegraphics[width=1\linewidth]{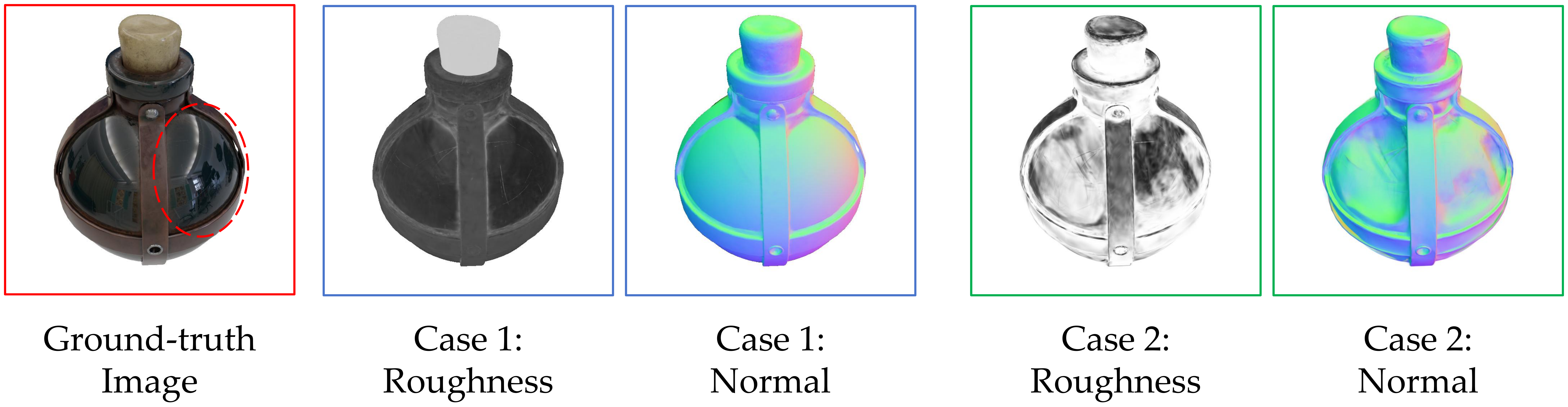}
    \label{fig:motivation_cmp}}
  
  \caption{\textbf{Ambiguities Between Macroscopic Normal and Microscopic Roughness.}  $\boldsymbol{\omega_o},\mathbf{n},\mathbf{t},r$ represent outgoing view direction, normal, reflected light, and roughness respectively. (a) represents large roughness and simple normal, while (b) is the opposite, with complex normal and each plane can be considered as a specular reflection with smaller roughness. Both can represent the same reflective object, leading to the occurrence of Case 1 and Case 2 in (c). }\label{fig:motivation_sam}
\end{figure}

The failure in reconstructing glossy objects is mainly attributed to the ambiguities arising from the coupling of microscopic and macroscopic normal distributions introduced by the micro-facet model~\cite{cook1982reflectance}. The initial motivation behind the introduction of the micro-facet model is to represent complex materials in situations where the geometry is not excessively complex, thus reducing the storage overhead associated with geometry. The success of this model relies on the ability of the microscopic normal distribution $D(r, \mathbf{t})$ characterized by roughness $r$ to produce visual effects similar to the macroscopic normal $\mathbf{n}$. However, achieving similar visual effects poses a challenge for inverse rendering tasks. 

As shown in Fig. \ref{fig:motivation_normal} and \ref{fig:motivation_roughness}, a low-frequency macroscopic normal distribution with a large roughness and a high-frequency macroscopic normal distribution with a small roughness can model the same glossy appearance. This leads to different roughness-normal pairs for the individual object as shown in Fig. \ref{fig:motivation_cmp}.  The introduction of environmental light information leads to visual color inconsistencies in parts that were expected to have the same roughness, while parts that were expected to have different roughness exhibit visual color consistency as shown in Fig. \ref{fig:motivation_cmp}. This complicates the decoupling of microscopic roughness and macroscopic normal, making it challenging to obtain accurate solutions through simple inverse rendering gradient backpropagation.

Ideally, we can directly apply roughness priors to narrow down the solution space, effectively alleviating ambiguities. However, we still encounter challenges in estimating roughness, particularly when working with a single image, as it is impossible to completely separate this from the unknown lighting conditions. To address this issue, we propose a prior for segmenting the geometry of micro-facets, which helps eliminate cases like Case 2 in Figure \ref{fig:motivation_cmp}. In these cases, the solutions satisfy the rendering equations~\cite{kajiya1986rendering} but do not correspond to real-world glossy objects, unlike Case 1 shown in the same figure.
In specific terms, we segment the roughness of glossy objects, with each segment representing a consistent roughness value. This prior serves to constrain the micro-facet normal distribution and geometry of reflective objects.

\subsection{Micro-facet Geometry Segmentation Model}
Similar to semantic segmentation, the micro-facet geometry segmentation model can be formulated as a mapping function from the image domain to the output label domain: $I\to B$, which predicts a pixel-wise category label $B \in \{0,...,C\}^{H\times W}$, where $H$ and $W$ denote the image size, $C$ is the total number of roughness categories. Given that an object is usually composed of a limited number of materials, we assume that a single object can be composed of up to 6 roughness categories. Accordingly, we sort the values in the roughness map and select the top $C$ most frequently occurring values to to filter the corresponding pixels for each category. Pixels that are not selected are categorized as label $0$, excluding the background.

During training, we employ the cross-entropy loss defined as follows:
\begin{equation}
    \mathcal{L}_{ce}\left(I, B\right)=-\sum_{u=1}^{H} \sum_{v=1}^{W} \sum_{c=1}^{C} b_{(u, v), c} \log p_{(u, v)}\left(c \mid I\right),
\end{equation}
where $(u, v)$ are the pixel coordinates in the input image $I$, $c$ is the category index, $b_{(u, v), c} \in \left \{ 0,1 \right \}$ is entry in the one-hot vector of the ground-truth label, and $ p_{(u, v)}\left(c \mid I\right)$ is the predicted category probability for category $c$ at pixel $(u,v)$.

\begin{figure}[htb]
\begin{center}
    \addtolength{\tabcolsep}{-4pt}
    \begin{tabular}{cccc}
    
   Scene & Segmentation &  w/ prior & w/o prior\\
    \includegraphics[height=0.7in]{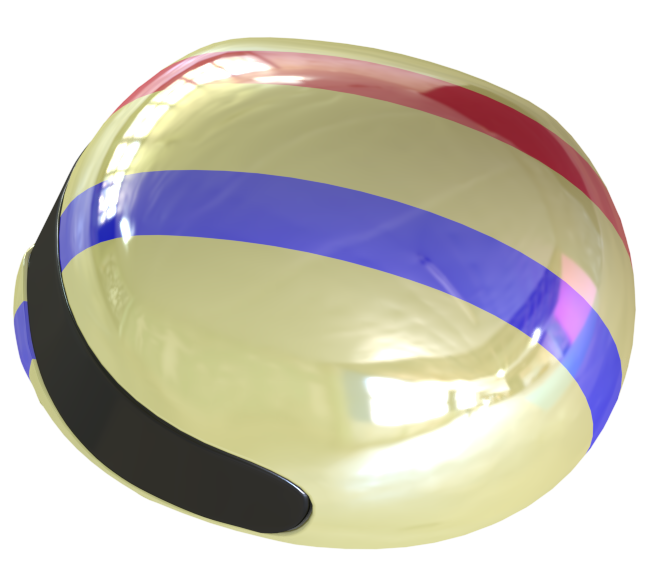} &
    \includegraphics[height=0.7in]{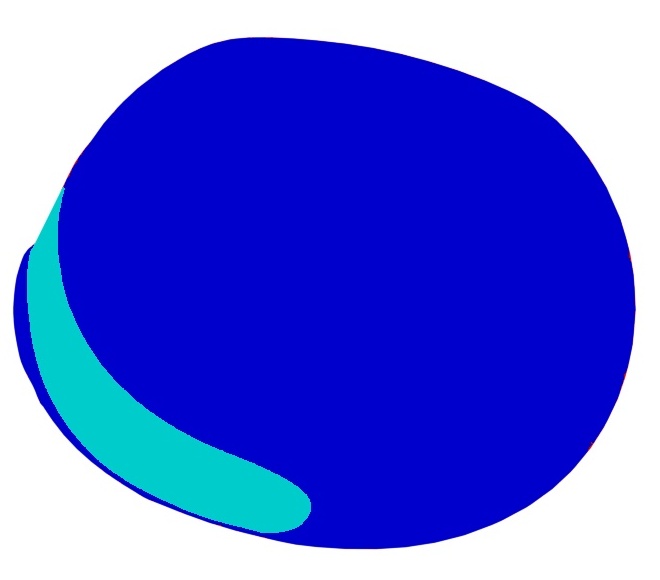} &
    \includegraphics[height=0.7in]{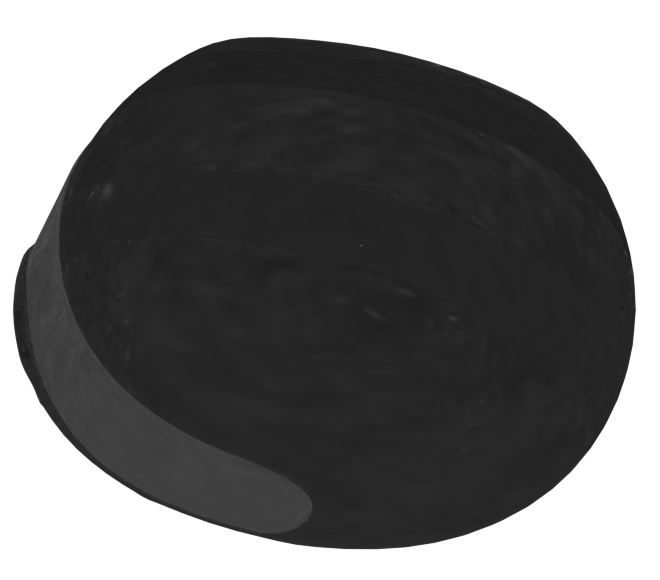} &
    \includegraphics[height=0.7in]{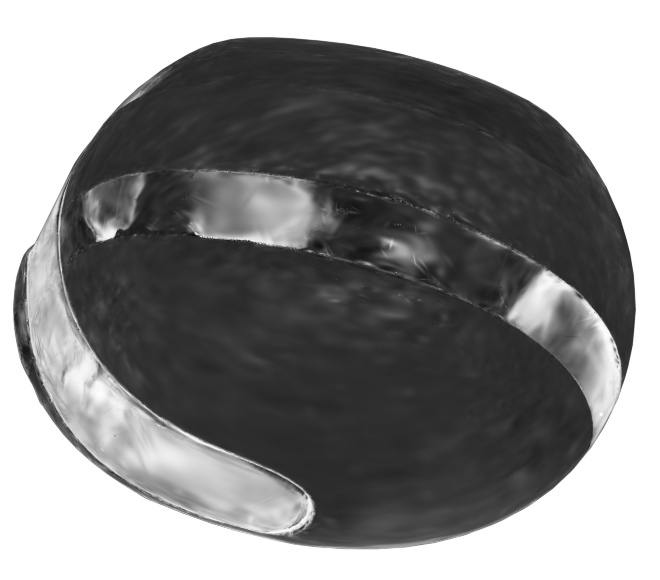}  \\ 
    
    \includegraphics[height=0.7in]{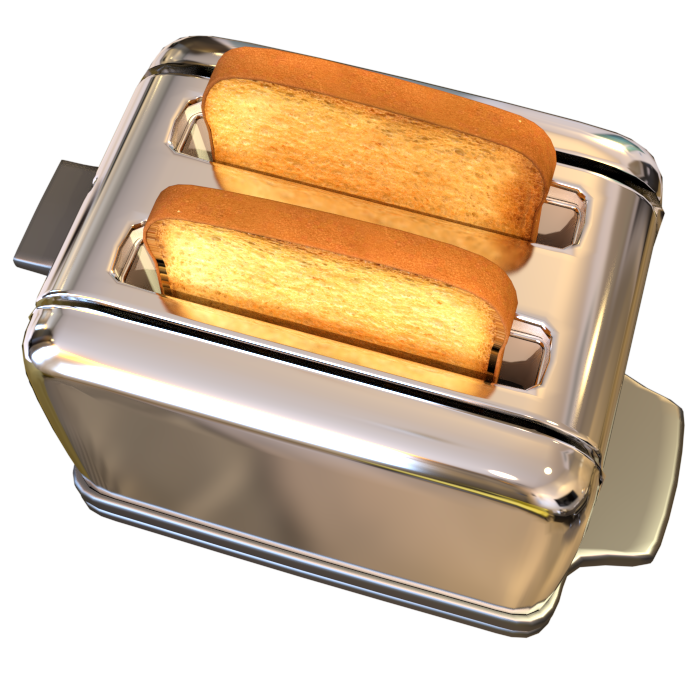} &
    \includegraphics[height=0.7in]{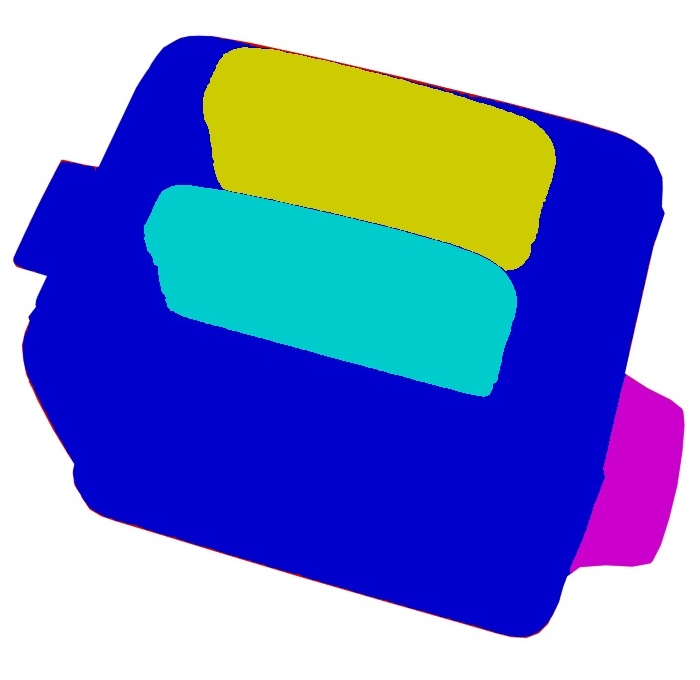} &
    \includegraphics[height=0.7in]{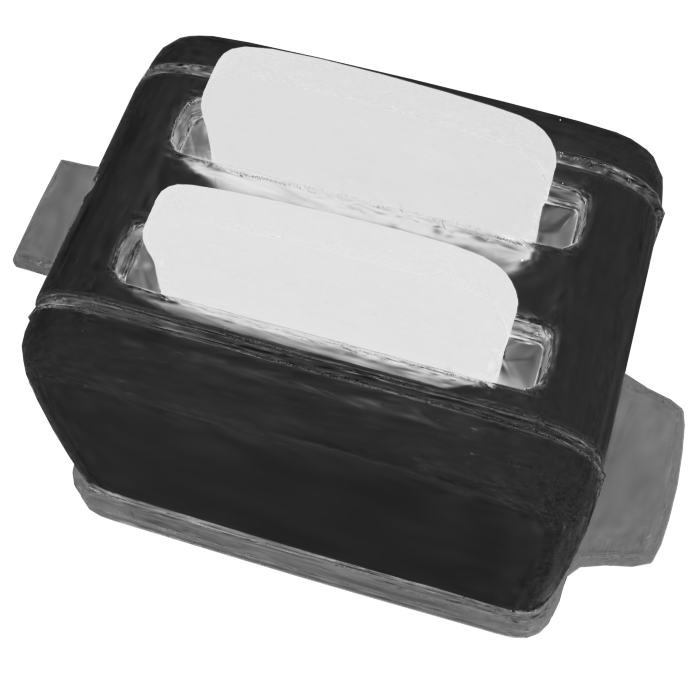}&
    \includegraphics[height=0.7in]{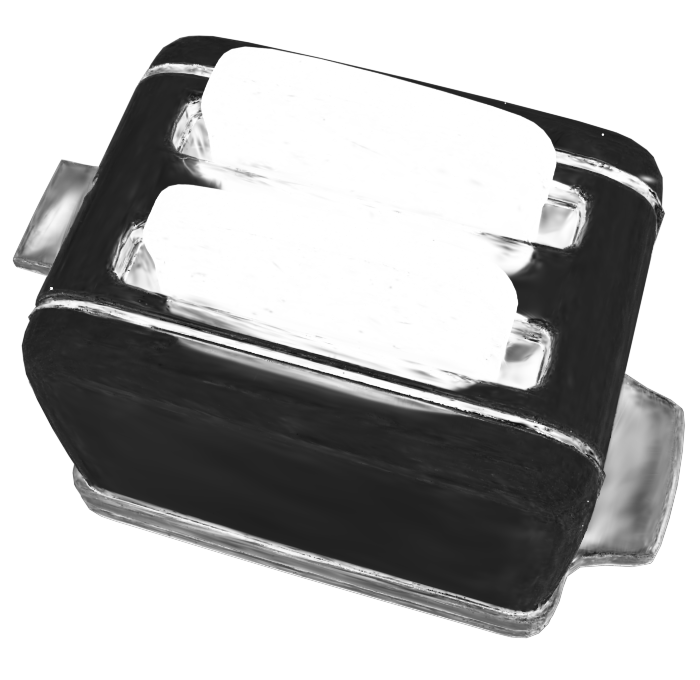} \\ 
    
    \includegraphics[height=0.7in]{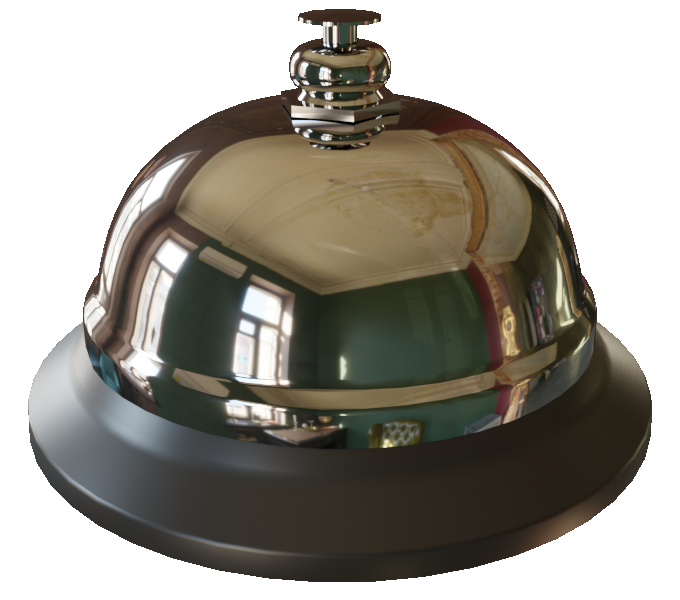} &
    \includegraphics[height=0.7in]{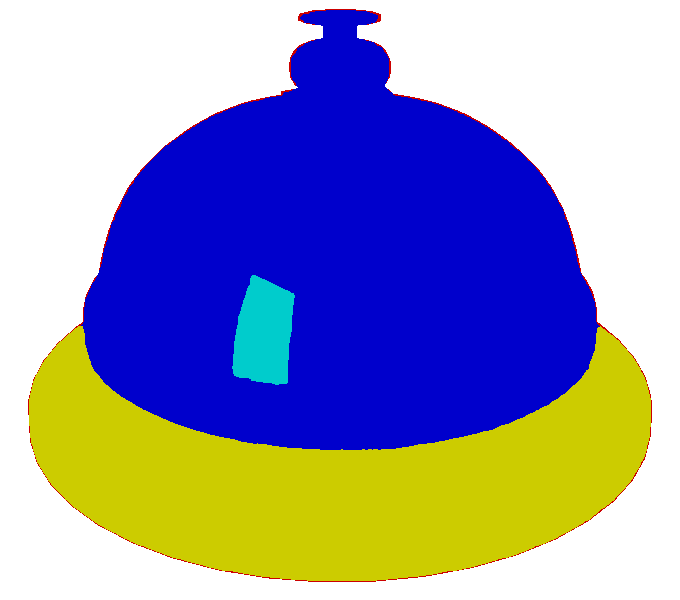} &
    \includegraphics[height=0.7in]{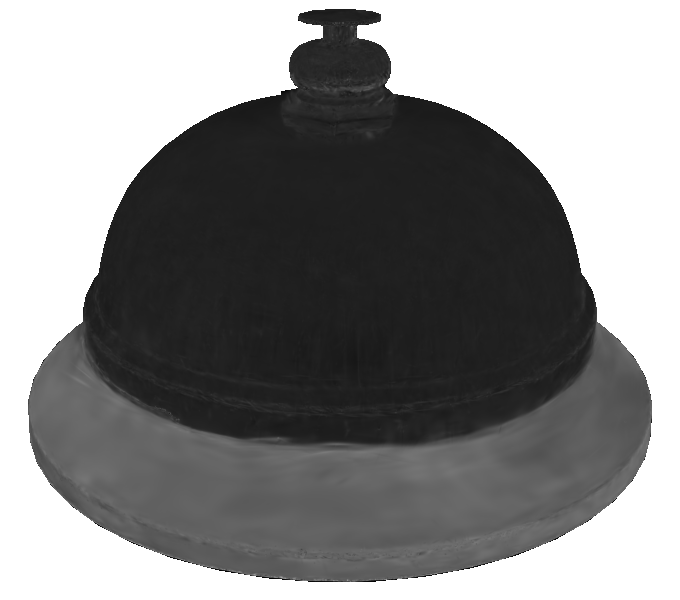} &
    \includegraphics[height=0.7in]{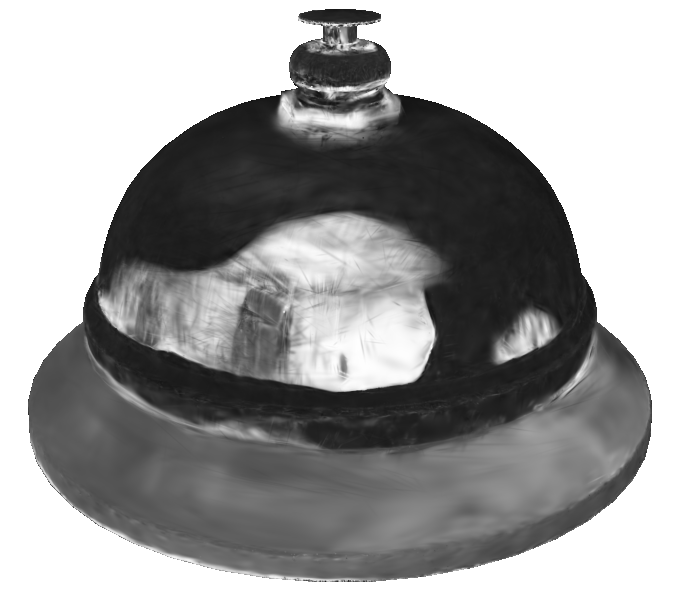} \\ 
    
    \includegraphics[height=0.7in]{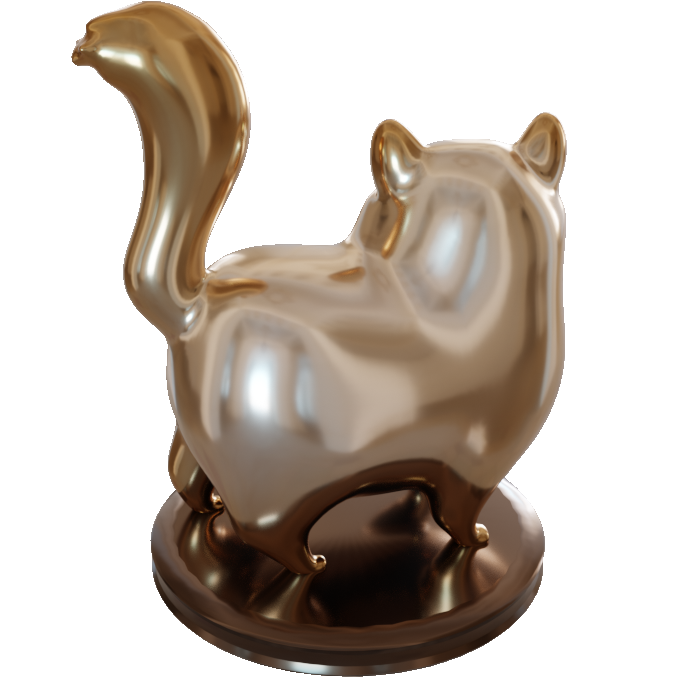} &
    \includegraphics[height=0.7in]{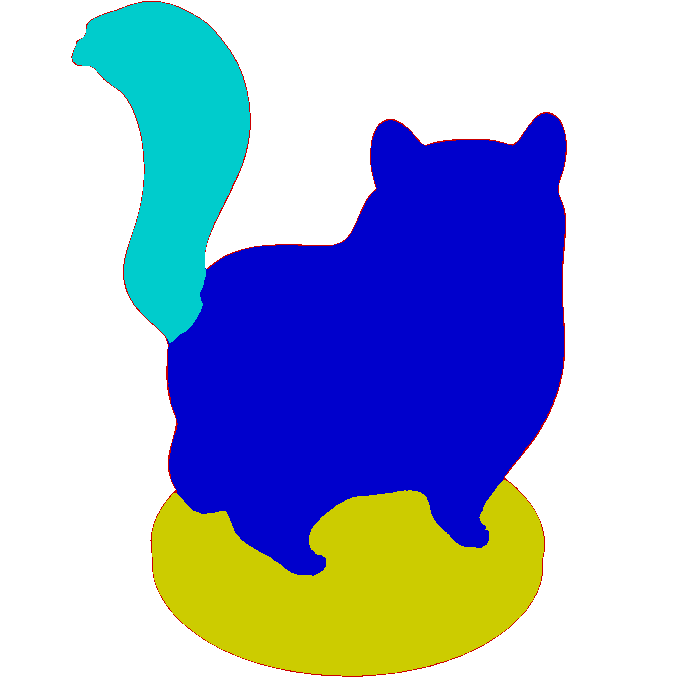} &
    \includegraphics[height=0.7in]{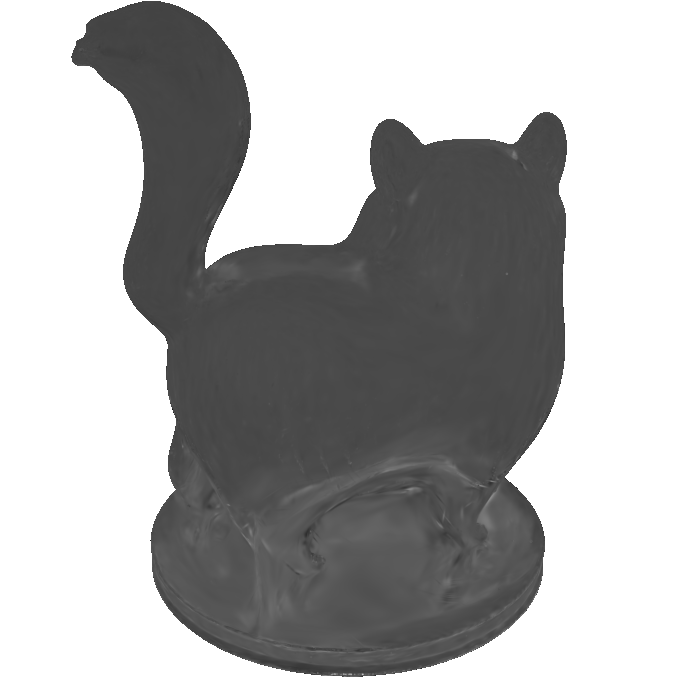}  &
    \includegraphics[height=0.7in]{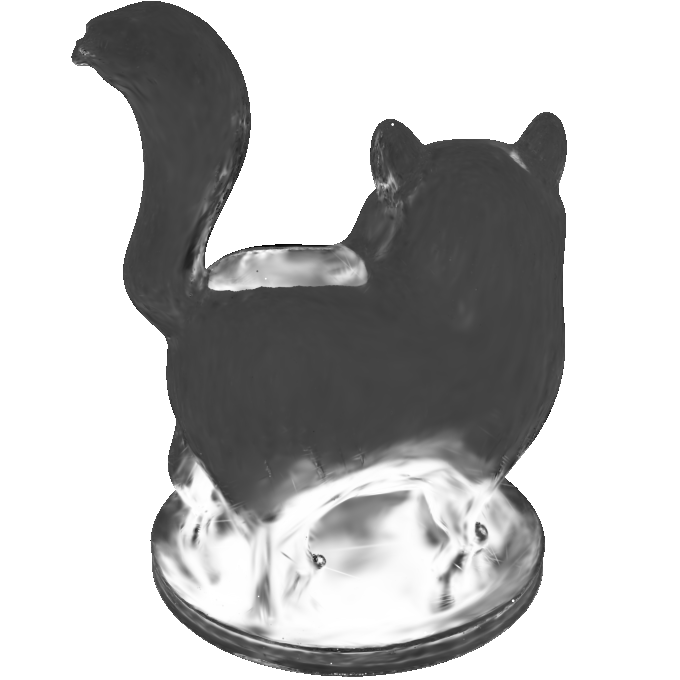} \\ 
    \end{tabular}
\end{center}
\caption{Visualizing the results of micro-facet geometry segmentation and the effect of the segmentation prior. Without the segmentation prior, the reconstructed roughness is noisy and lacks smoothness. Note that there is no available ground-truth roughness in these examples.}
\label{fig:segment_vis}
\end{figure}

Since roughness is an intrinsic characteristic of a material, it can be challenging to observe directly. To address this issue, we utilize existing pre-trained vision models as the backbone for our optimization process. 
DINOv2~\cite{oquab2023dinov2} models have been widely adopted across various semantic-related tasks, such as image retrieval and semantic segmentation, demonstrating robust performance even when using frozen weights without any fine-tuning. To leverage this impressive semantic capability in our segmentation model, we propose incorporating the DINOv2 encoder for feature extraction. Specifically, we utilize the DINOv2 encoder to capture rich feature representations and pair it with the DPT decoder~\cite{DPT} to perform effective segmentation. This design enables the network to learn to distinguish similar roughness regions by leveraging semantic-aware representations derived from DINOv2, utilizing our custom-built dataset. As shown in Fig. \ref{fig:segment_vis}, the segmentation model can accurately partition regions of varying roughness, thereby ensuring that the reconstructed roughness is both smooth and reasonable.

\subsection{The Inverse Rendering Framework}
Based on the micro-facet geometry segmentation prior and the normal map prefiltering strategy, we propose a method for reconstructing high-fidelity geometries and materials of glossy objects from a set of posed images $I^s$. An overview of our approach is presented in Fig.~\ref{fig:pipeline}. 
Initially, we obtain the geometric and material attributes of each Gaussian through network inference. Subsequently, using $\alpha$-blending as described in Eqn.~(\ref{eq:alpha_blending}), we prefilter the maps for each attribute. We then utilize these maps along with the rendering equation~\cite{kajiya1986rendering}, to perform shading and produce the final image, denoted as $\hat{I}$. Due to the fact that the maps represent the attributes of the true surface rather than the 3D Gaussians, and considering the nonlinearity of shading with respect to the normal, our approach yields more accurate results compared to directly shading on 3D Gaussians. Furthermore, the calculation of environment mipmaps follows the same methodology as that of \cite{liang2023gs,jiang2023gaussianshader}. 

During training, the learnable parameters and MLPs are jointly optimized by minimizing the reconstruction loss $\mathcal{L}$, represented as:
\begin{equation}
\label{eq:loss}
    \mathcal{L} = \mathcal{L}_c +\lambda _s\mathcal{L}_s +\lambda_{e}\mathcal{L}_{e}+\lambda_{n}\mathcal{L}_{n}.
\end{equation}
$\mathcal{L}_{c}$ denotes the combined image loss, similar to the approach used in Scaffold-GS~\cite{lu2023scaffold}, defined as follows:
\begin{equation}
    \mathcal{L}_{c} = \mathcal{L}_{1} + \lambda _{SSIM}\mathcal{L}_{SSIM} + \lambda _{vol}\mathcal{L}_{vol}.
\end{equation}
In this equation, the parameters $\lambda _{SSIM}$ and $\lambda _{vol}$ are set to 0.2 and 0.001, respectively.

The roughness constraint loss $\mathcal{L}_{e}$ leverages the consistent roughness within segmented regions to supervise the roughness map $I_\rho$. Specifically, it is defined as follows:
 \begin{gather}
    \mathbb{S }(c) = \{ I_\rho (u,v) \mid \hat{B} (u,v) = c\},\\
    \mathcal{L}_{e} = \sum_{c \in C}\sqrt{\frac{1}{\left | \mathbb{S}(c) \right | -1} \sum_{i=1}^{\left | \mathbb{S}(c) \right | } (z_i - \bar{z})^2}, 
\end{gather}
where $z_i \in \mathbb{S}(c)$ and $\bar{z}$ represents the mean value of the elements within $\mathbb{S}(c)$. Additionally, $\hat{B}$ represents the segmented labels obtained from the micro-facet geometry segmentation model applied to the images $\hat{I}$. During the inverse rendering process, minimizing loss $\mathcal{L}_{e}$ involves fitting a more accurate macroscopic normal distribution, thereby achieving improved geometry reconstruction of glossy objects.

$\mathcal{L}_{n}$ is a normal loss that correlates the local geometry with the predicted normals by reducing the discrepancy between the gradient normal maps $\hat{n}_D$ that are calculated from the rendered depth maps and the rendered normal maps $\hat{n}$, defined as:
\begin{equation}
    \mathcal{L}_{n}=\left \| \hat{n} -\hat{n}_D\right \|. 
\end{equation}
The parameters $\lambda _s$, $\lambda _e$, and $\lambda _n$ are all set to 0.01 in our experiments.

\section{Implementation details}
In this section, we present our datasets, and training details.

\paragraph{Datasets and Metrics} We conduct experiments on two synthetic datasets, Shiny Blender \cite{verbin2022ref} and Glossy Synthetic \cite{liu2023nero}, and two real datasets, Stanford-ORB \cite{kuang2024stanford} and NeILF-HDR \cite{zhang2023neilf++}. To further demonstrate the robustness of our method, we also captured two real scenes under indoor lighting. Following the official guidelines, images in the Stanford-ORB dataset are resized to a resolution of $512 \times 512$ for training and testing. We utilize tone mapping to convert high-dynamic range (HDR) images from the NeILF-HDR dataset into low-dynamic range (LDR) images. To evaluate the quality of the estimated normals, we compare estimated normal images with the ground truth using the mean angular error (MAE). Regarding rendering quality, we use PSNR, SSIM and LPIPS \cite{zhang2018unreasonable} to evaluate the similarity between the rendered images and the ground truth images.

\paragraph{Micro-facet Geometry Segmentation Prior Model Training} We collect training data by randomly picking 100,000 objects with roughness maps from the Objaverse dataset \cite{deitke2023objaverse}. We render each object in ten random poses against five random HDR environment maps to capture a variety of realistic lighting conditions. The base learning rate of the pre-trained encoder is set as $5e-6$, while the randomly initialized decoder uses a $10\times$ larger learning rate. We employ the AdamW optimizer and reduce the learning rate by a factor of $0.8$ every five epochs. Additionally, we apply horizontal flipping as a form of data augmentation. The training process runs for approximately three days on four NVIDIA A100 GPUs.

\paragraph{Inverse Rendering} The GlossyGS model is trained on a single NVIDIA V100 GPU with 32GB memory, completing the training process in approximately 1 hour for images at an $800 \times 800$ resolution. It achieves a rendering speed of around $30$ FPS at the same resolution on the V100 GPU. Additionally, the learnable environment map has a resolution of $256 \times 512$.

\section{Results}

\begin{table}[]
  \small
  \centering
\caption{Quantitative comparison of normal reconstruction and relighting results on the Shiny Blender~\cite{verbin2022ref} dataset. \textbf{Bold} means the best performance and \underline{underline} means the second best. \label{tab:refnerf}}
\begin{tabular}{c|ccccc}

\hline
               & NeRO  & NMF & Gshader       & GSIR          & Ours           \\
\hline
PSNR$\uparrow$                              & 21.53                    & \underline{25.17}         & 21.17                       & 21.95                    & \textbf{25.72}           \\
SSIM$\uparrow$                              & 0.858                    & \underline{0.916}        & 0.867                       & 0.857                    & \textbf{0.93}            \\
LPIPS$\downarrow$                             & 0.144                    & \underline{0.119}         & 0.133                       & 0.152                    & \textbf{0.103}           \\
normal MAE$\downarrow$                        & 3.02                     & \textbf{1.81} & 6.81                        & 9.06                     & \underline{2.82}            \\
training time$\downarrow$ & 12h                   & 4h         & 1.2h              & \underline{1.1h}            & \textbf{1h}            \\
rendering FPS$\uparrow$ & 0.03                     & 0.04          & 16           & \underline{30}              & \textbf{32}    \\
\hline
\end{tabular}
\end{table}

\begin{table}[]
  \small
  \centering
\caption{
\label{tab:nero}Quantitative comparison of novel view synthesis results on the Glossy Synthetic \cite{liu2023nero} dataset. \textbf{Bold} means the best performance.}
\begin{tabular}{c|ccccc}
\hline
       & NeRO & NMF & Gshader & GSIR  & Ours           \\
      \hline
PSNR$\uparrow$   & 29.73 & 28.22 & 28.65   & 25.35 & \textbf{30.46} \\
SSIM$\uparrow$    & 0.904 & 0.926 & 0.943   & 0.899 & \textbf{0.960} \\
LPIPS$\downarrow$  & 0.324 & 0.091 & 0.072   & 0.101 & \textbf{0.049} \\
           \hline
\end{tabular}
\end{table}

In this section, we compare our method with the following two types of baseline methods. (1) \textbf{3D-GS-based methods}: Gshader~\cite{jiang2023gaussianshader} and GSIR~\cite{liang2023gs}. (2) \textbf{NeRF-based methods}: neural microfacet fields (NMF)~\cite{mai2023neural} and NeRO~\cite{liu2023nero}.  For all these methods, we use their official implementations. While NeRF-based methods can achieve impressive reconstruction results, they are notably time-consuming. 
\subsection{Results on synthetic data}
We conduct quantitative and qualitative evaluations on two synthetic datasets, examining normal reconstruction, material decomposition, novel view synthesis, and relighting.
\begin{figure*}[p]
\begin{center}
    \addtolength{\tabcolsep}{-4pt}
    \begin{tabular}{m{2.5cm}<{\centering} m{2.5cm}<{\centering} m{2.5cm}<{\centering} m{2.5cm}<{\centering} m{2.5cm}<{\centering} m{2.5cm}<{\centering}m{2.5cm}<{\centering}}
    
    Scene & NeRO & NMF & Gshader & GSIR &Ours & GT \\
    \includegraphics[width=0.9in]{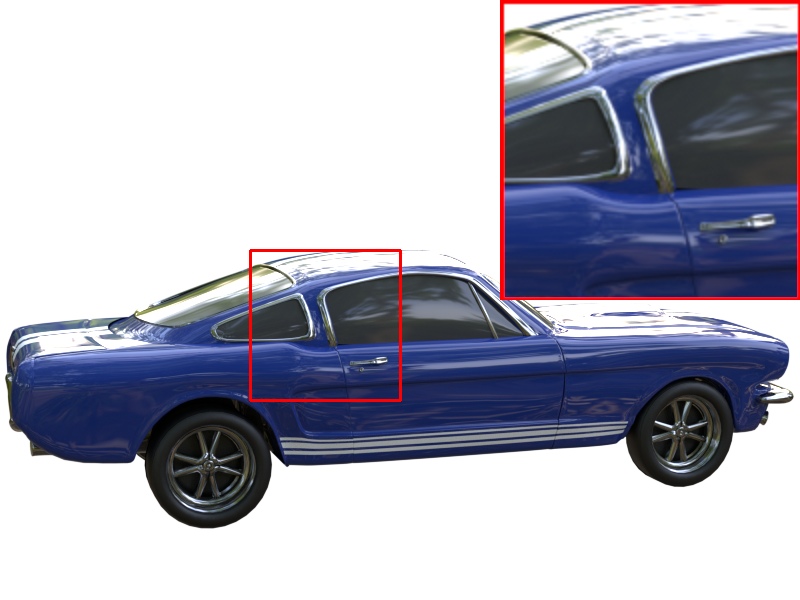} &
    \includegraphics[width=0.9in]{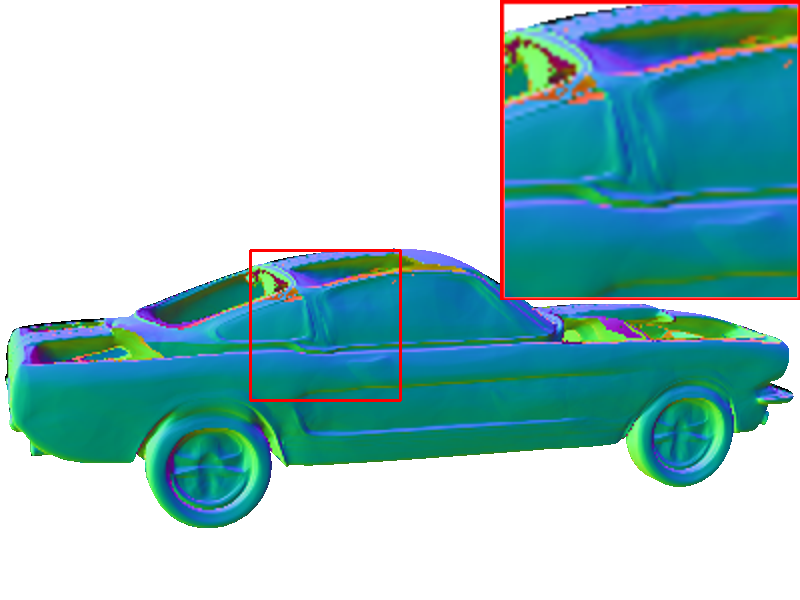} &
    \includegraphics[width=0.9in]{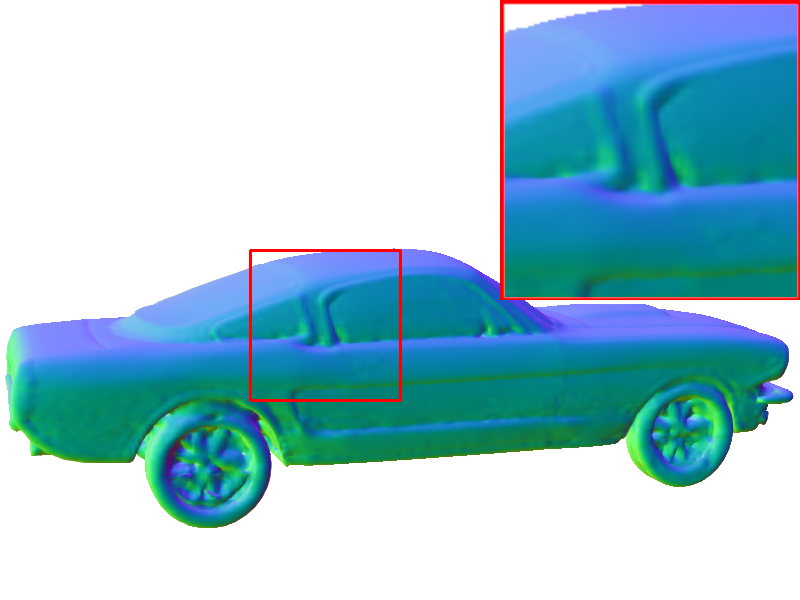} &
    \includegraphics[width=0.9in]{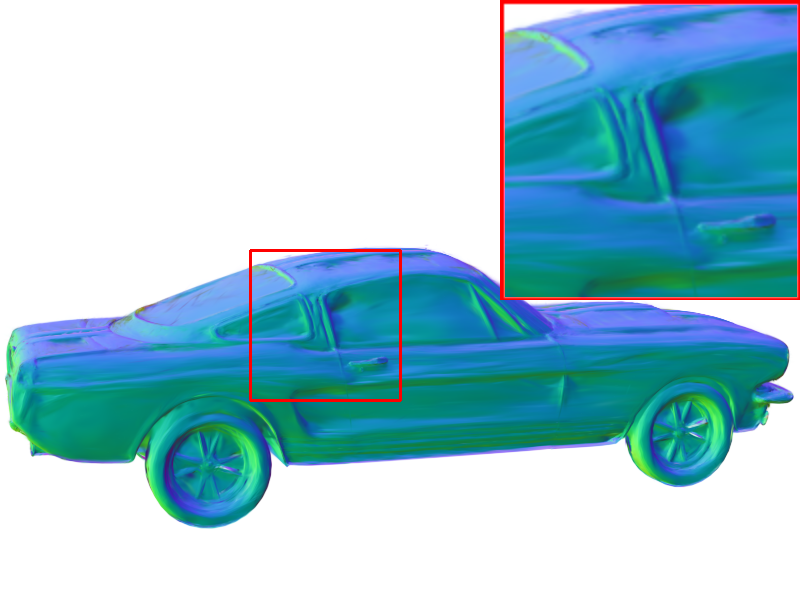} &
    \includegraphics[width=0.9in]{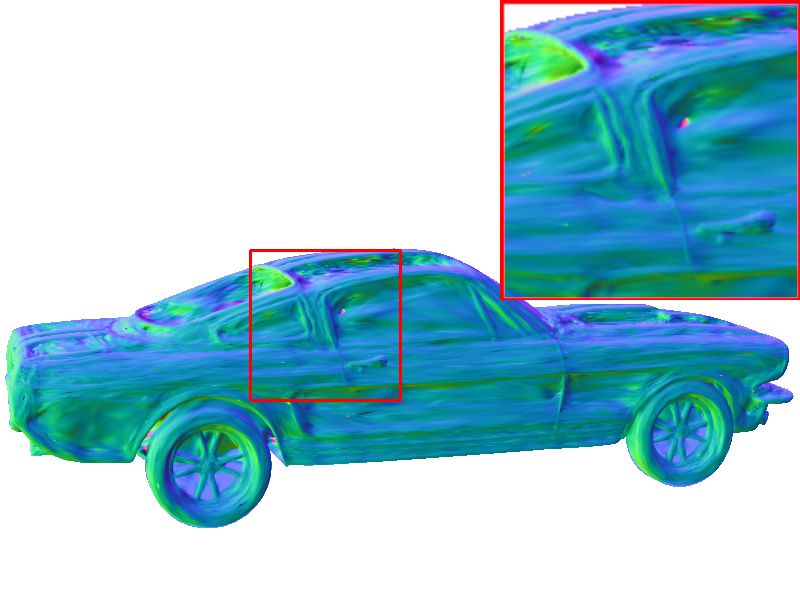} &
    \includegraphics[width=0.9in]{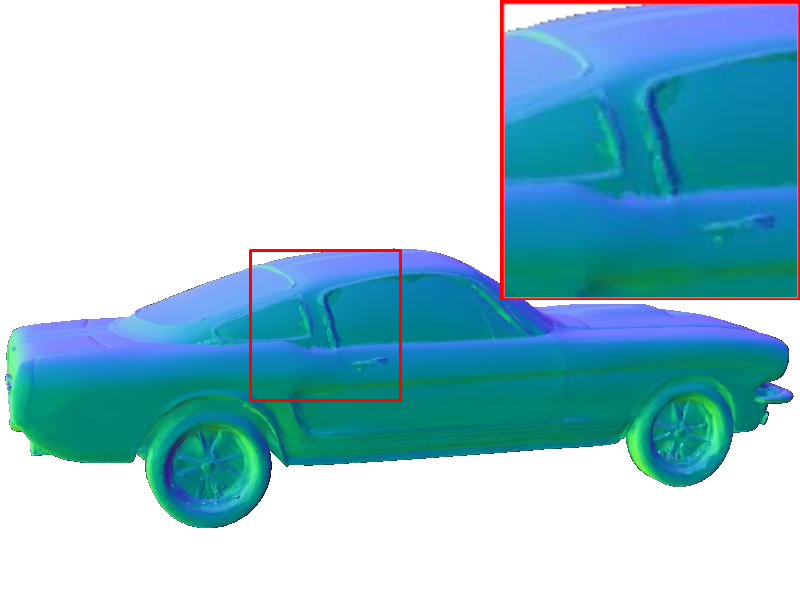} &
    \includegraphics[width=0.9in]{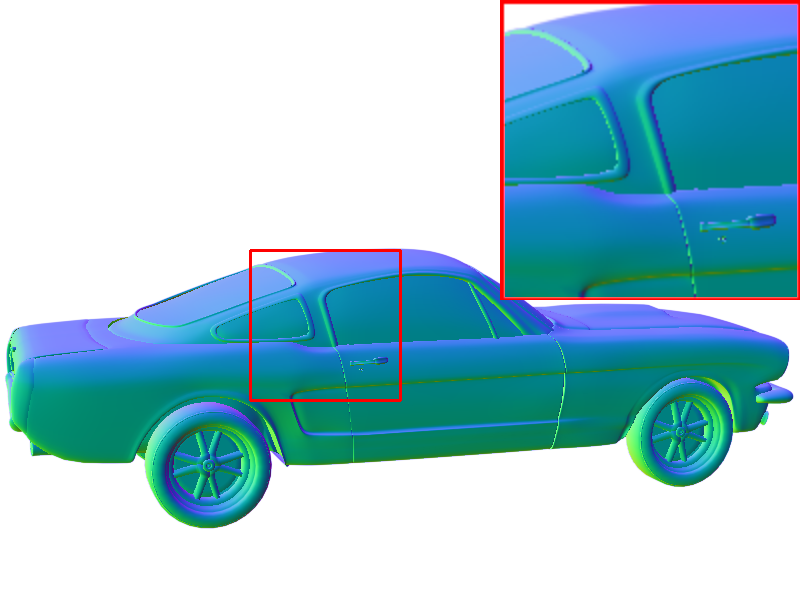} \\
    Car & & & & & & \\
    \includegraphics[height=1in]{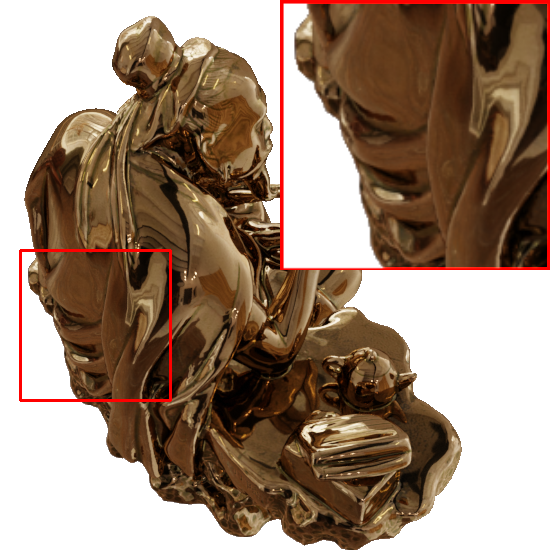} &
    \includegraphics[height=1in]{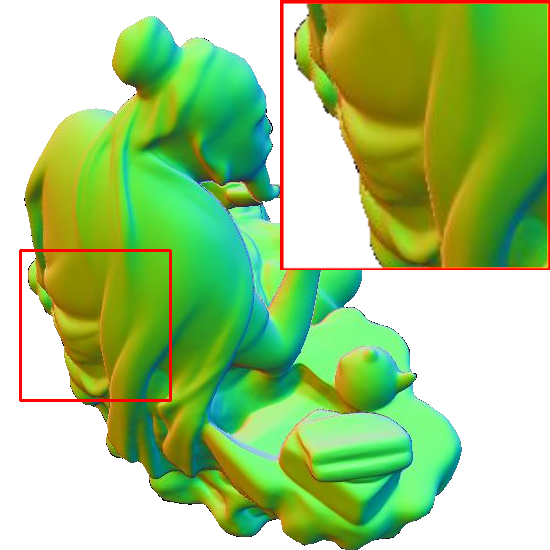} &
    \includegraphics[height=1in]{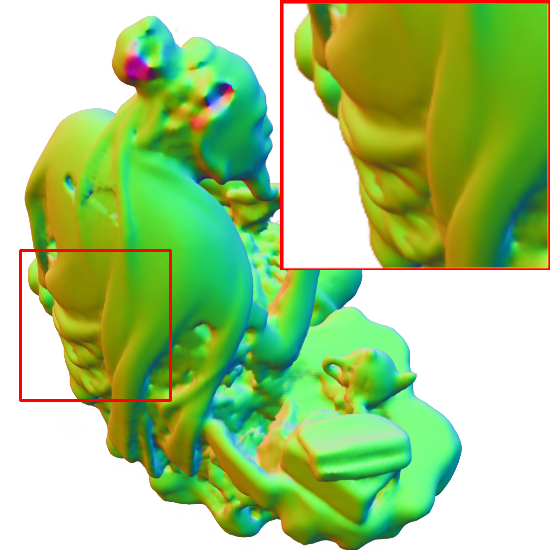} &
    \includegraphics[height=1in]{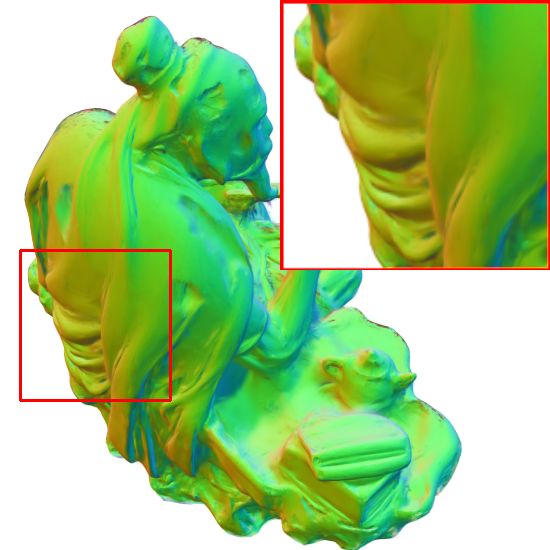} &
    \includegraphics[height=1in]{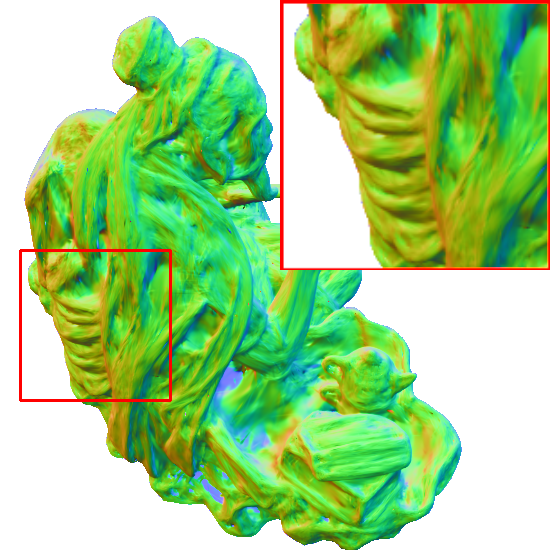} &
    \includegraphics[height=1in]{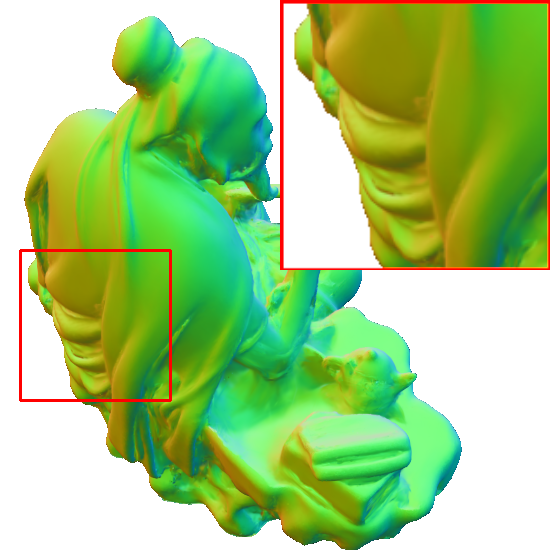} &
    \resizebox{0.2\columnwidth}{!}{N/A} \\
    Luyu & & & & & & \\
    
    \includegraphics[height=1in]{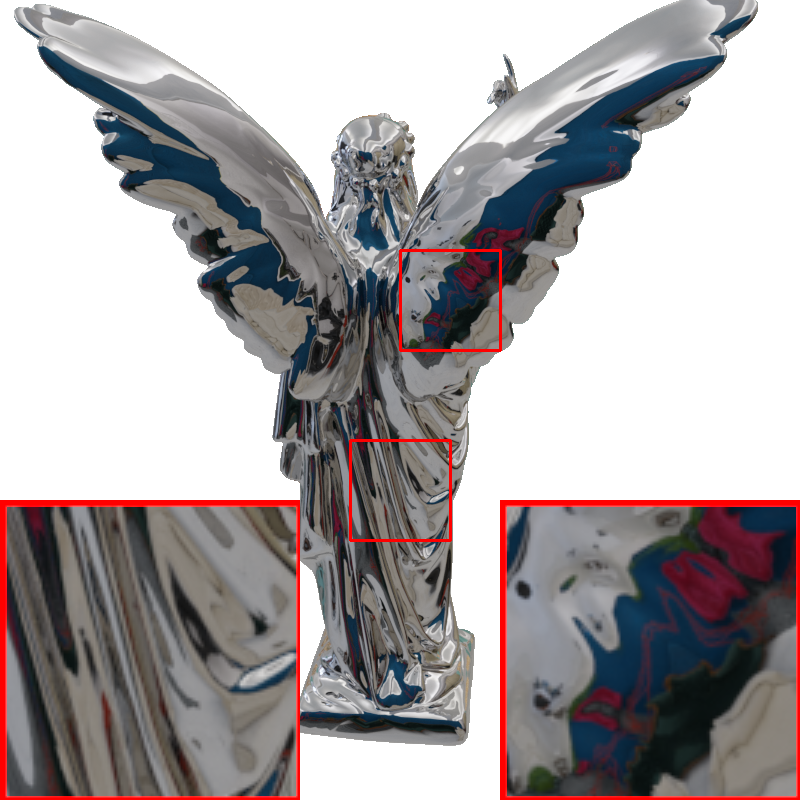} &
    \includegraphics[height=1in]{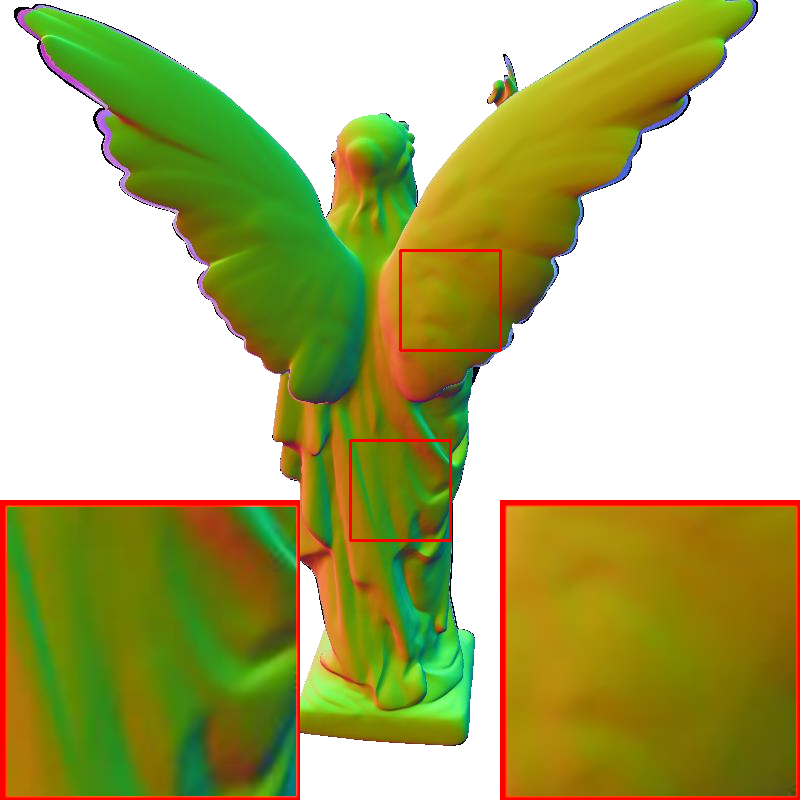} &
    \includegraphics[height=1in]{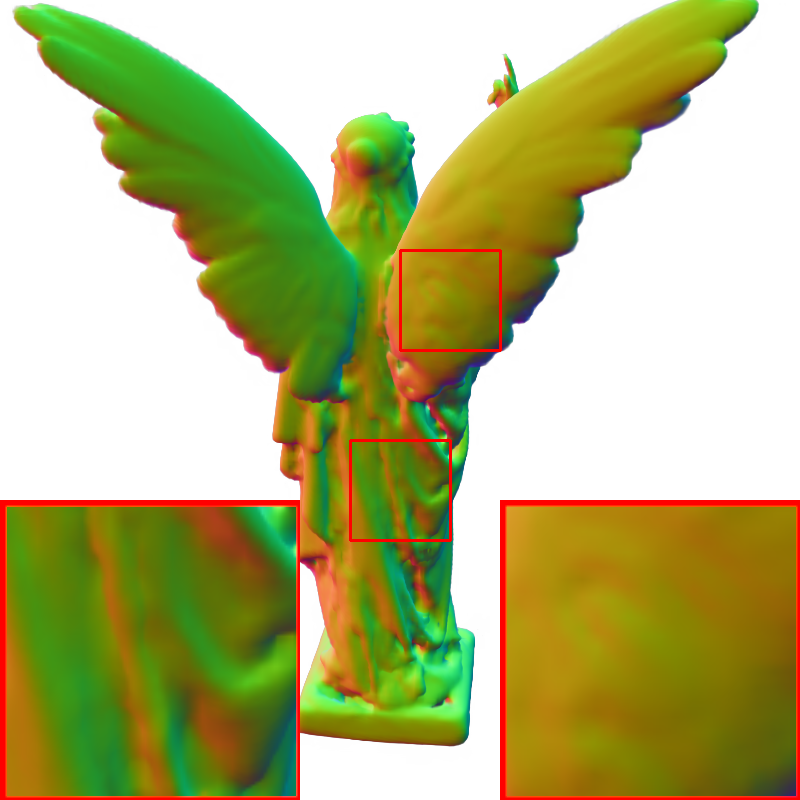} &
    \includegraphics[height=1in]{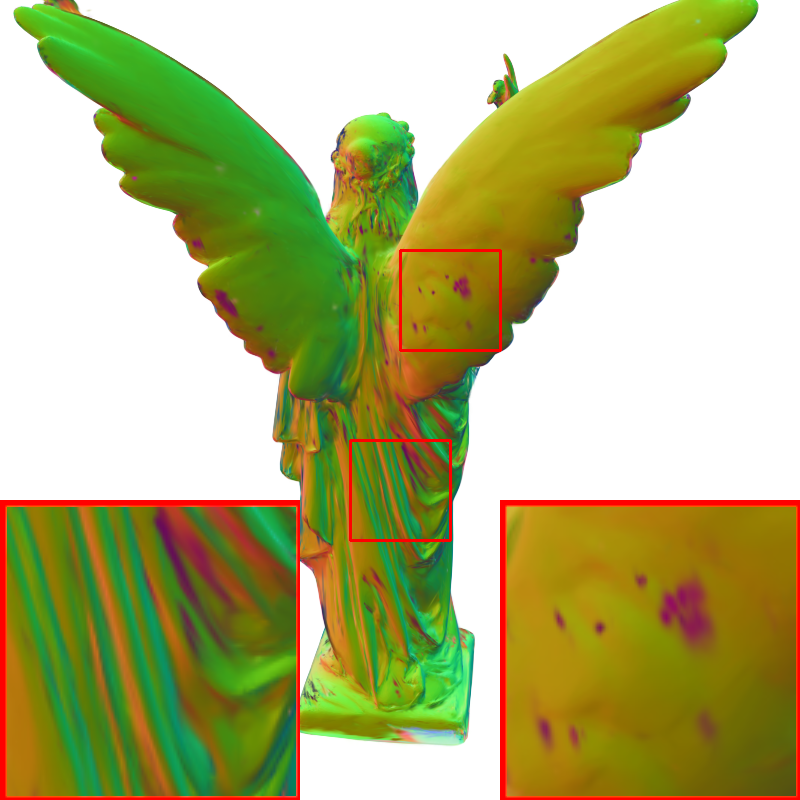} &
    \includegraphics[height=1in]{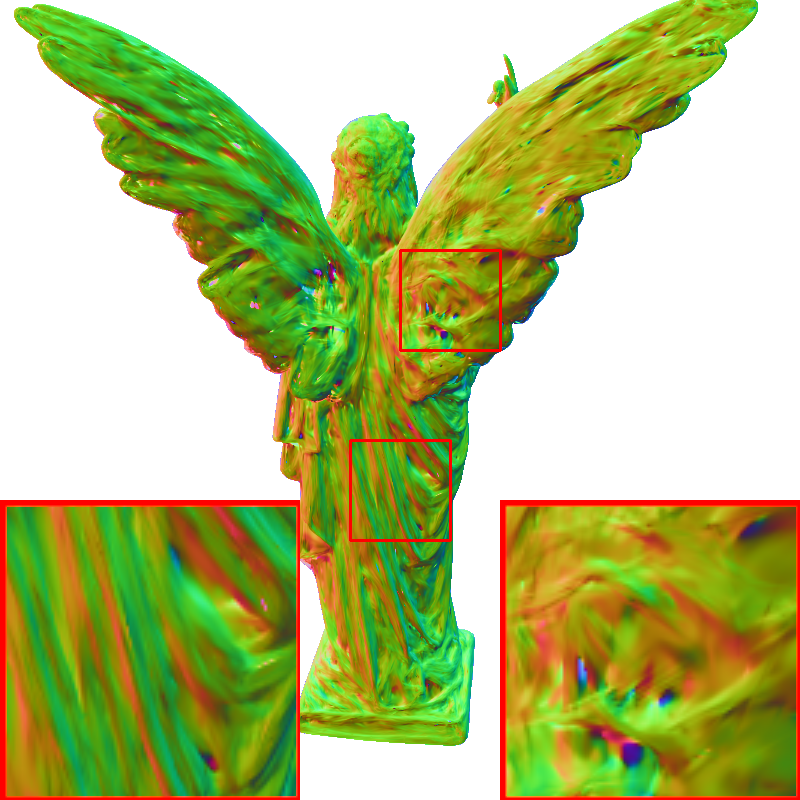} &
    \includegraphics[height=1in]{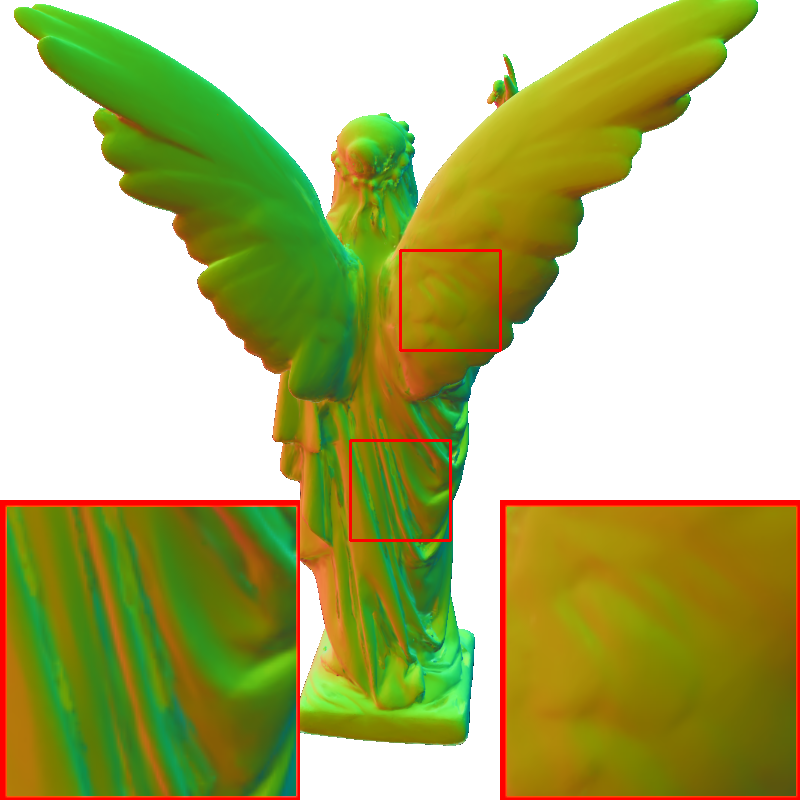} &
    \resizebox{0.2\columnwidth}{!}{N/A} \\
    Angel & & & & & & \\
    \end{tabular}
\end{center}
\caption{Comparison of normal reconstruction among our method, NMF~\cite{mai2023neural}, NeRO \cite{liu2023nero}, Gshader \cite{jiang2023gaussianshader} and GSIR \cite{liang2023gs} on the Shiny Blender \cite{verbin2022ref} (the first row), Glossy Synthetic datasets \cite{liu2023nero}(the second and third rows). Note that there is no available ground-truth normal in the Glossy Synthetic dataset.}
\label{fig:normal_syn}
\end{figure*}

\begin{figure*}[p]
\begin{center}
        \addtolength{\tabcolsep}{-4pt}
    \begin{tabular}{m{2.5cm}<{\centering} m{2.5cm}<{\centering} m{2.5cm}<{\centering} m{2.5cm}<{\centering} m{2.5cm}<{\centering} m{2.5cm}<{\centering}m{2.5cm}<{\centering}}
    
    Scene & NeRO albedo & GSIR albedo & Our albedo &  Our metallic & Our roughness & Our relighting\\
    \includegraphics[width=1in]{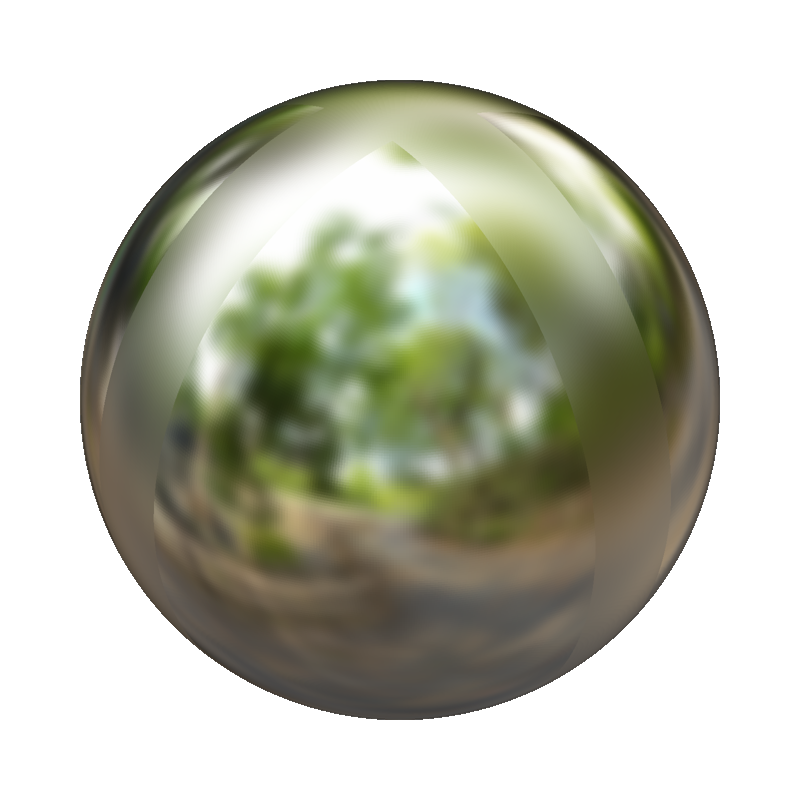} &
    \includegraphics[width=1in]{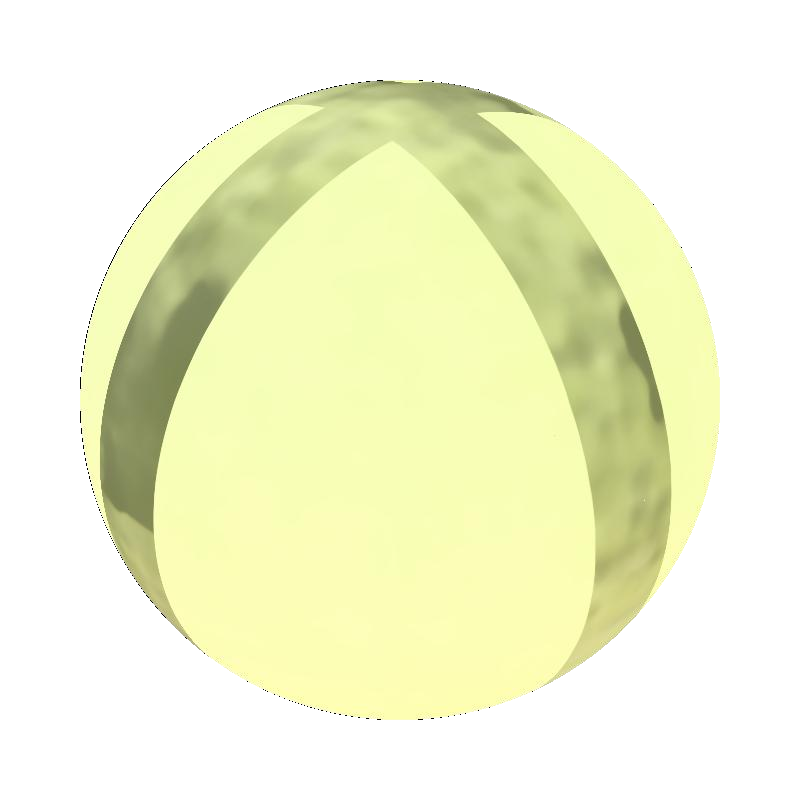} &
    \includegraphics[width=1in]{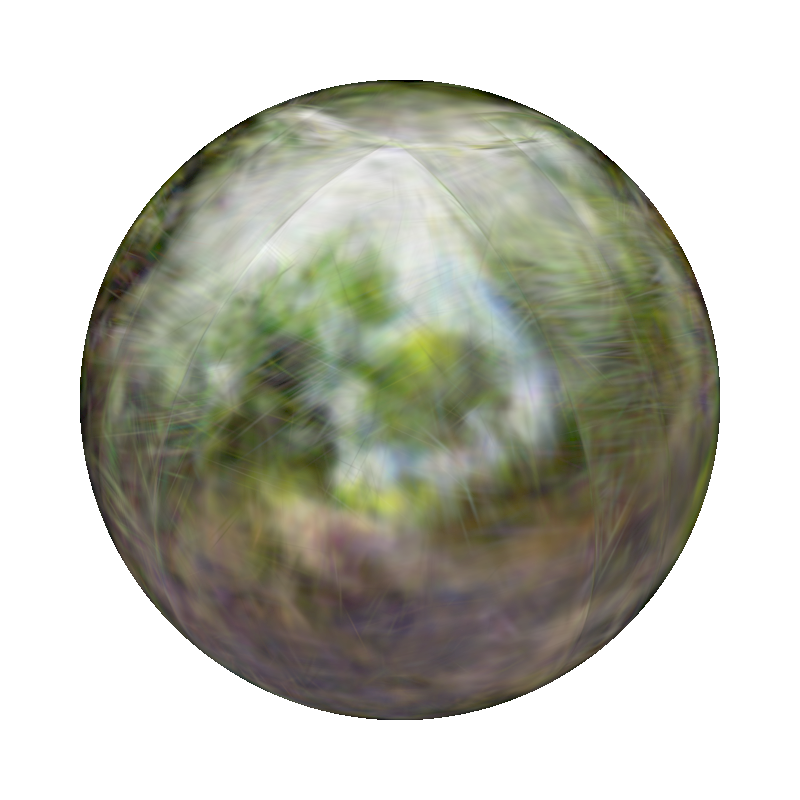} &
    \includegraphics[width=1in]{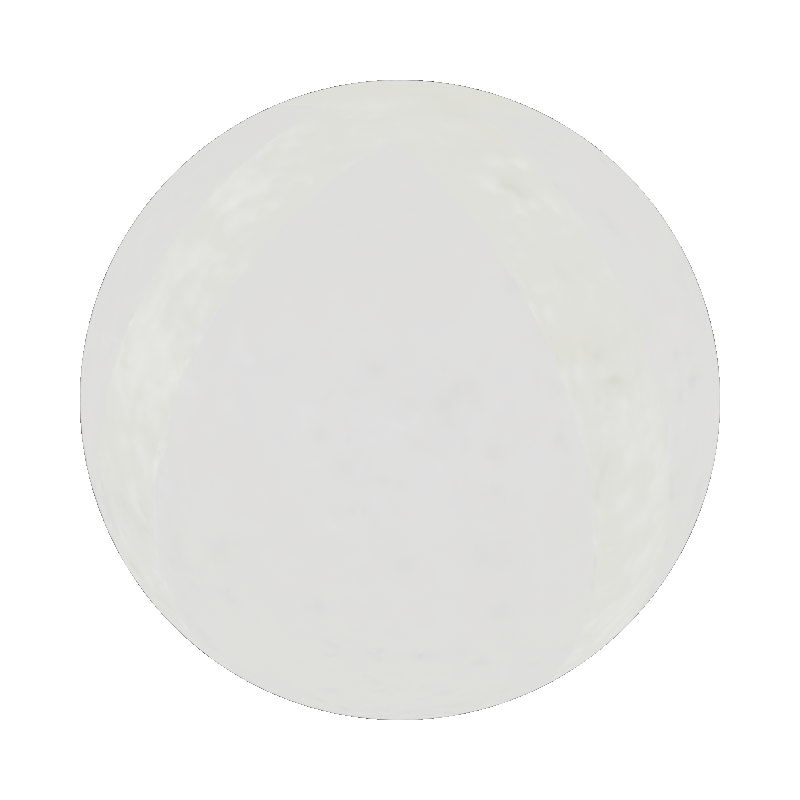} &
    \includegraphics[width=1in]{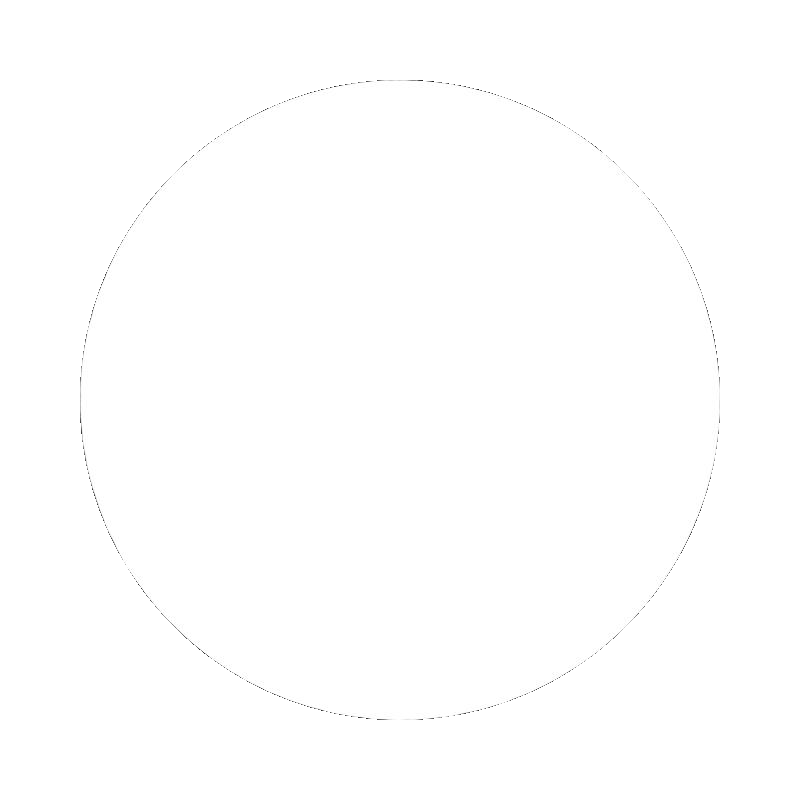} &
    \includegraphics[width=1in]{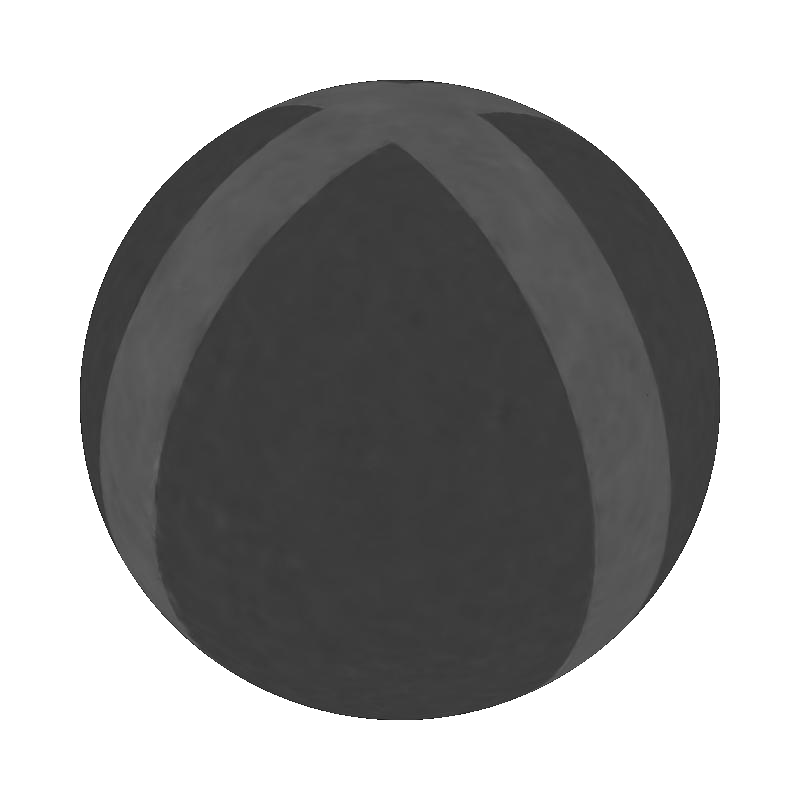} &
    \includegraphics[width=1in]{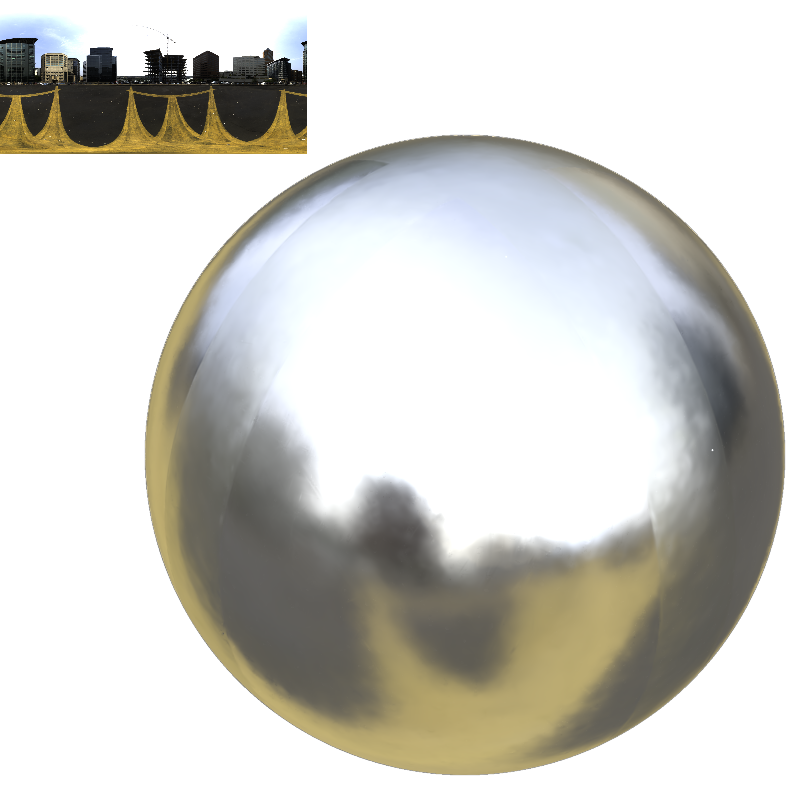}
    \\
    
    \includegraphics[width=1in]{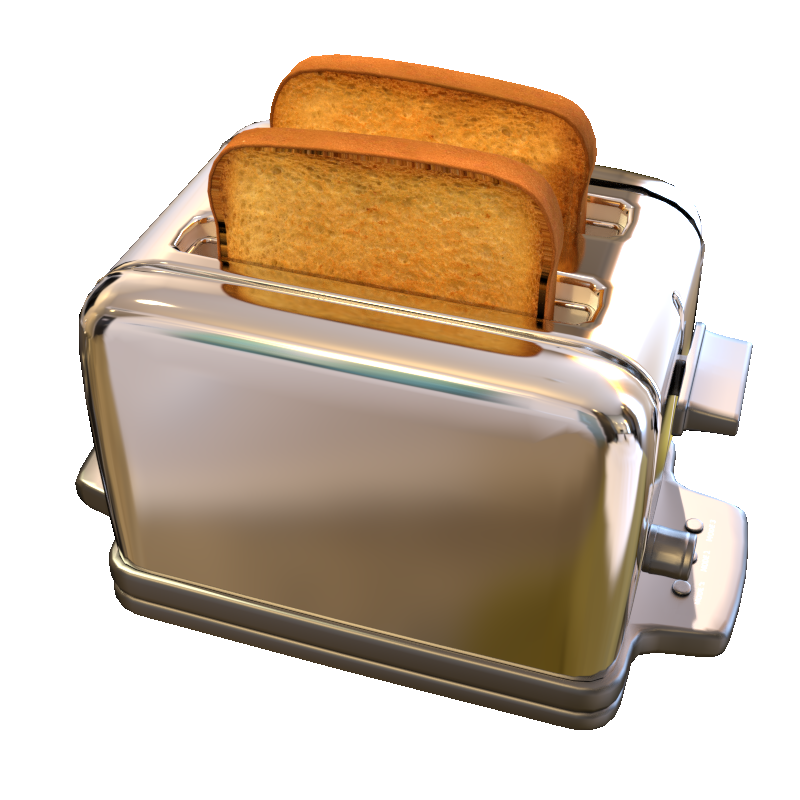} &
    \includegraphics[width=1in]{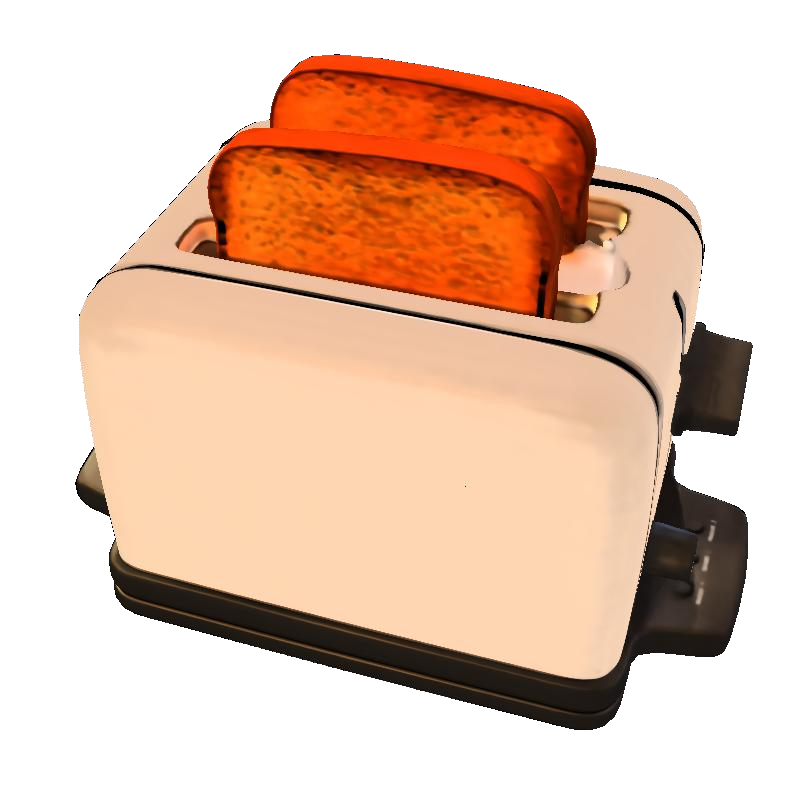} &
    \includegraphics[width=1in]{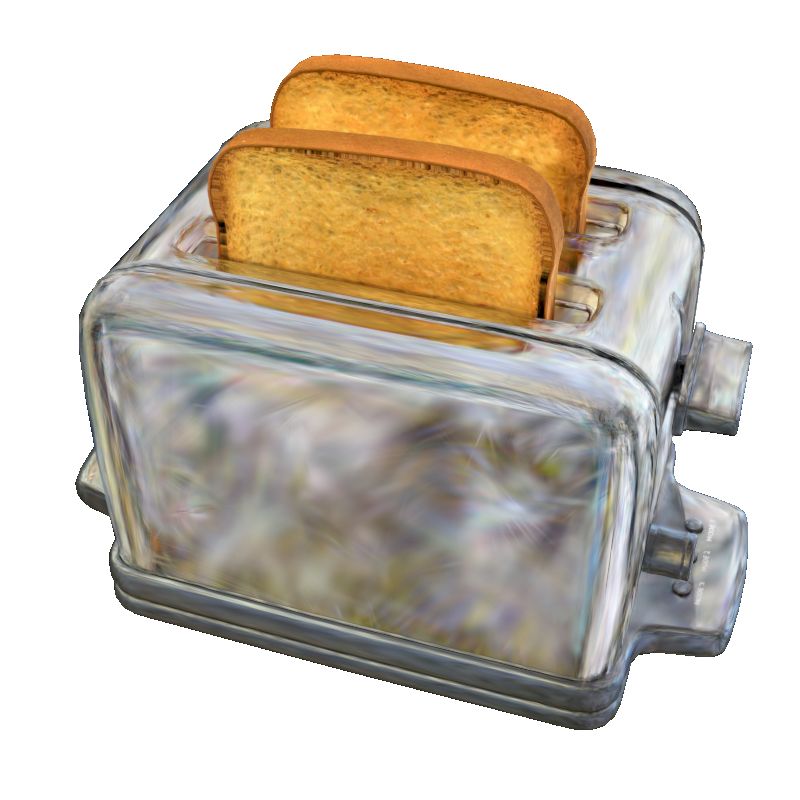} &
    \includegraphics[width=1in]{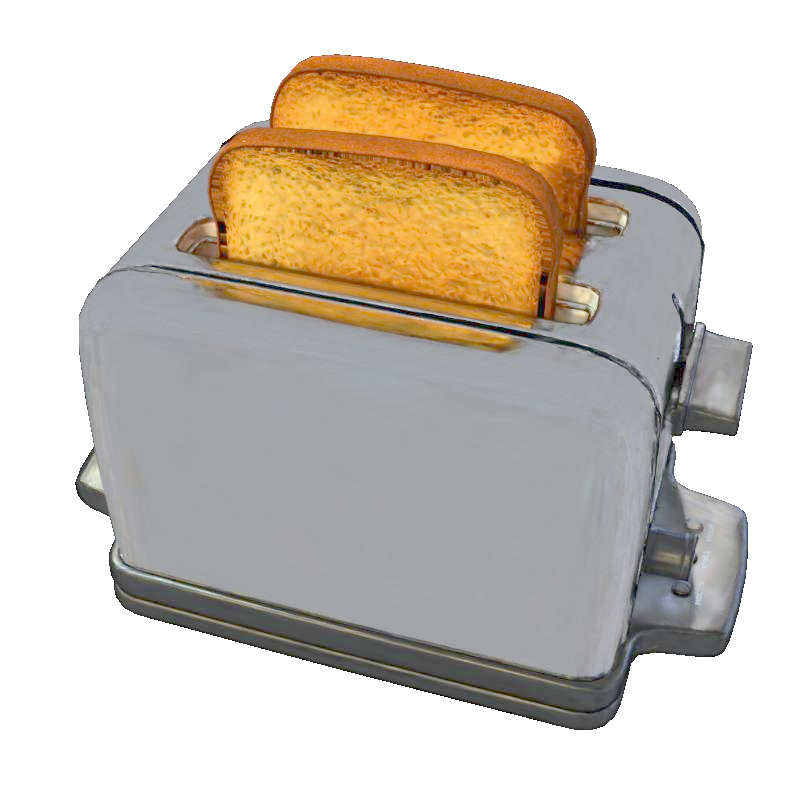} &
    \includegraphics[width=1in]{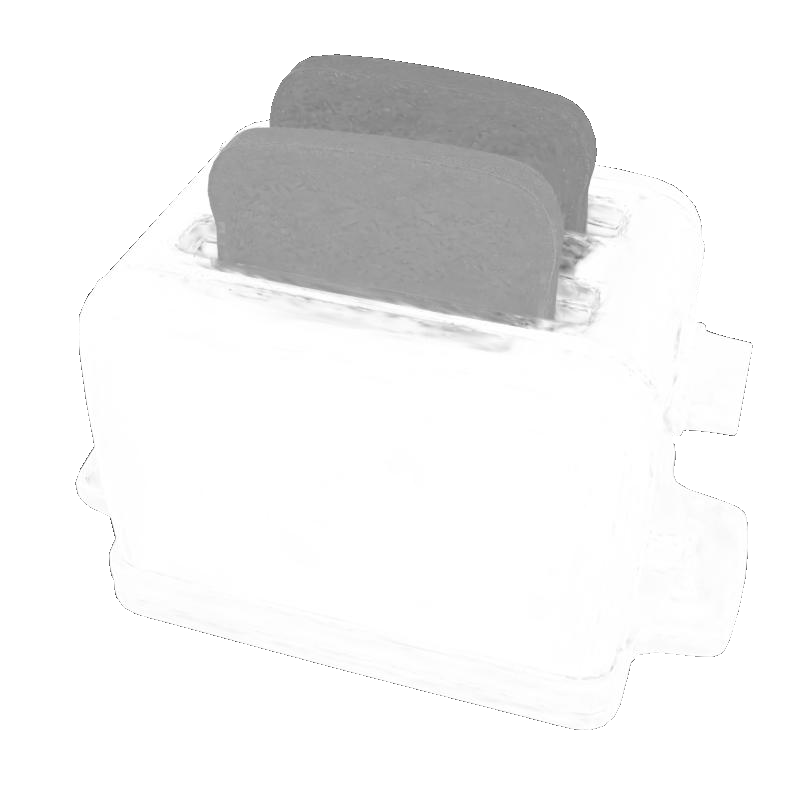} &
    \includegraphics[width=1in]{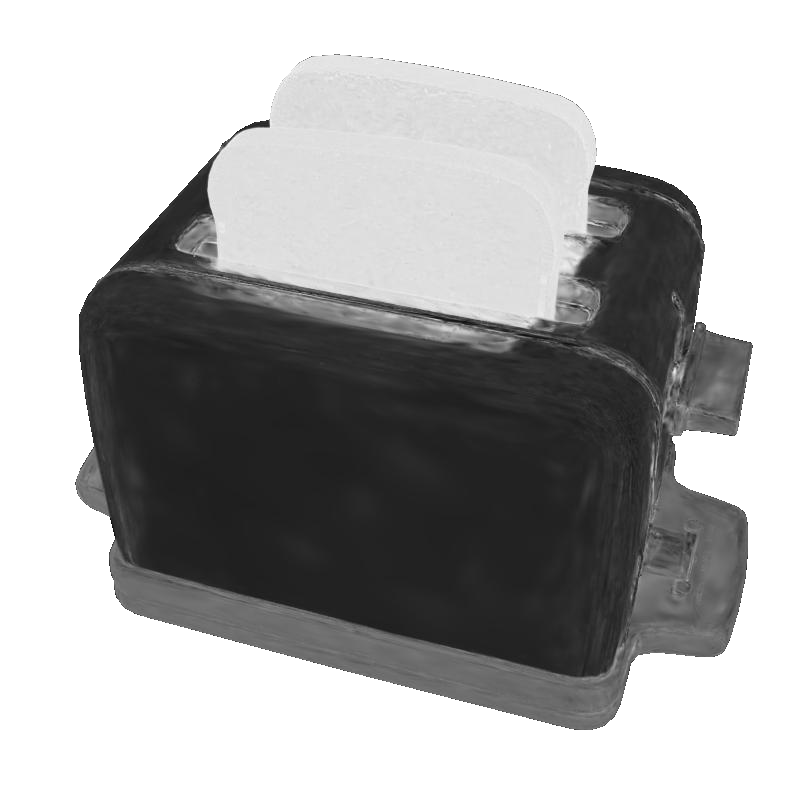} &
    \includegraphics[width=1in]{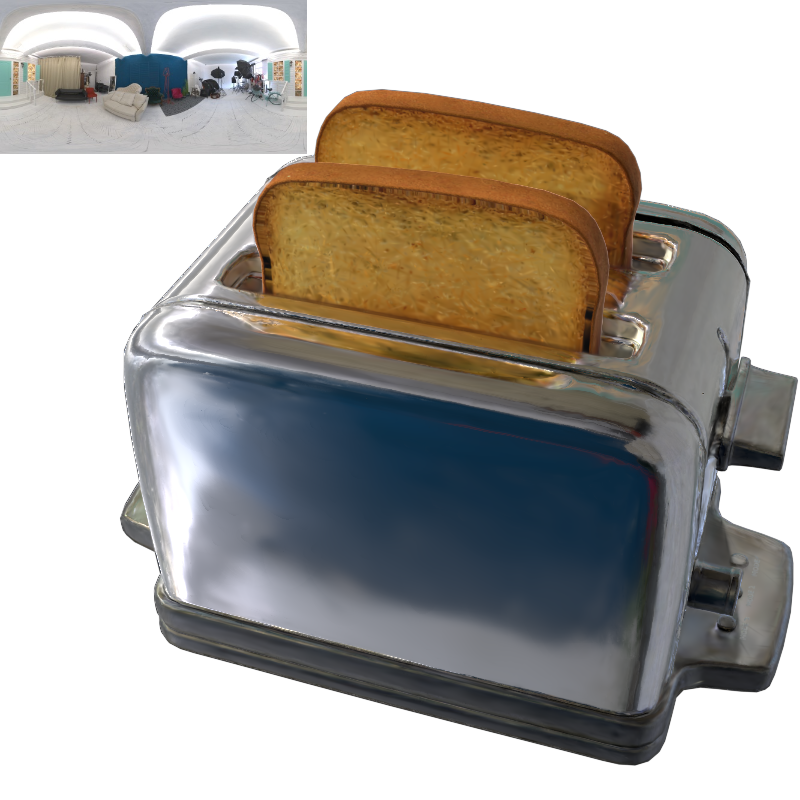}\\
    
    
    \includegraphics[width=1in]{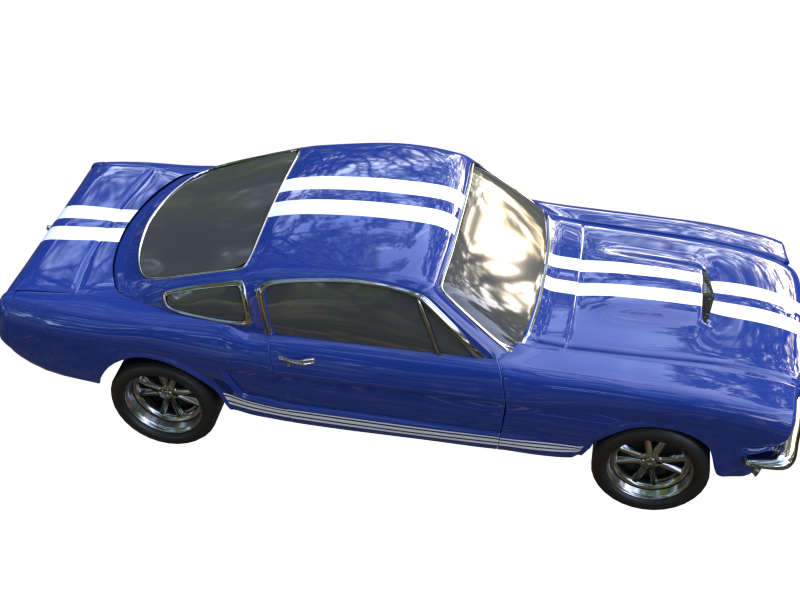} &
    \includegraphics[width=1in]{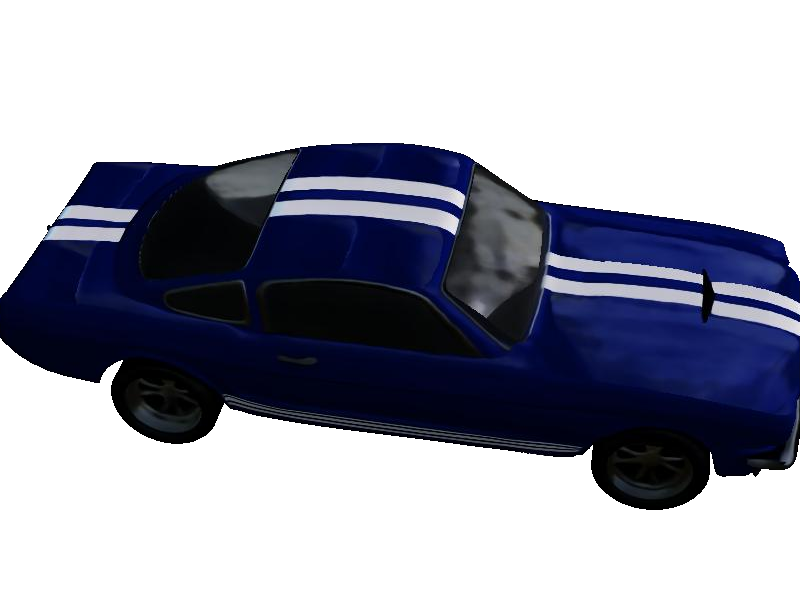} &
    \includegraphics[width=1in]{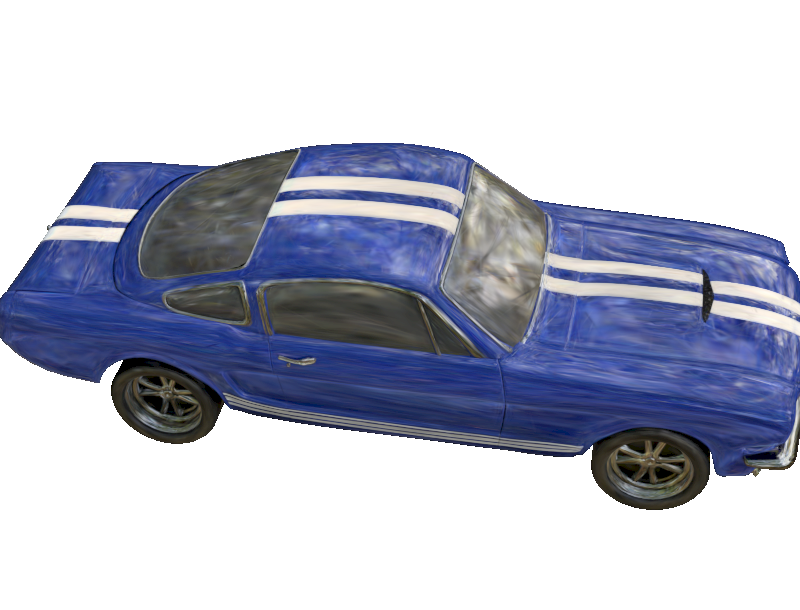} &
    \includegraphics[width=1in]{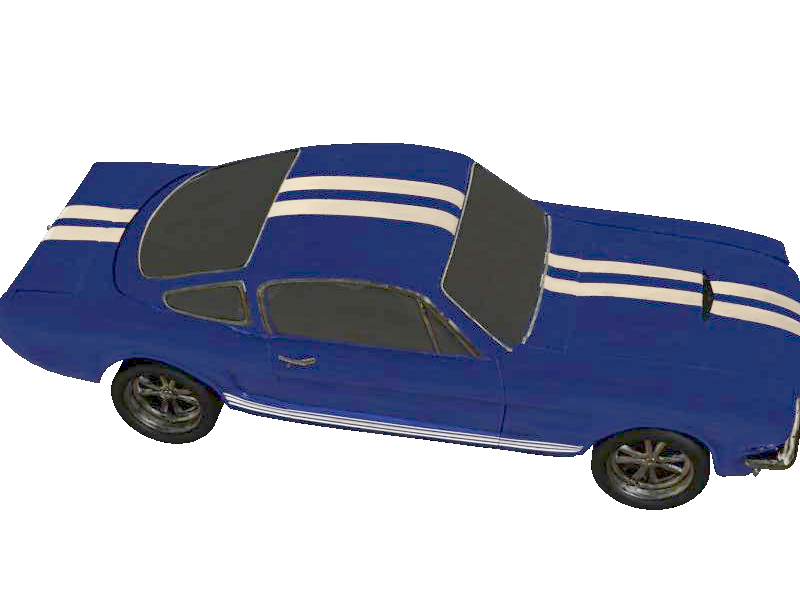} &
    \includegraphics[width=1in]{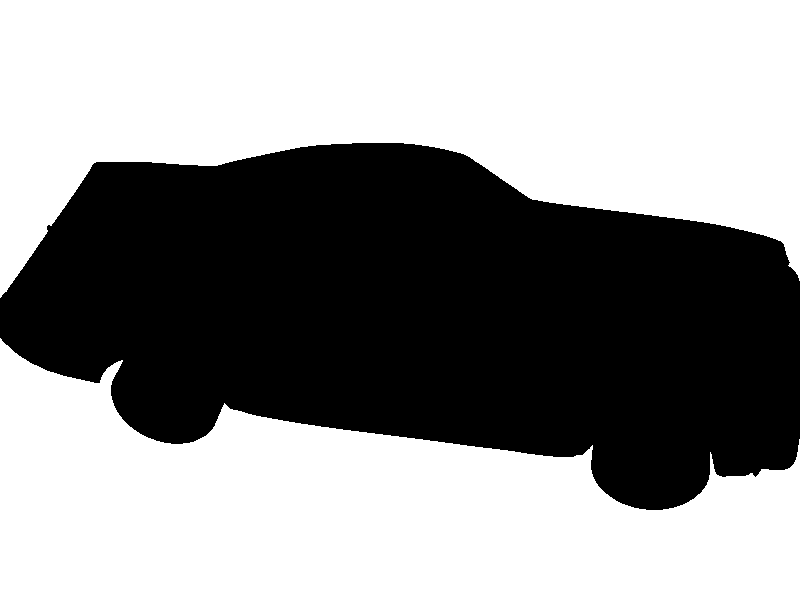} &
    \includegraphics[width=1in]{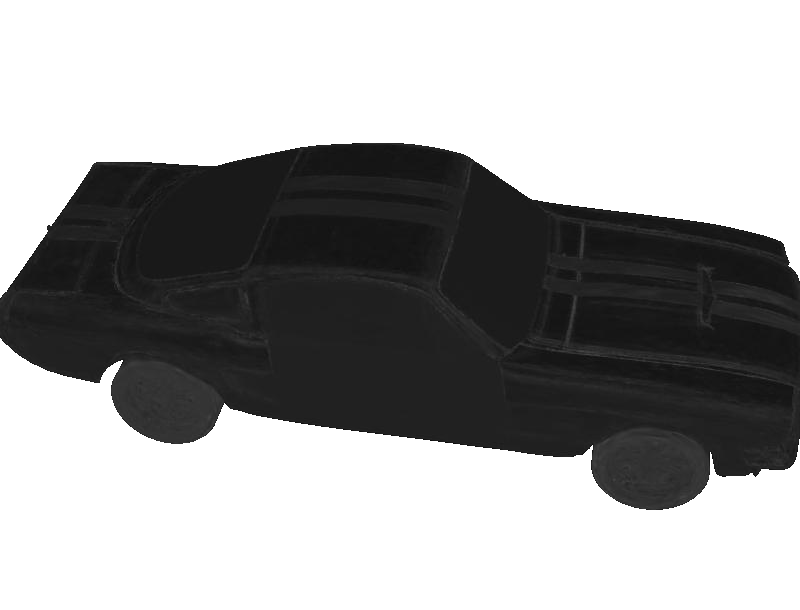}  &
    \includegraphics[width=1in]{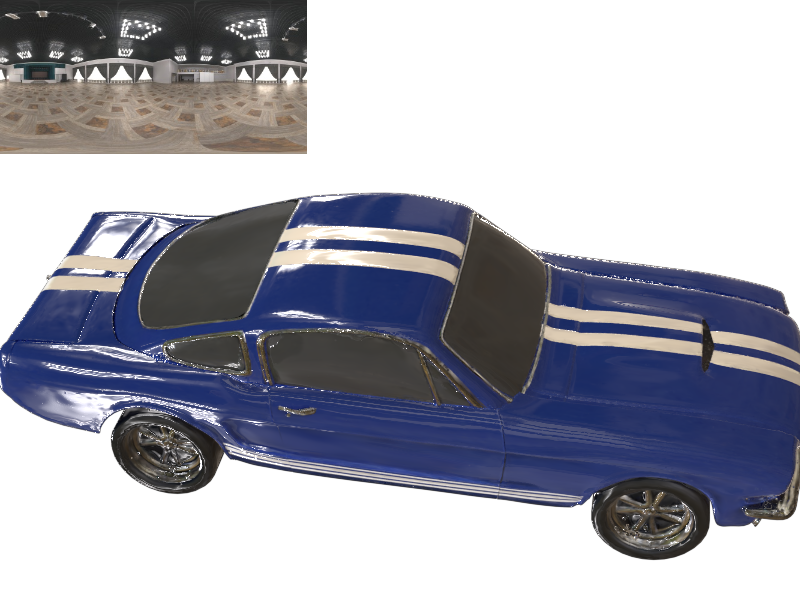} \\

    \end{tabular}
\end{center}
\caption{Comparison of the estimated BRDFs among our method, NeRO \cite{liu2023nero} and GSIR \cite{liang2023gs} on the Shiny Blender dataset \cite{verbin2022ref}.}
\label{fig:material}
\end{figure*}

\begin{figure*}[p]
\begin{center}
    \addtolength{\tabcolsep}{-4pt}
    \begin{tabular}{ccccc}
    
    Scene & Gshader & GSIR &Ours & Reference \\
    \includegraphics[height=0.7in]{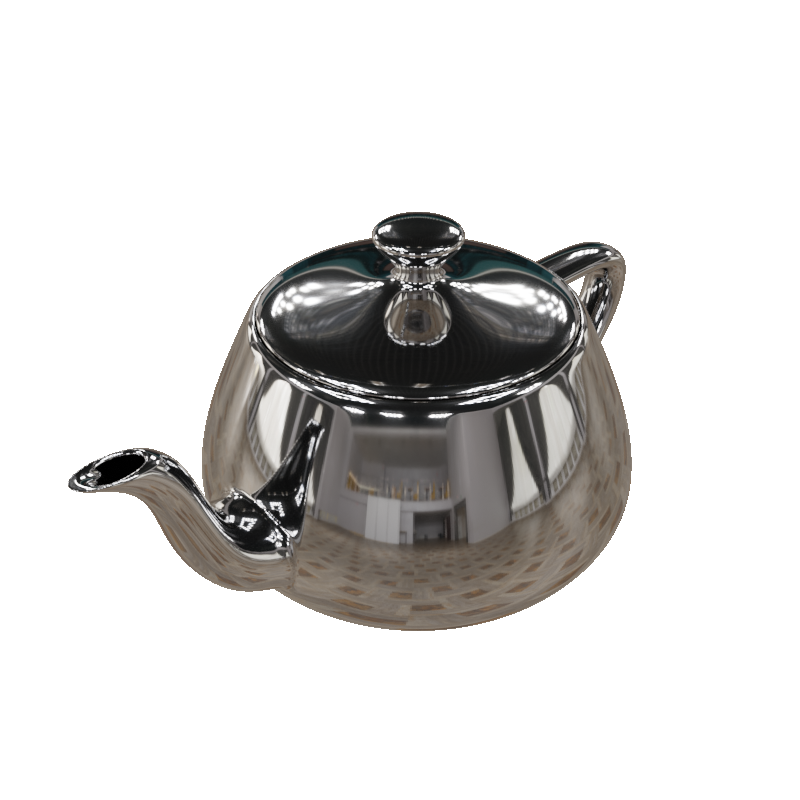} &
    \includegraphics[height=0.7in]{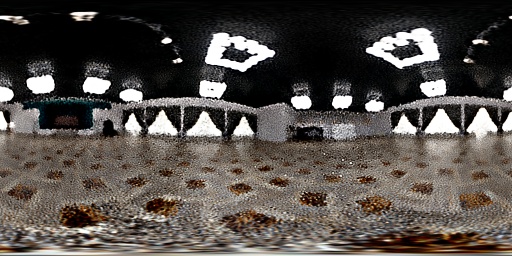} &
    \includegraphics[height=0.7in]{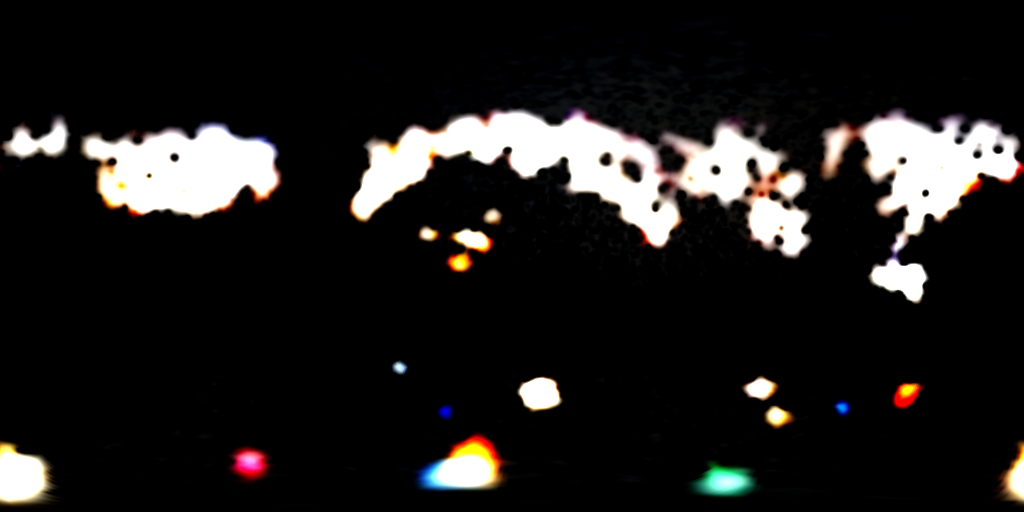} &
    \includegraphics[height=0.7in]{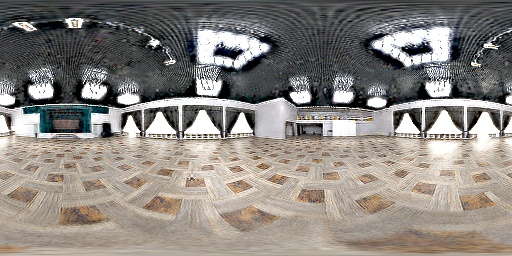}&
    \includegraphics[height=0.7in]{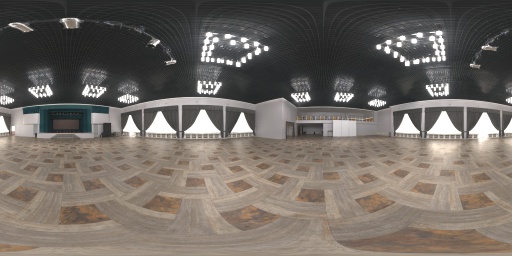}\\
    
    \includegraphics[height=0.7in]{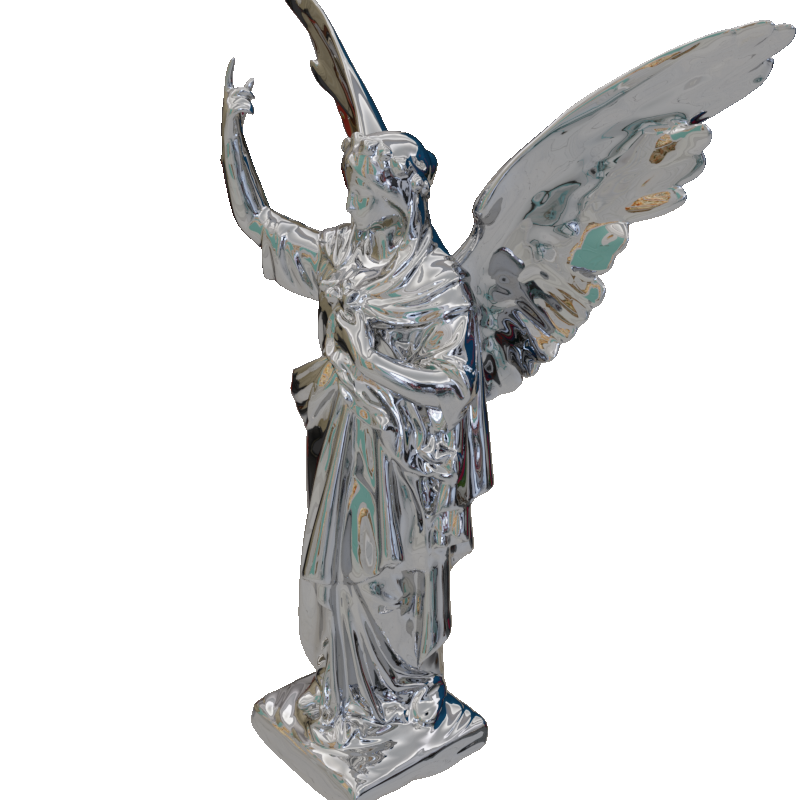}  &
    \includegraphics[height=0.7in]{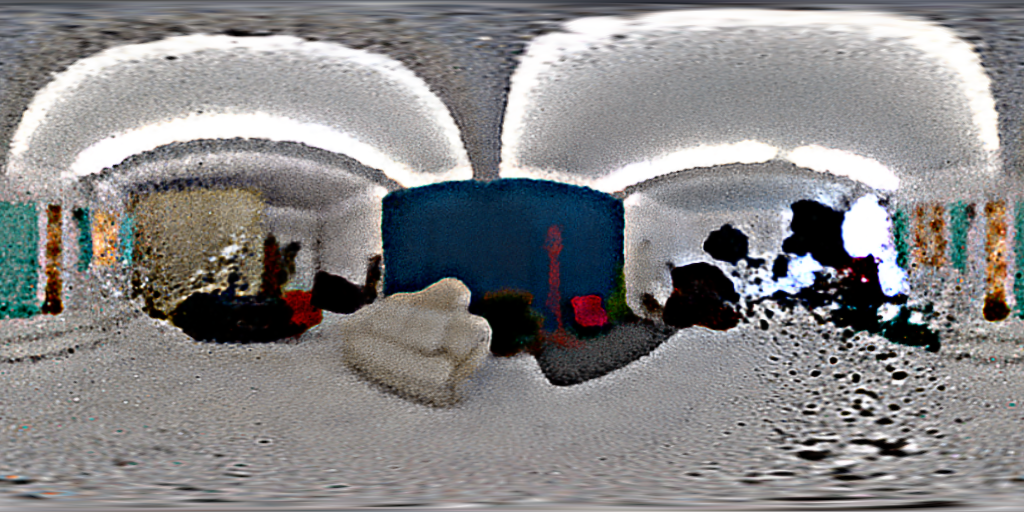} &
    \includegraphics[height=0.7in]{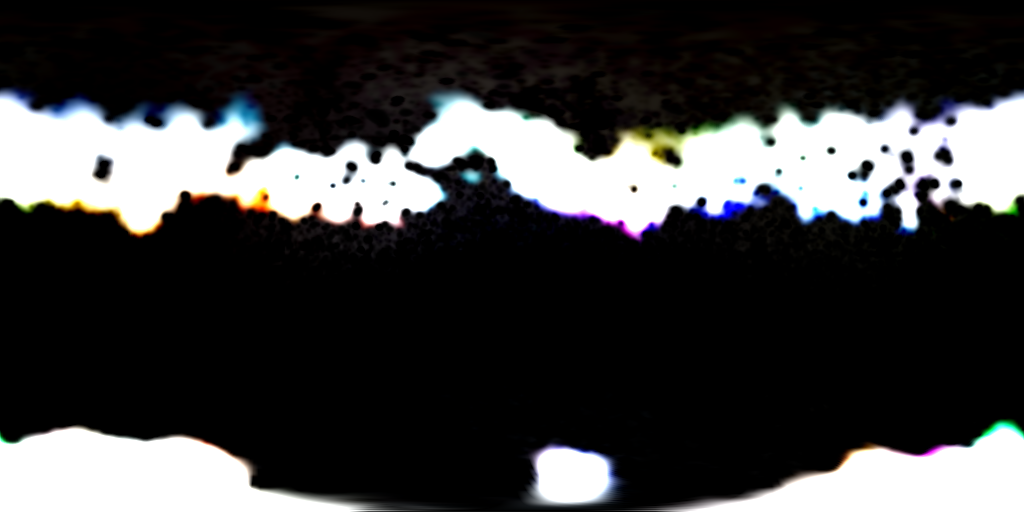} &
    \includegraphics[height=0.7in]{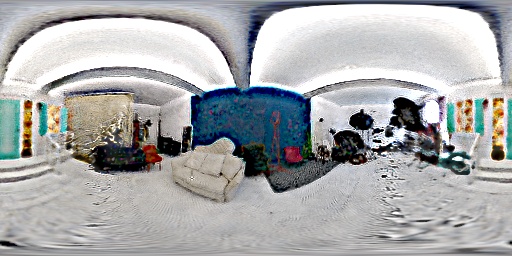}&
    \includegraphics[height=0.7in]{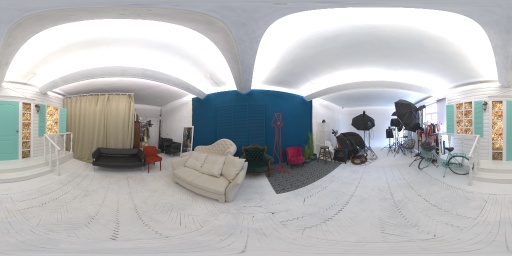}\\

    \end{tabular}
\end{center}
\caption{Comparison of environment maps among our method, Gshader \cite{jiang2023gaussianshader} and GSIR \cite{liang2023gs} on the Glossy Synthetic dataset \cite{liu2023nero}.}
\label{fig:envmap}
\end{figure*}

Visual comparisons of the normal reconstruction on synthetic datasets are illustrated in Fig. \ref{fig:normal_syn}, highlighting our method's capability in reconstructing surfaces exhibiting substantial reflection, alongside the recovery of detailed geometries (the nuanced folds of the clothing in the \textit{Angel} scene). NeRO \cite{liu2023nero} exhibits remarkable reconstruction ability when applied to simple specular objects, but when applied to more complex scenarios, it tends to converge on local minima, leading to inaccurate geometries (pointed by the red box in the \textit{Luyu} scene). NMF~\cite{mai2023neural} can roughly recover geometry, but it tends to lose fine details (the missing door handle in the \textit{Car} scene). Gshader \cite{jiang2023gaussianshader} often incorrectly overfits highlights by exploiting geometric indentations, due to its forward shading mode. GSIR \cite{liang2023gs} excels in reconstructing non-specular surfaces but face difficulties with surfaces showing strong reflections. 
We report the quantitative metrics on the Shiny Blender~\cite{verbin2022ref} dataset in Tab. \ref{tab:refnerf}. Our method demonstrates superior performance in relighting quality compared to baseline methods, highlighting the effectiveness of inverse rendering. Importantly, our method achieves the shortest training time, demonstrating a $4\times$ improvement over NeRF-based methods, which showcases its efficiency. Furthermore, we attain the highest relighting rendering frame rate, thanks to our hybrid geometry and material representation.

Additionally, we showcase the estimated BRDFs (albedo, roughness and metallic) in Fig. \ref{fig:material}. Thanks to the micro-facet geometry segmentation prior and the roughness constraint loss, the roughness maps tend to be smooth and physically-plausible, which reduces the ambiguities in inverse rendering and enhances the decoupling of albedo and metallic. This contrasts with other methods that inaccurately assign lighting effects to albedo maps. The reconstructed environment maps, as illustrated in Fig. \ref{fig:envmap}, further support our method's ability to effectively decouple the BRDFs, allowing for a nearly complete restoration of the environment map. In contrast, GShader can only generate noisy images with many holes, while GSIR struggles with poor environmental maps due to its failure to decouple light from the albedo.

\begin{figure*}[htb]
\begin{center}
    \addtolength{\tabcolsep}{-4pt}
    \begin{tabular}{m{2cm}<{\centering} m{2.5cm}<{\centering} m{2.5cm}<{\centering} m{2.5cm}<{\centering} m{2.5cm}<{\centering} m{2.5cm}<{\centering} m{2.5cm}<{\centering}}
    
    New lighting & NeRO & NMF & Gshader & GSIR &Ours  & GT\\
    \centering
    \includegraphics[width=0.6in]{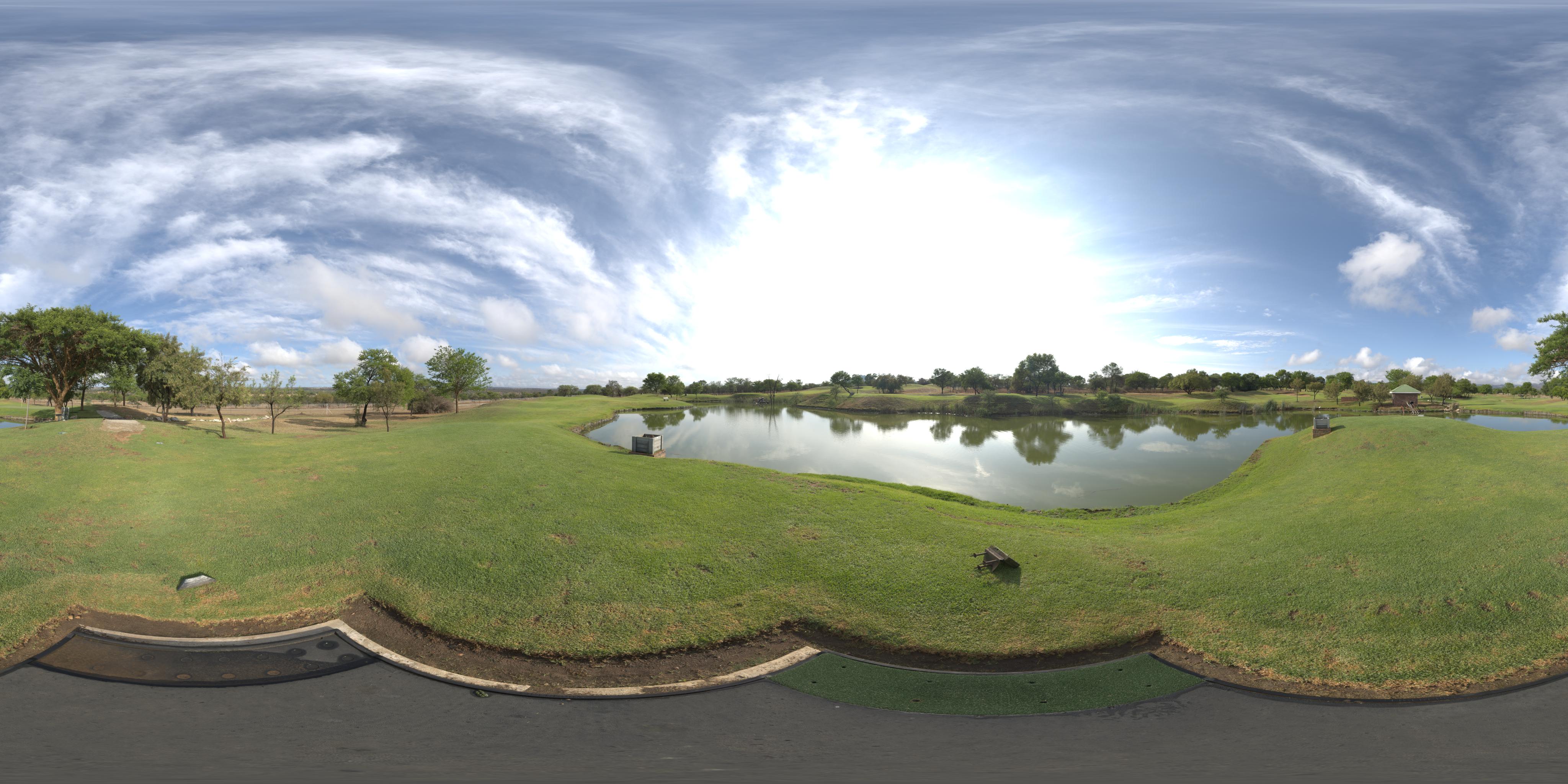} &
    \includegraphics[width=1.1in]{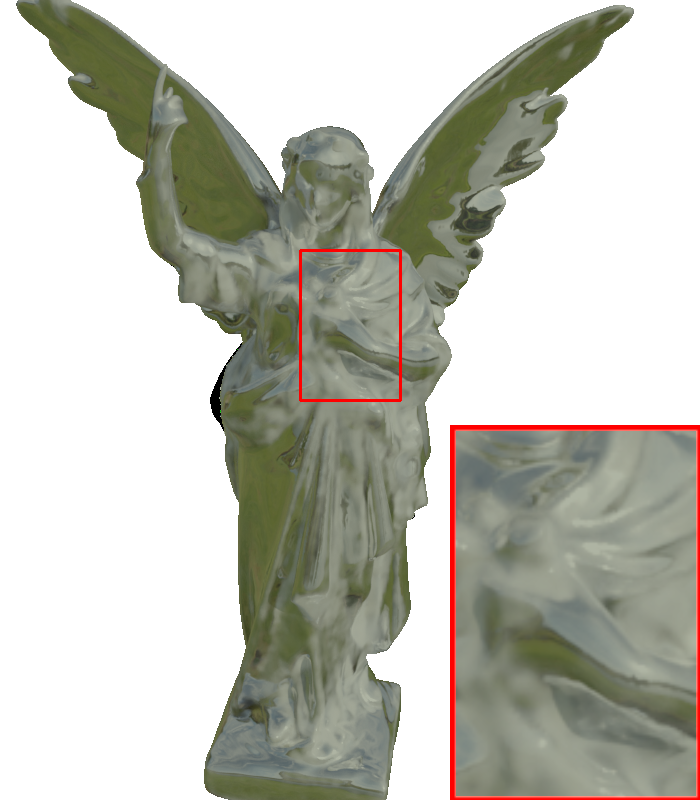} &
    \includegraphics[width=1.1in]{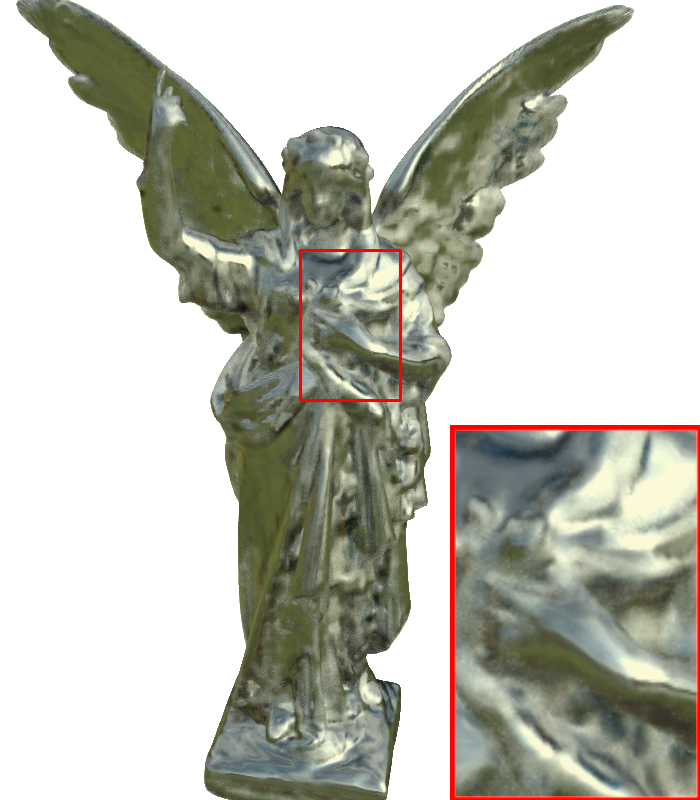} &
    \includegraphics[width=1.1in]{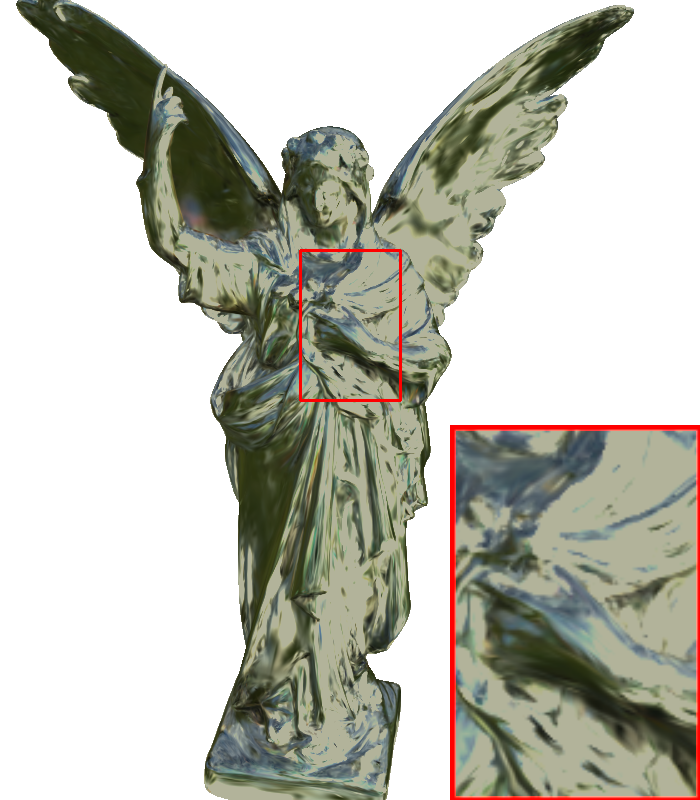} &
    \includegraphics[width=1.1in]{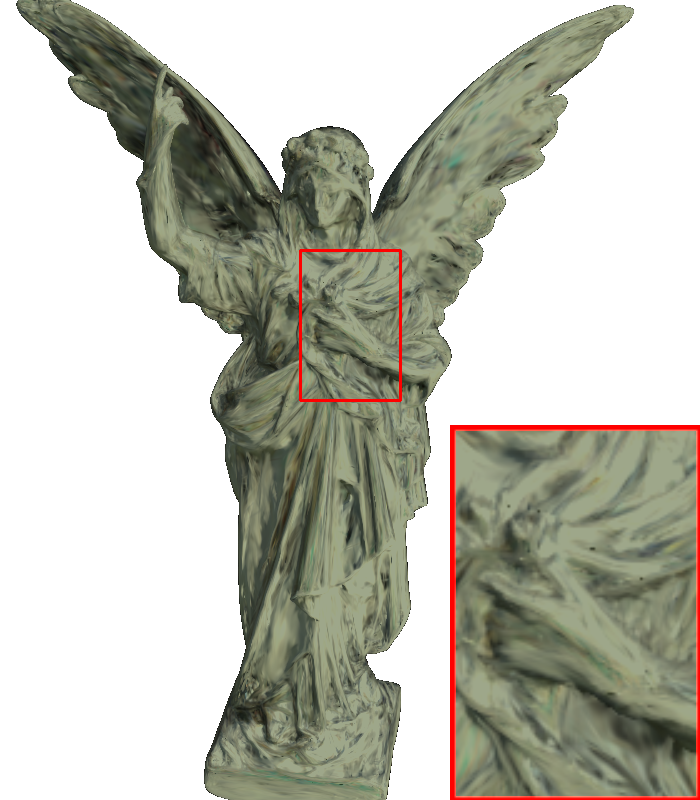} &
    \includegraphics[width=1.1in]{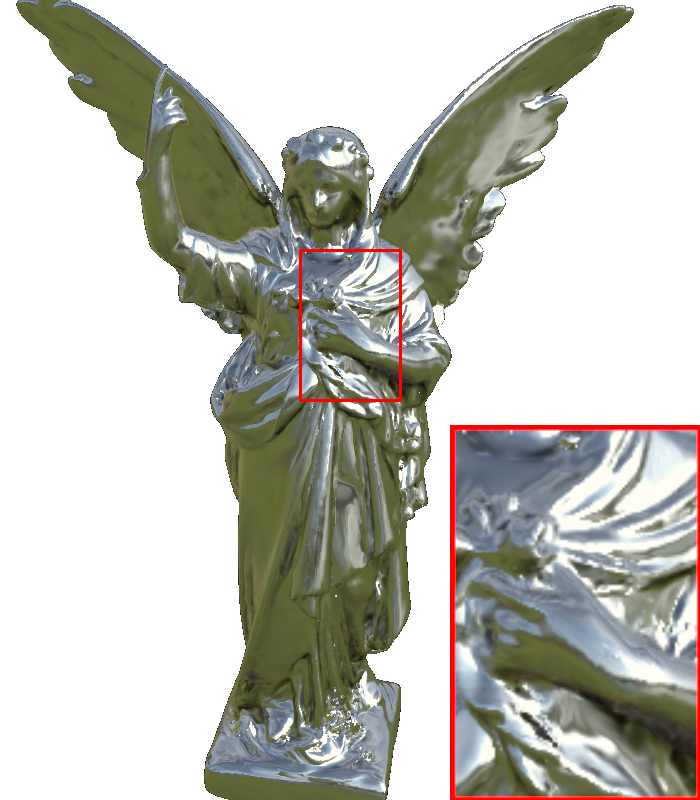} &
    \includegraphics[width=1.1in]{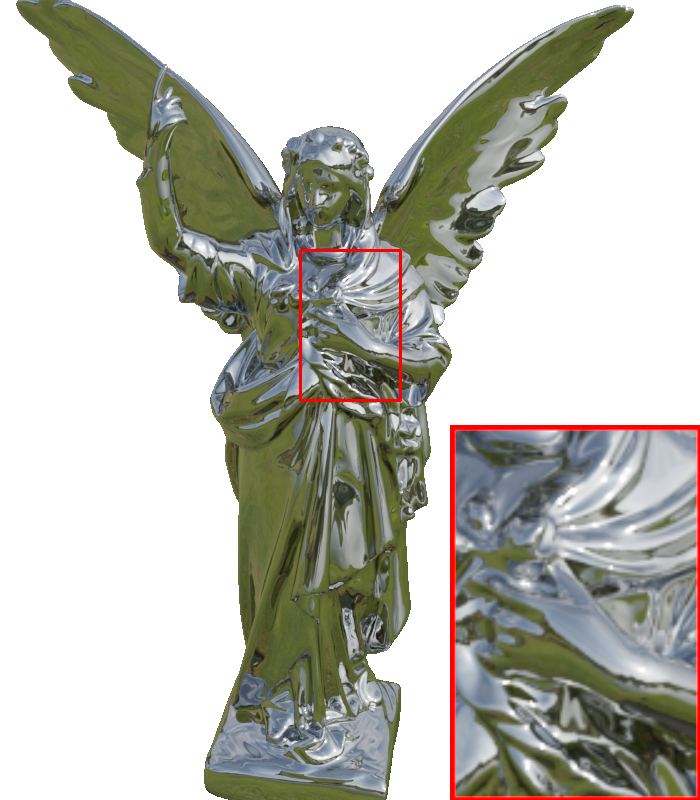}\\
    & Angel& & & & &  \\

    \includegraphics[width=0.6in]{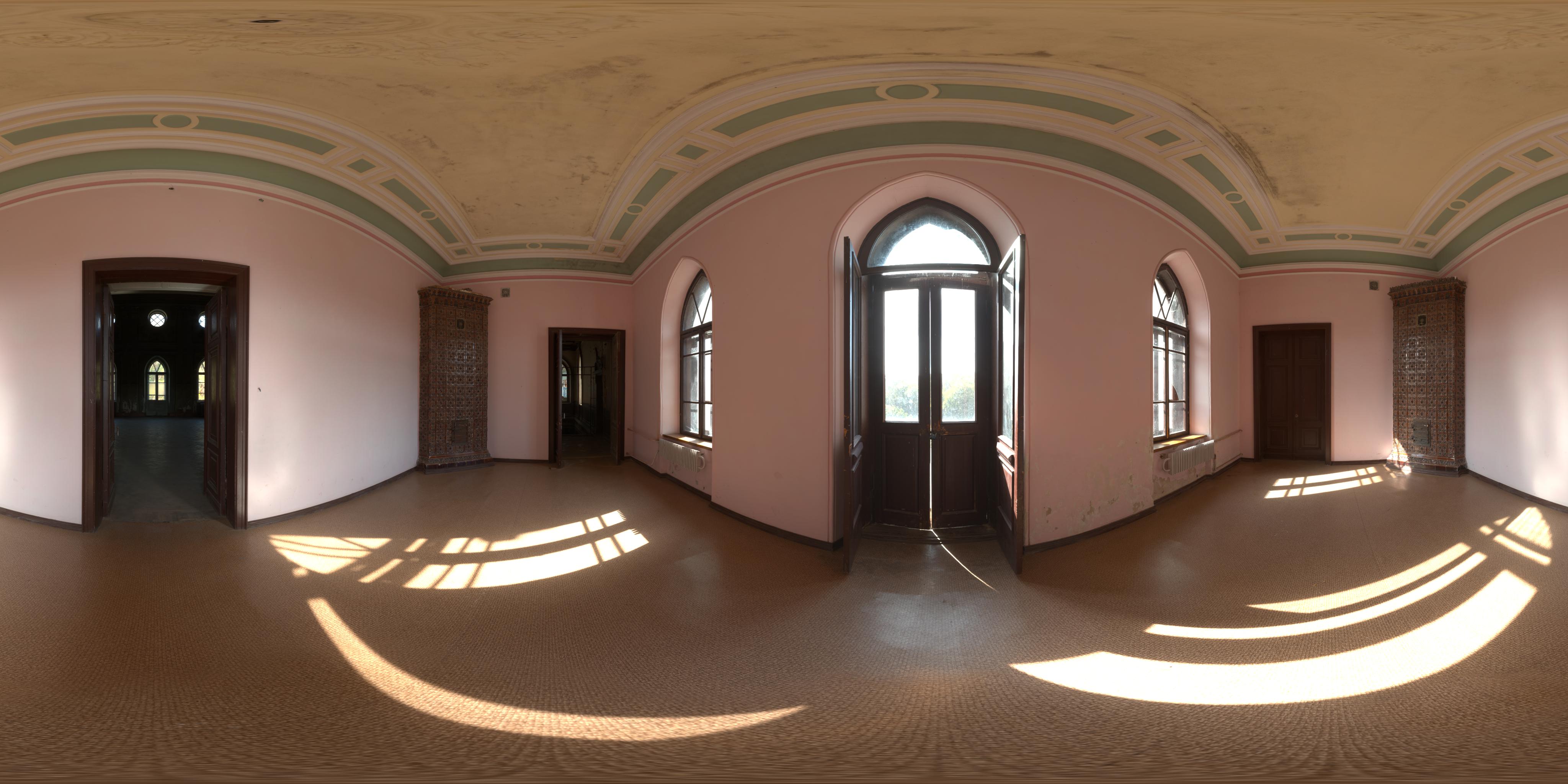} &
    \includegraphics[width=1.1in]{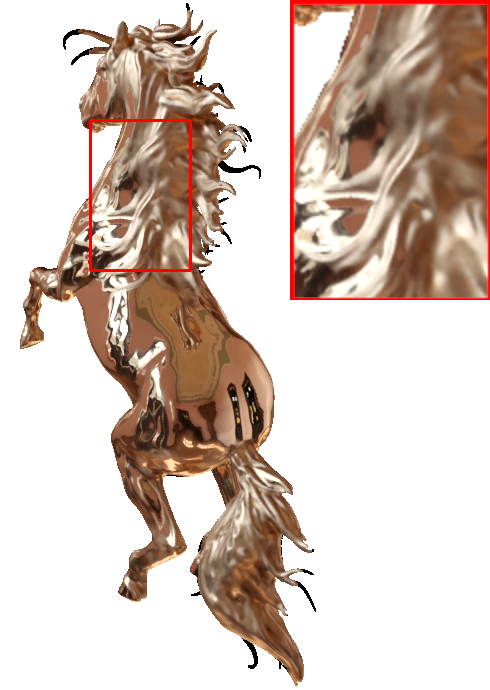} &
    \includegraphics[width=1.1in]{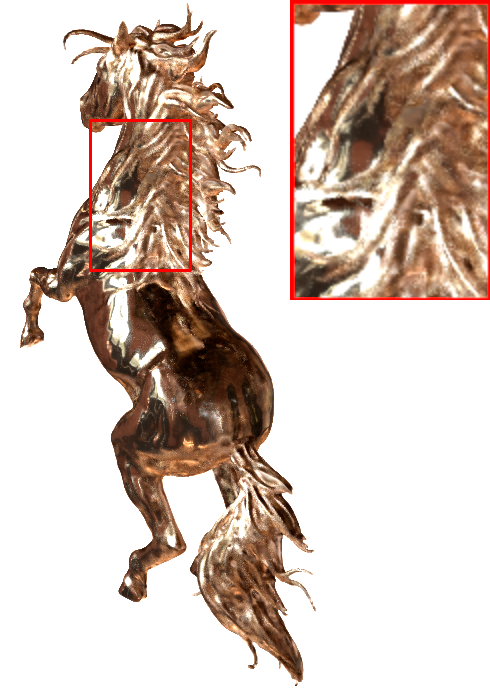} &
    \includegraphics[width=1.1in]{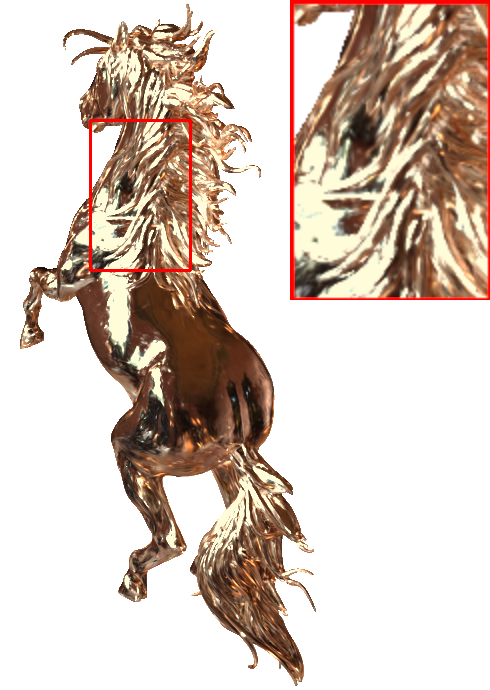} &
    \includegraphics[width=1.1in]{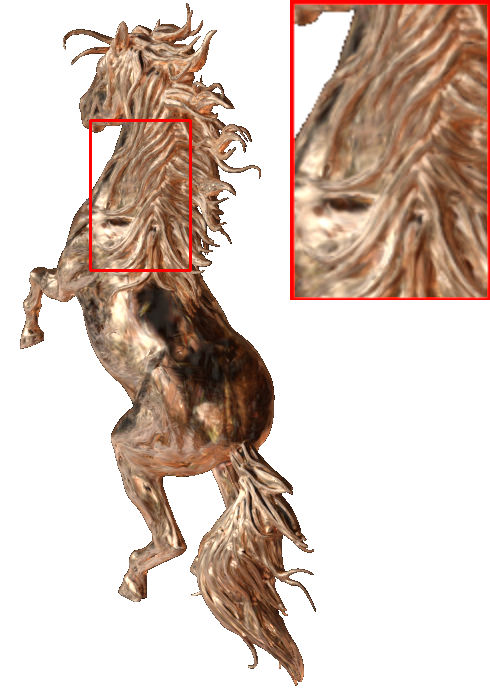} &
    \includegraphics[width=1.1in]{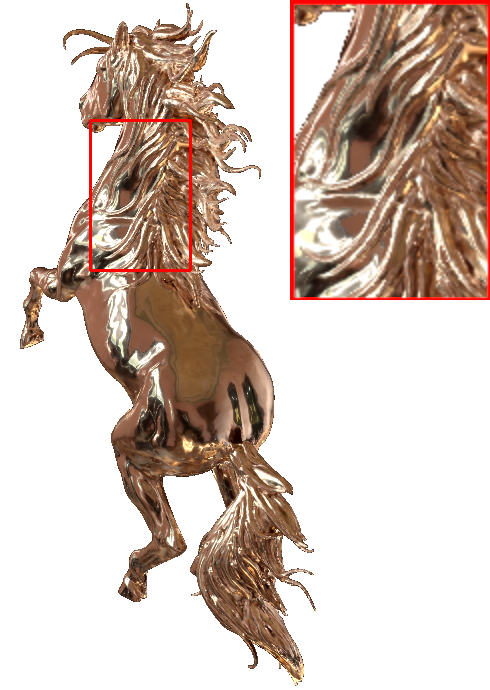} &
    \includegraphics[width=1.1in]{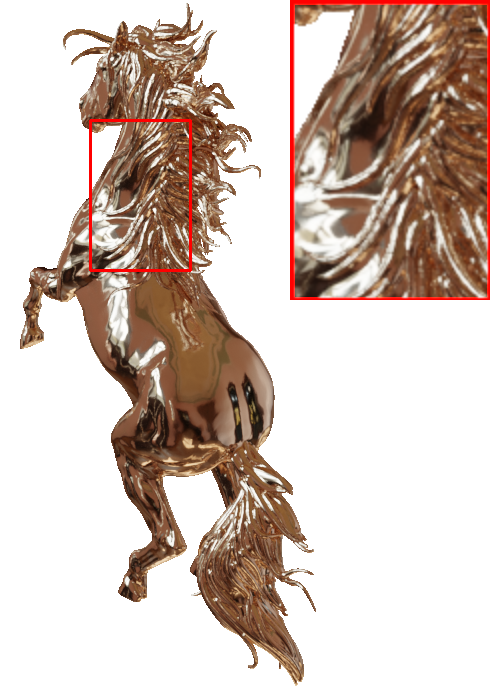} \\
    &Horse & & & & &  \\

    \includegraphics[width=0.6in]{figures/experiment/relight/golf.jpg} &
    \includegraphics[width=1.1in]{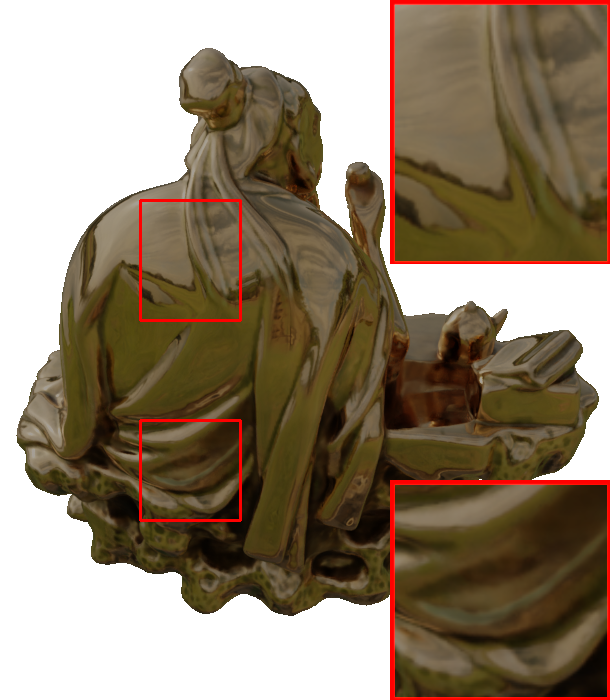} &
    \includegraphics[width=1.1in]{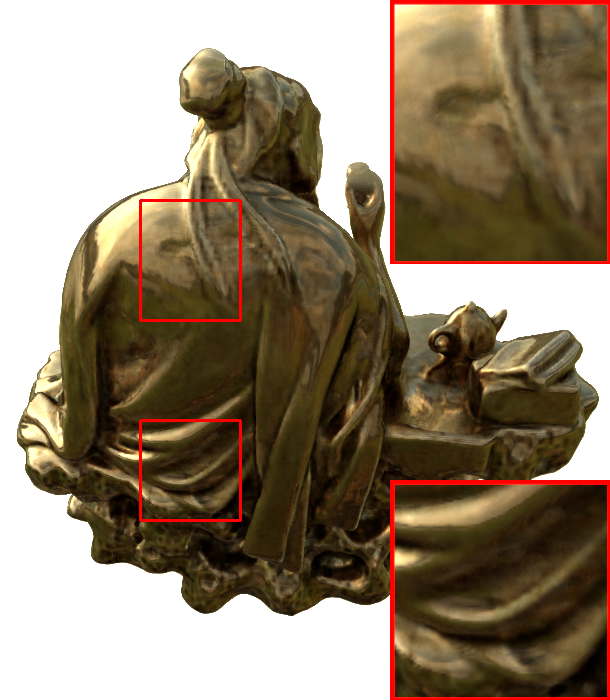} &
    \includegraphics[width=1.1in]{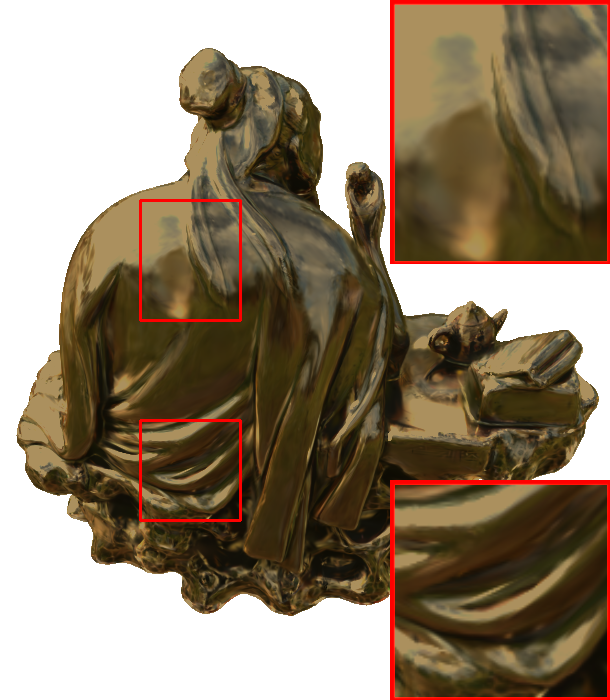} &
    \includegraphics[width=1.1in]{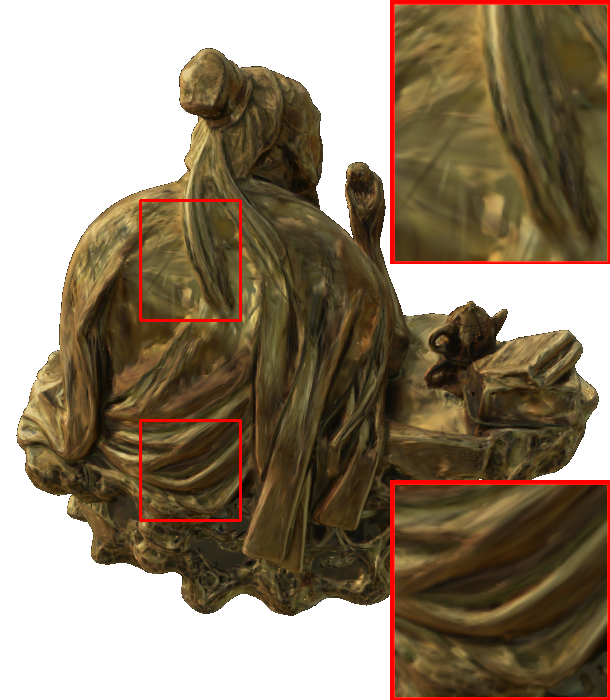} &
    \includegraphics[width=1.1in]{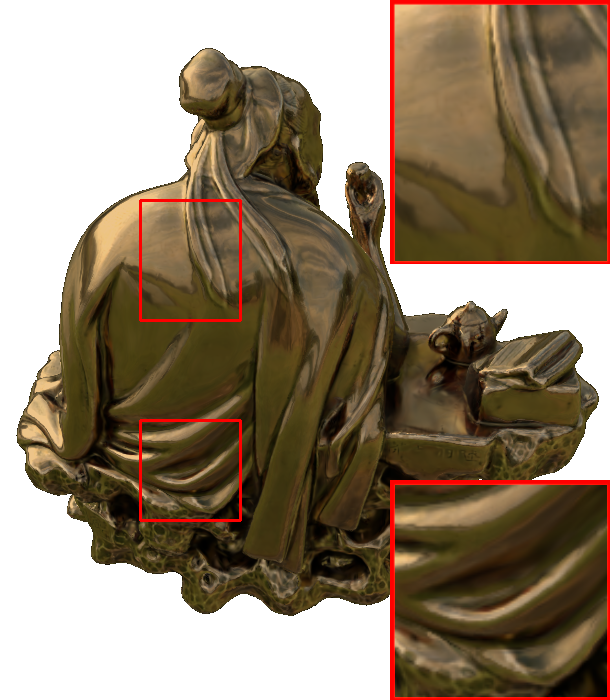} &
    \includegraphics[width=1.1in]{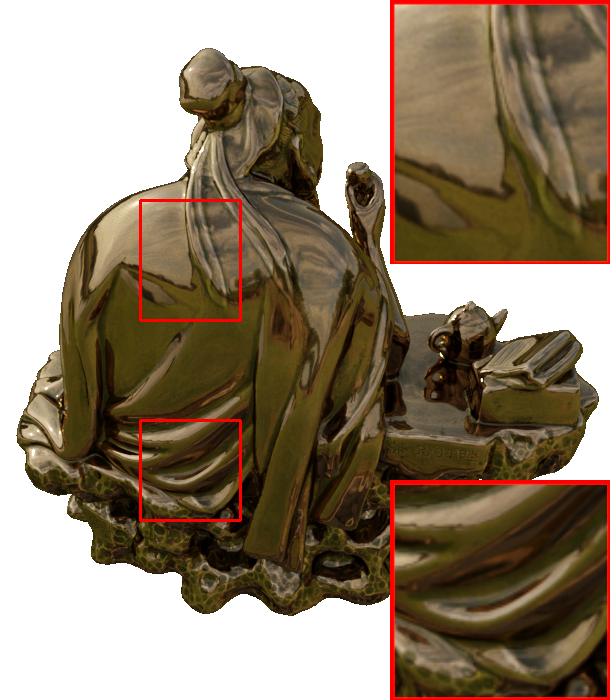} \\
    
    & Luyu& & & & &  \\

    \end{tabular}
\end{center}
\caption{Comparison of relighting results among our method, NMF~\cite{mai2023neural}, NeRO \cite{liu2023nero}, Gshader \cite{jiang2023gaussianshader} and GSIR \cite{liang2023gs} on the Glossy Synthetic dataset \cite{liu2023nero}.}
\label{fig:relight_nero}
\end{figure*}

\begin{figure*}[htb]
\begin{center}
    \addtolength{\tabcolsep}{-4pt}
    \begin{tabular}{m{2.916cm}<{\centering} m{2.916cm}<{\centering} m{2.916cm}<{\centering} m{2.916cm}<{\centering} m{2.916cm}<{\centering} m{2.916cm}<{\centering}}
    
     NeRO & NMF & Gshader & GSIR & Ours &GT \\
    \includegraphics[height=0.8in]{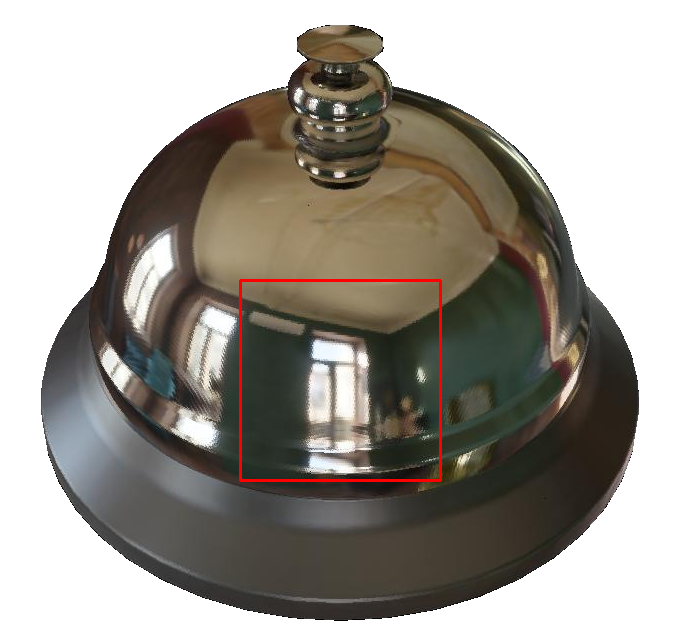} &
    \includegraphics[height=0.8in]{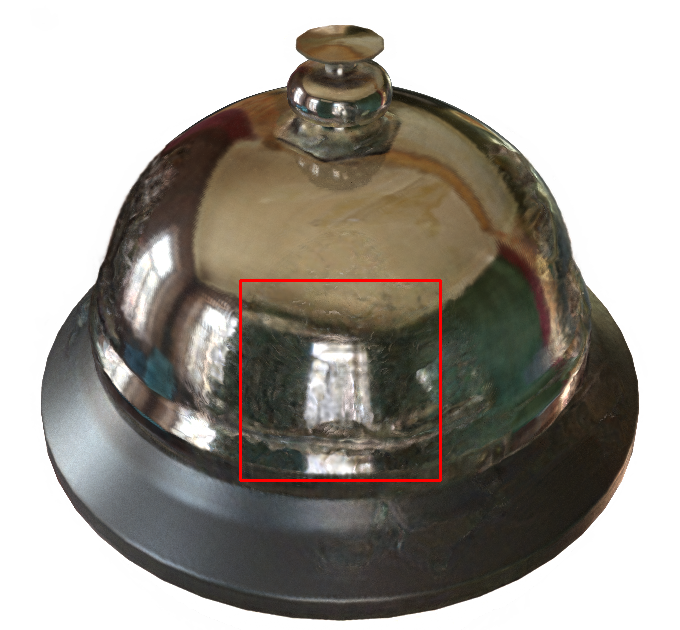} &
    \includegraphics[height=0.8in]{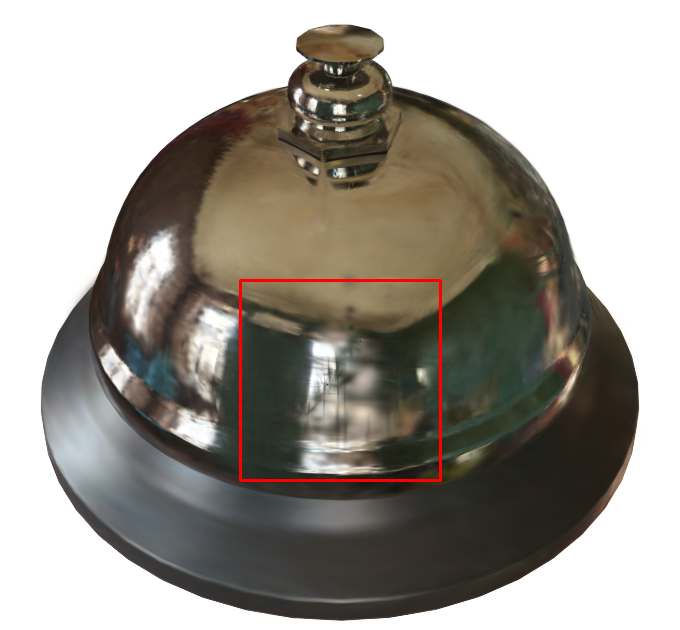} &
    \includegraphics[height=0.8in]{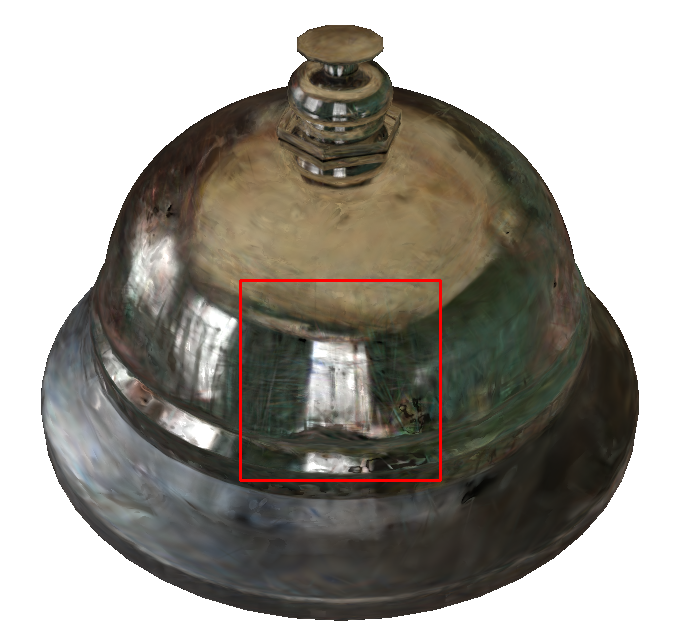} &
    \includegraphics[height=0.8in]{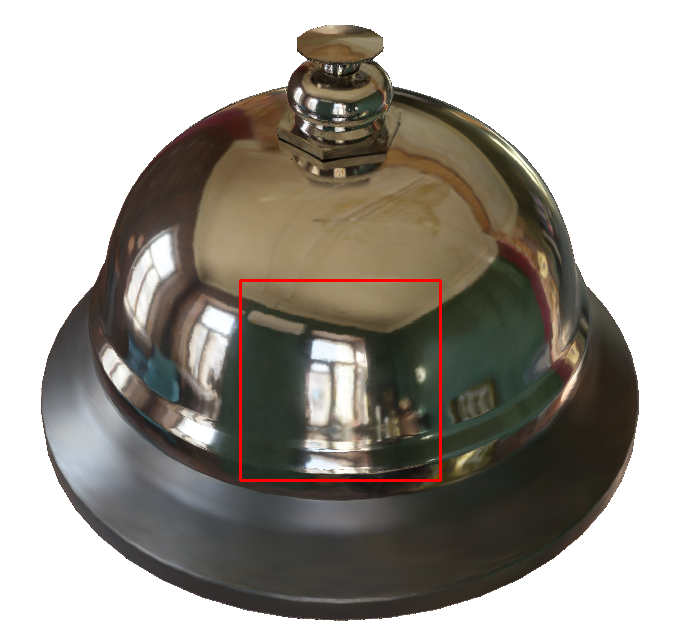}&
    \includegraphics[height=0.8in]{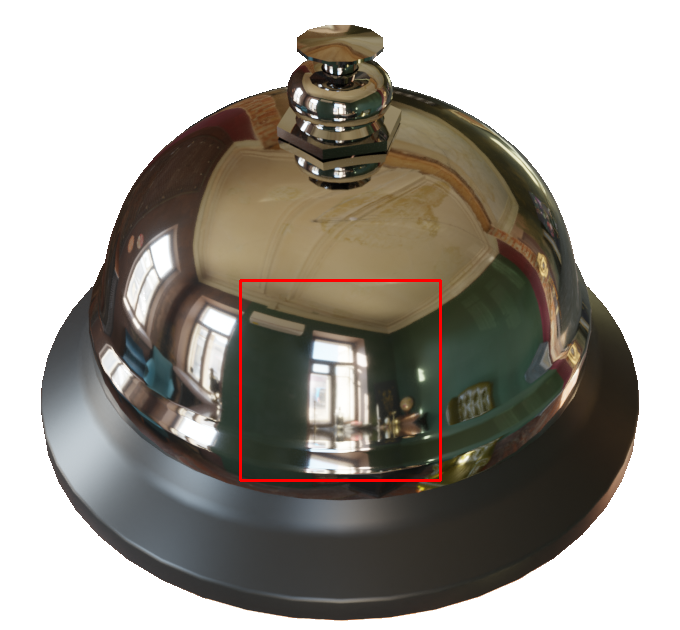} \\

    \includegraphics[width=1.2in]{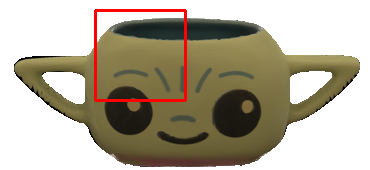} &
    \includegraphics[width=1.2in]{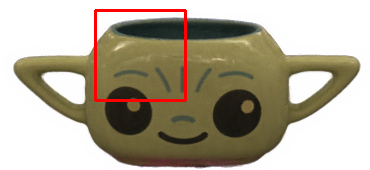} &
    \includegraphics[width=1.2in]{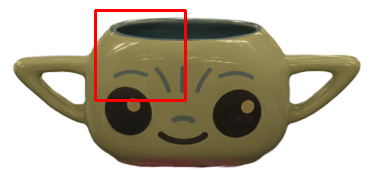} &
    \includegraphics[width=1.2in]{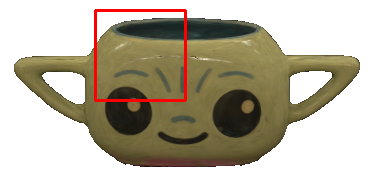} &
    \includegraphics[width=1.2in]{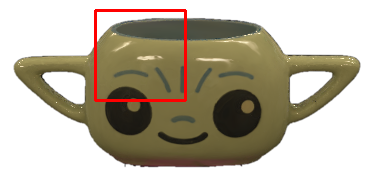} &
    \includegraphics[width=1.2in]{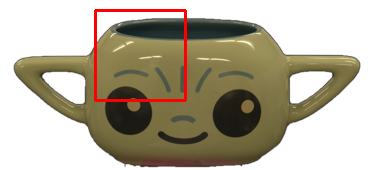}\\
    \end{tabular}
\end{center}
\caption{Comparison of novel view synthesis among our method, NMF~\cite{mai2023neural}, NeRO \cite{liu2023nero}, Gshader \cite{jiang2023gaussianshader} and GSIR \cite{liang2023gs} on the Glossy Synthetic \cite{liu2023nero} (the first row), and Stanford-ORB dataset \cite{kuang2024stanford}(the second row).}
\label{fig:nvs}
\end{figure*}

Our method excels in novel view synthesis and relighting, outperforming others, as evidenced in Tab.~\ref{tab:refnerf} and Tab.~\ref{tab:nero}. Fig.~\ref{fig:nvs} highlights our success in capturing specular reflections in novel view synthesis. Furthermore, as shown in Fig.~\ref{fig:relight_nero}, our relighting results produce realistic visualizations and effectively recover specular highlights. NeRO mainly suffers from a loss of geometric details, leading to blurred highlight details (pointed by the red box in the \textit{Horse} scene). NMF and GShader both encounter issues with geometries and materials, resulting in inferior relighting quality. Meanwhile, GSIR produces diffuse materials and struggles to model high-frequency effects.


\subsection{Results on real data}

We first evaluate our method on the NeILF-HDR dataset \cite{zhang2023neilf++} and several real scenes captured by ourself, as shown in Fig. \ref{fig:neilf}. We compare our method with the other four methods on normal reconstruction. As seen, our approach effectively handles surfaces with strong highlights and recovers geometric details. 
 In contrast, GShader \cite{jiang2023gaussianshader} demonstrates unstable geometry recovery, resulting in some voids, as shown in the \textit{Qilin} and \textit{Brassgourd} scene. GS-IR \cite{liang2023gs} struggles to generate smooth geometric surfaces. NeRF-based methods (NeRO \cite{liu2023nero} and NMF \cite{mai2023neural}) can only generate coarse geometries without details (pointed by the red box in the \textit{GoldenDragon} scene). The relighting results (as shown in the seventh column of Fig. \ref{fig:neilf}) further demonstrate the reliability of our material reconstruction, as the specular highlights change appropriately with variations in lighting conditions.
\begin{table}[]
    \centering
  \small
  \setlength\tabcolsep{2pt}
\caption{\label{tab:stanford}Quantitative comparison of normal reconstruction and relighting results on the Stanford-ORB \cite{kuang2024stanford} dataset. \textbf{Bold} means the best performance.}
\begin{tabular}{c|ccccc}
\hline
           & NeRO  & NMF & Gshader & GSIR  & Ours           \\
           \hline
PSNR$\uparrow$       & 27.35 & 25.64 & 25.76   & 28.04 & \textbf{30.3}  \\
SSIM$\uparrow$       & 0.969 & 0.944 & 0.957   & 0.962 & \textbf{0.975} \\
LPIPS$\downarrow$      & 0.051 & 0.048 & 0.041   & 0.033 & \textbf{0.022} \\
normal MAE$\downarrow$ & 1.84 & 1.81  & 2.01    & 2.69  & \textbf{1.75} \\
           \hline
\end{tabular}
\end{table}
 
\begin{figure*}[htb]
\begin{center}
    \addtolength{\tabcolsep}{-4pt}
    \begin{tabular}{m{2.5cm}<{\centering} m{2.5cm}<{\centering} m{2.5cm}<{\centering} m{2.5cm}<{\centering} m{2.5cm}<{\centering} m{2.5cm}<{\centering}m{2.5cm}<{\centering}}
    
    Scene & NeRO & NMF & Gshader & GSIR &Ours & GT \\
    
    \includegraphics[height=1in]{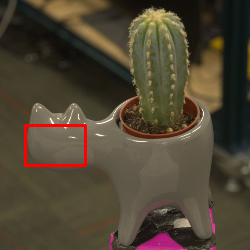}  &
    \includegraphics[height=1in]{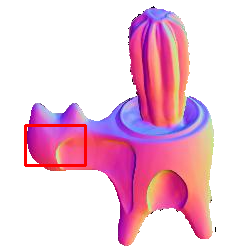} &
    \includegraphics[height=1in]{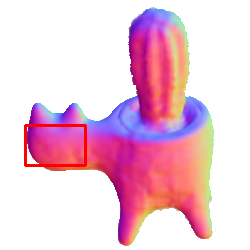} &
    \includegraphics[height=1in]{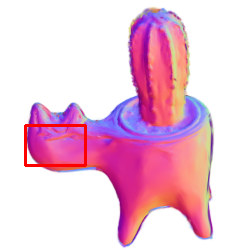} &
    \includegraphics[height=1in]{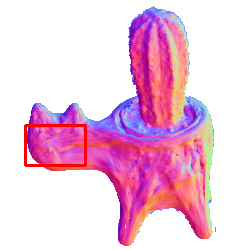} &
    \includegraphics[height=1in]{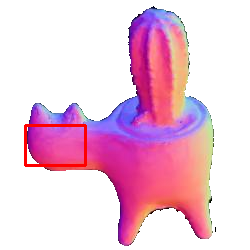}&
    \includegraphics[height=1in]{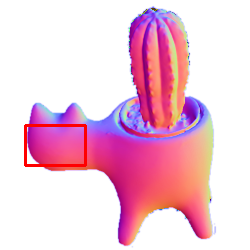}\\
    
    \includegraphics[height=1in]{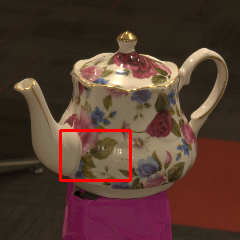} &
    \includegraphics[height=1in]{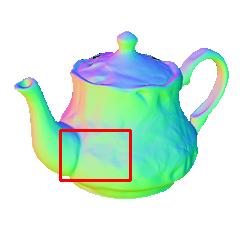} &
    \includegraphics[height=1in]{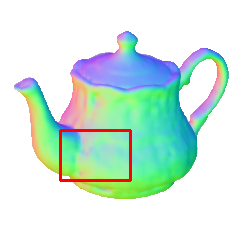} &
    \includegraphics[height=1in]{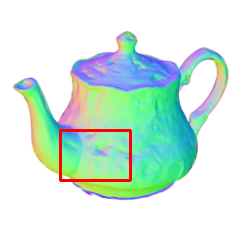} &
    \includegraphics[height=1in]{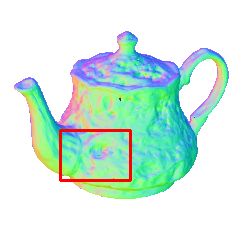} &
    \includegraphics[height=1in]{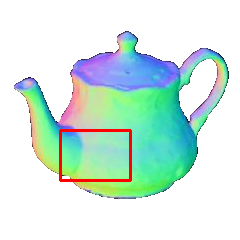} &
    \includegraphics[height=1in]{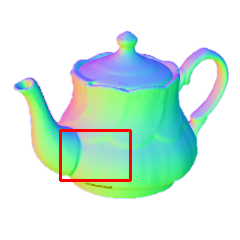}\\
    \end{tabular}
\end{center}
\caption{Comparison of normal reconstruction among our method, NMF~\cite{mai2023neural}, NeRO \cite{liu2023nero}, Gshader \cite{jiang2023gaussianshader} and GSIR \cite{liang2023gs} on the Stanford-ORB dataset \cite{kuang2024stanford}.}
\label{fig:normal_orb}
\end{figure*}

\begin{figure*}[htb]
\begin{center}
        \addtolength{\tabcolsep}{-4pt}
    \begin{tabular}{m{2cm}<{\centering} m{2.5cm}<{\centering} m{2.5cm}<{\centering} m{2.5cm}<{\centering} m{2.5cm}<{\centering} m{2.5cm}<{\centering} m{2.5cm}<{\centering}}
    
    New lighting & NeRO & NMF & Gshader & GSIR &Ours & GT \\
    
    \centering
    \includegraphics[width=0.6in]{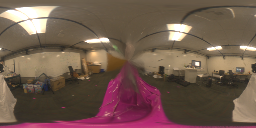} &
    \includegraphics[width=0.7in]{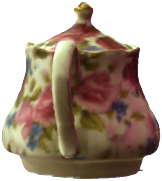} &
    \includegraphics[width=0.7in]{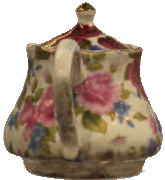} &
    \includegraphics[width=0.7in]{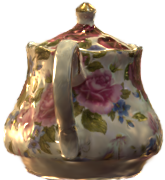} &
    \includegraphics[width=0.7in]{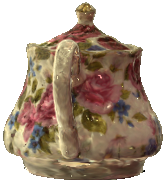} &
    \includegraphics[width=0.7in]{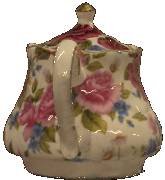} &
    \includegraphics[width=0.7in]{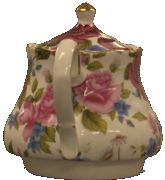} 
    \\
    
    \includegraphics[width=0.6in]{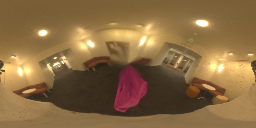} &
    \includegraphics[width=0.9in]{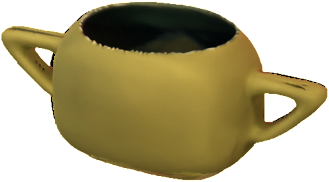} &
    \includegraphics[width=0.9in]{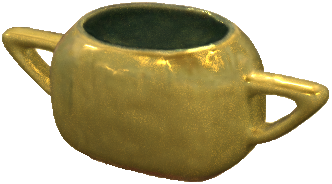} &
    \includegraphics[width=0.9in]{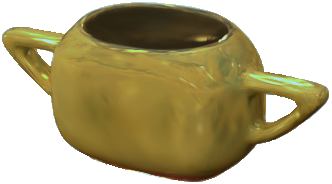} &
    \includegraphics[width=0.9in]{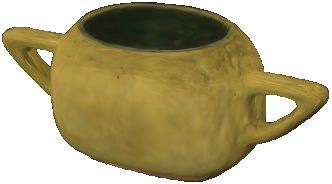} &
    \includegraphics[width=0.9in]{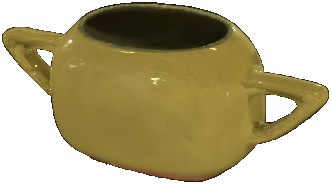} &
    \includegraphics[width=0.9in]{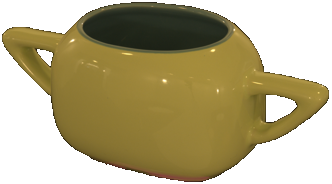}\\

    \end{tabular}
\end{center}
\caption{Comparison of relighting among our method, NMF~\cite{mai2023neural}, NeRO \cite{liu2023nero}, Gshader \cite{jiang2023gaussianshader} and GSIR \cite{liang2023gs} on the Stanford-ORB \cite{kuang2024stanford} dataset.}
\label{fig:relight_stanfordorb}
\end{figure*}

We also evaluate our method on four real glossy objects from the Stanford-ORB~\cite{kuang2024stanford} dataset. The quantitative measurements of normal reconstruction and relighting results are shown in Tab. \ref{tab:stanford}. Our method outperforms all the other methods. We also present the visual results in Fig.~\ref{fig:normal_orb} and Fig.~\ref{fig:relight_stanfordorb}. Again, we see that our method accurately reconstructs geometries and materials, displaying glossiness in real-scene relighting that matches the original objects. Conversely, prior methods struggle to model geometry in highly glossy regions, resulting in incorrect material properties and low-quality relighting effects.

\begin{figure*}[htb]
\begin{center}
    \addtolength{\tabcolsep}{-4pt}
    \begin{tabular}{m{2.5cm}<{\centering} m{2.5cm}<{\centering} m{2.5cm}<{\centering} m{2.5cm}<{\centering} m{2.5cm}<{\centering} m{2.5cm}<{\centering} m{2.5cm}<{\centering}}
    
    Scene & NeRO& NMF & Gshader & GSIR &Ours & Our relighting\\
    
        \includegraphics[width=1in]{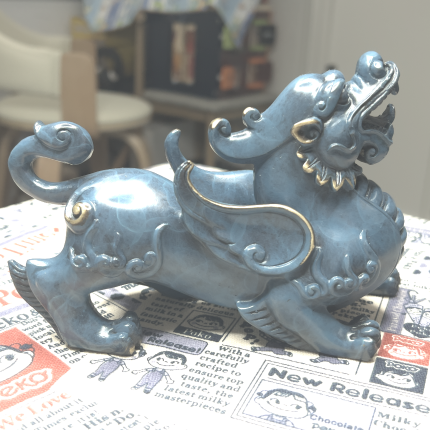} &
    \includegraphics[width=1in]{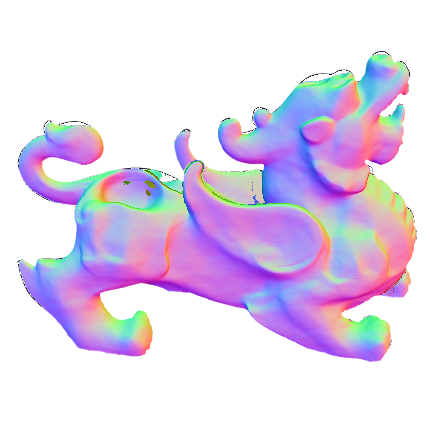} &
    \includegraphics[width=1in]{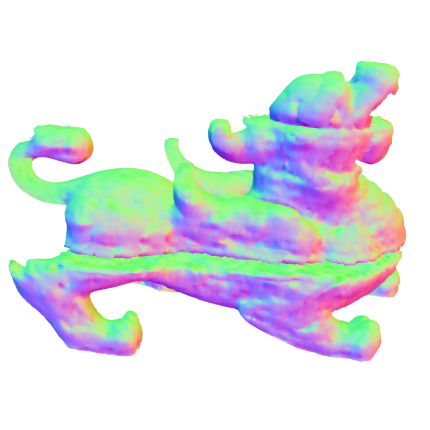} &
    \includegraphics[width=1in]{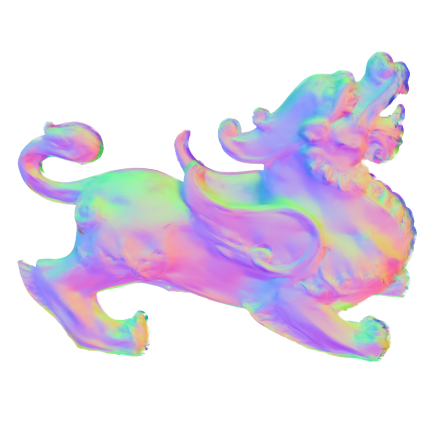}&
    \includegraphics[width=1in]{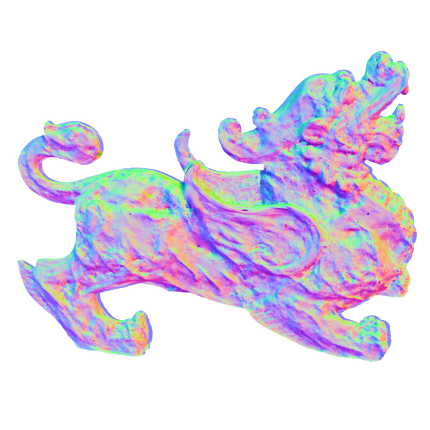} &
        \includegraphics[width=1in]{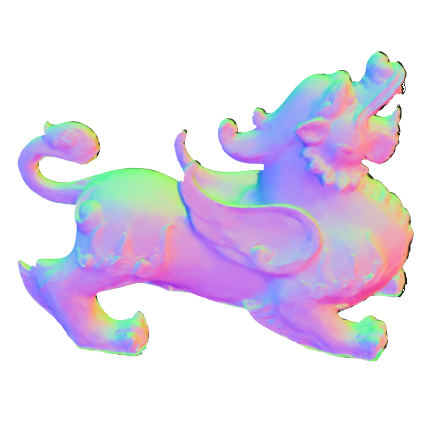} &
    \includegraphics[width=1in]{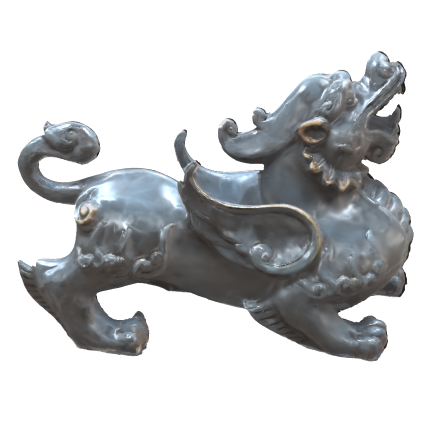} \\
    Qilin & & & & & & \\
    
        \includegraphics[width=1in]{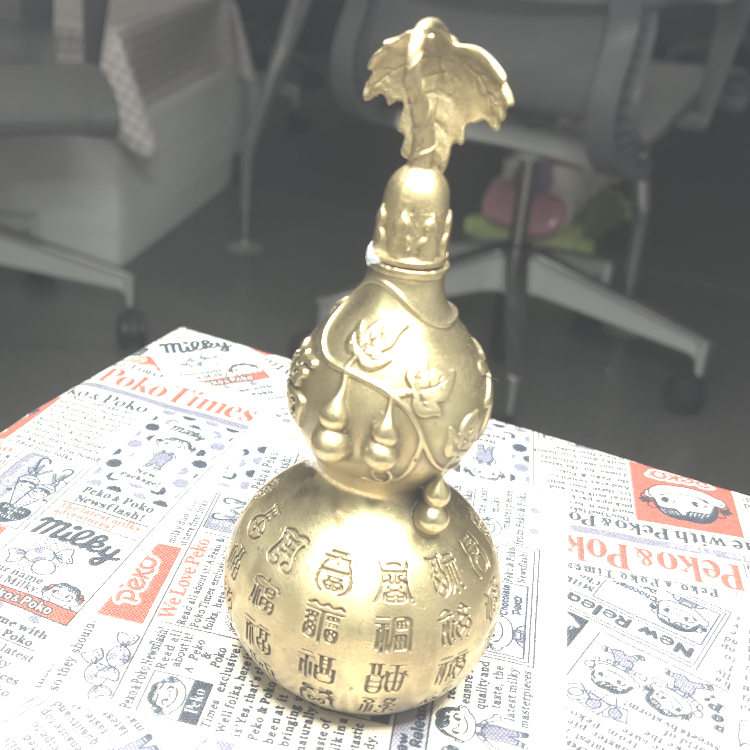} &
        \includegraphics[width=1in]{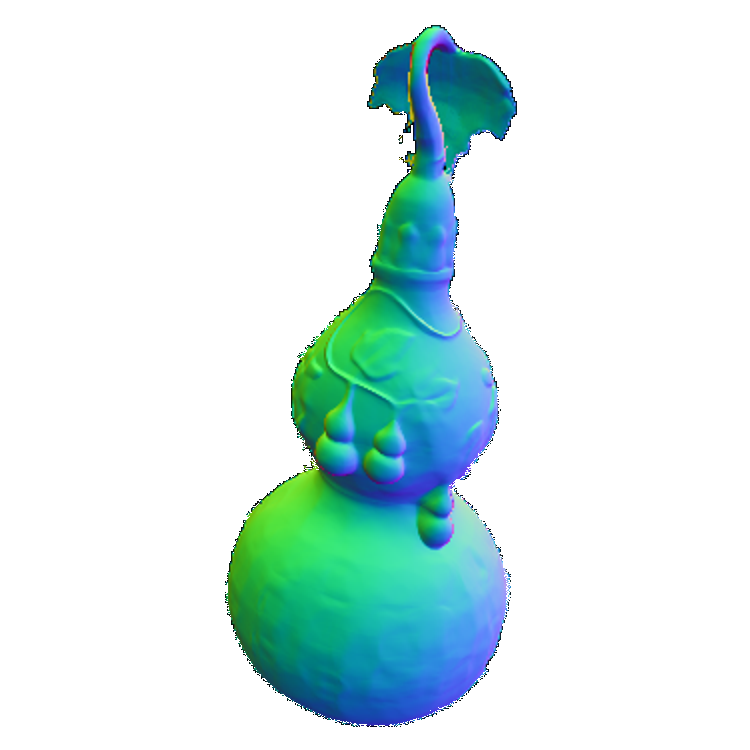} &
        \includegraphics[width=1in]{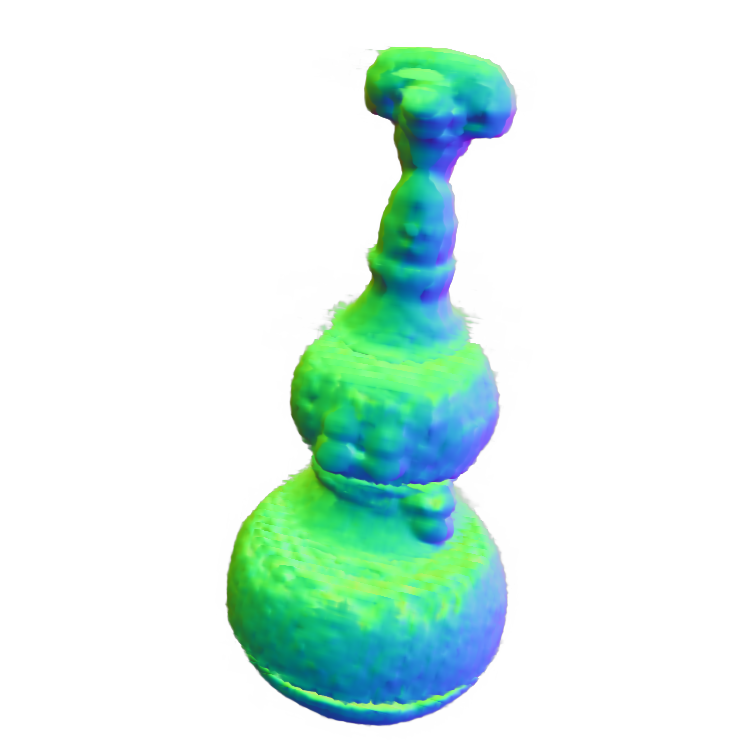} &
    \includegraphics[width=1in]{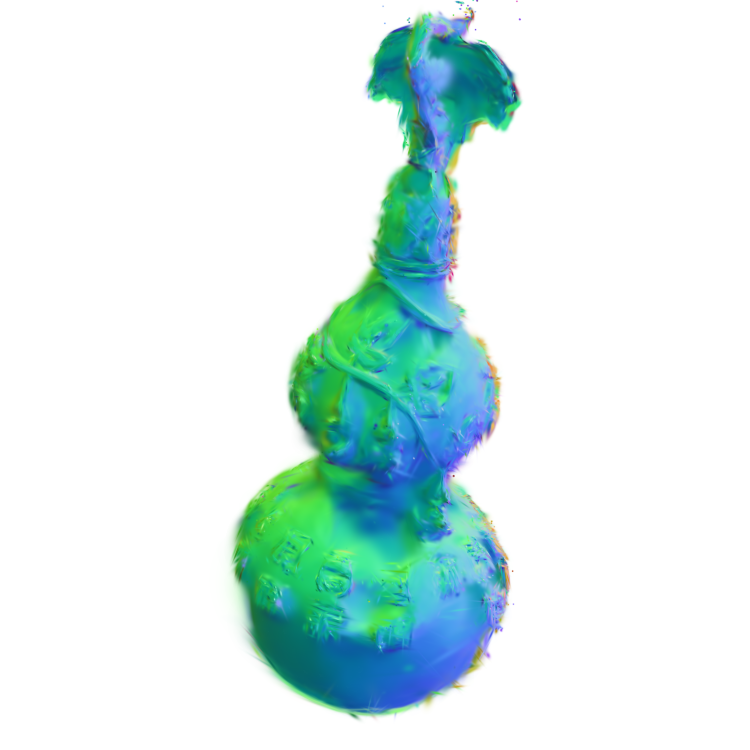} &
    \includegraphics[width=1in]{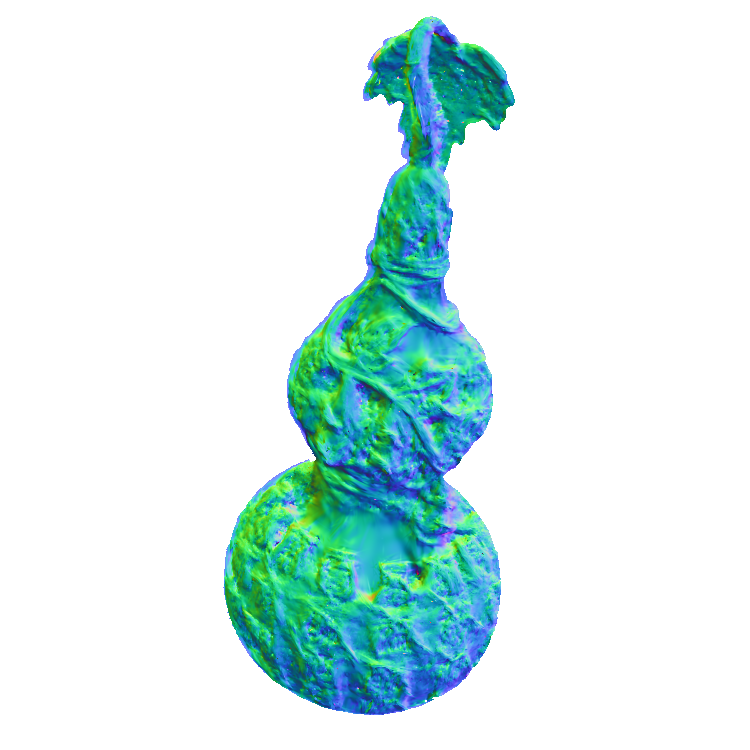} &
    \includegraphics[width=1in]{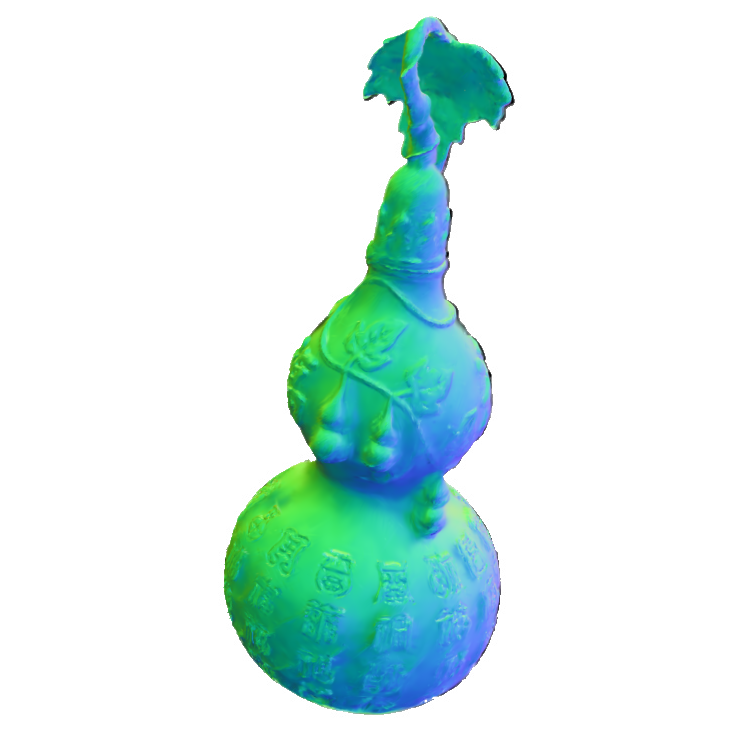}&
    \includegraphics[width=1in]{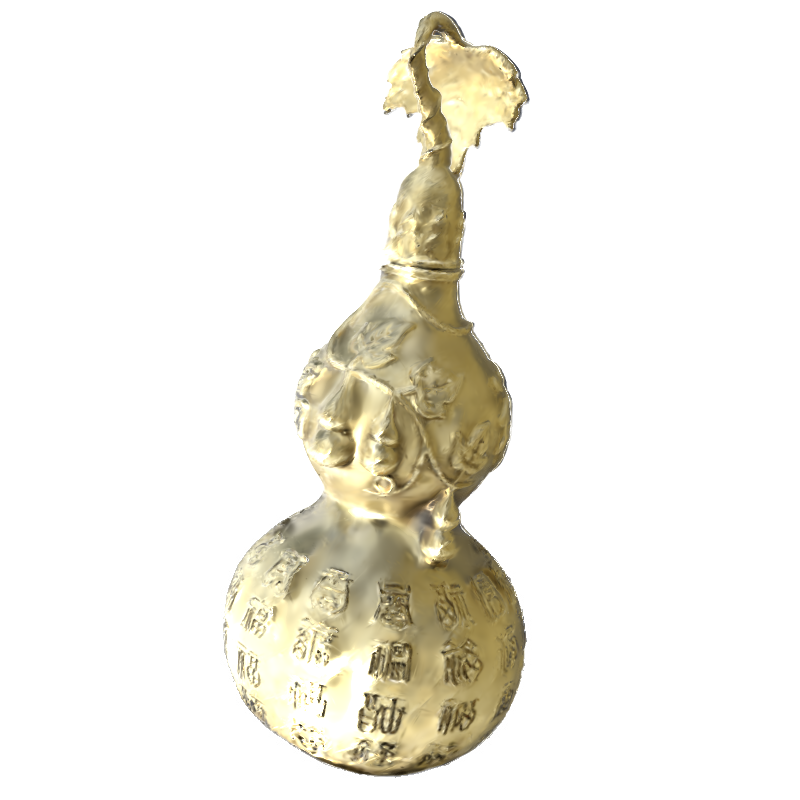} \\
    Brassgourd & & & & & & \\
    
        \includegraphics[width=1in]{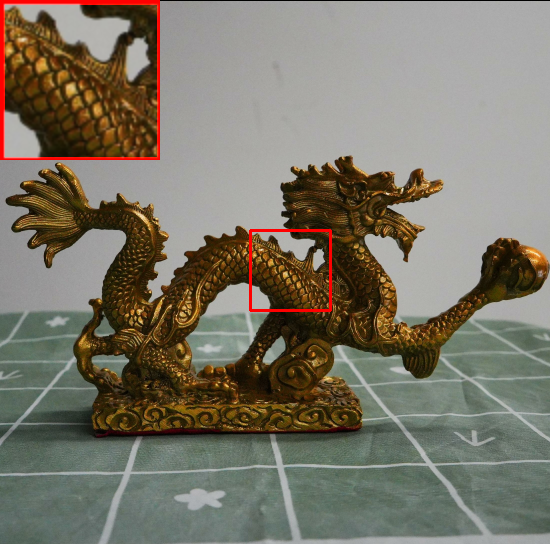} &
    \includegraphics[width=1in]{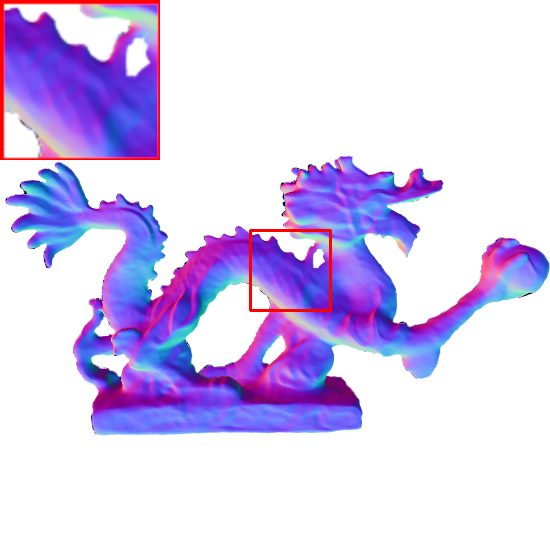} &
    \includegraphics[width=1in]{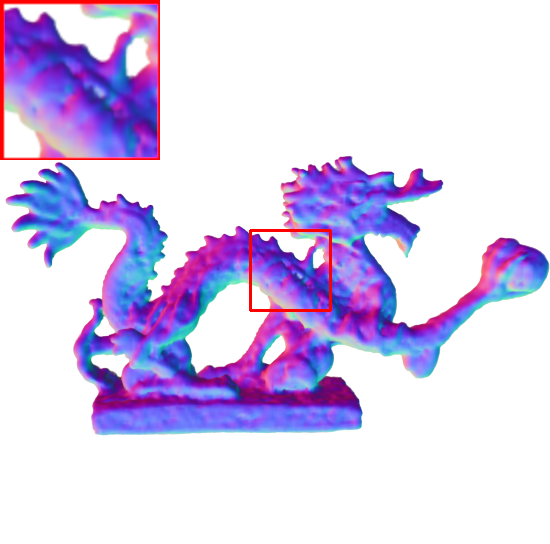} &
    \includegraphics[width=1in]{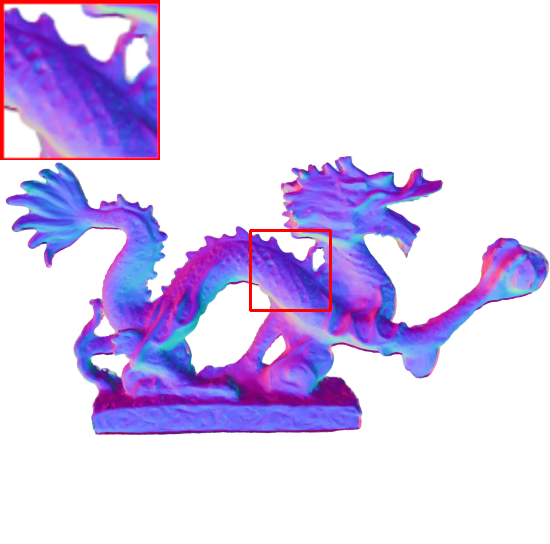}&
    \includegraphics[width=1in]{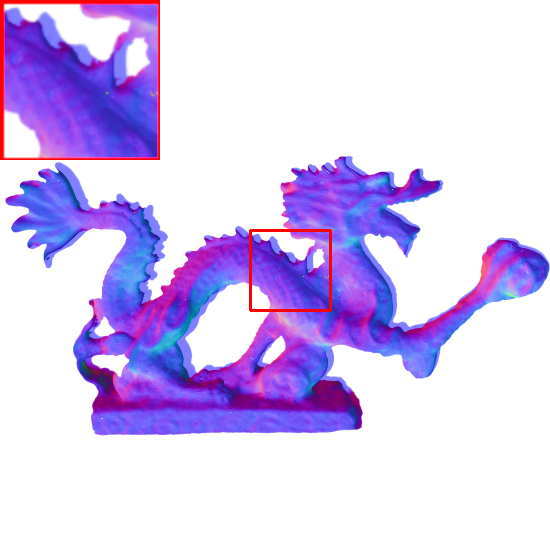} &
        \includegraphics[width=1in]{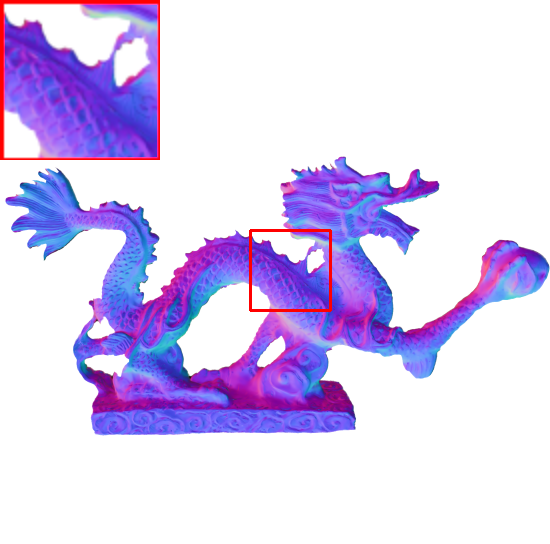} &
    \includegraphics[width=1in]{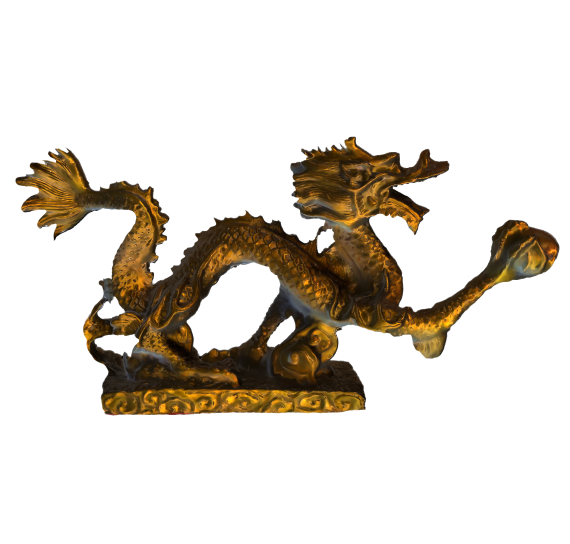} \\
    GoldenDragon & & & & & & \\
            
        \includegraphics[width=1in]{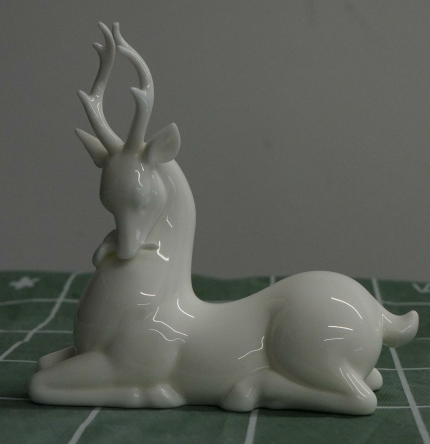} &
    \includegraphics[width=1in]{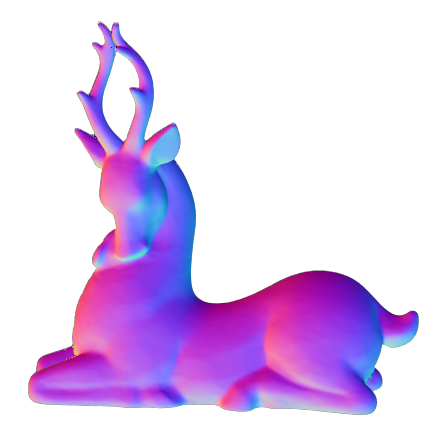} &
    \includegraphics[width=1in]{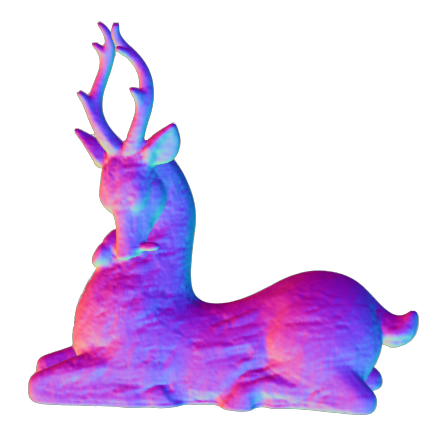} &
    \includegraphics[width=1in]{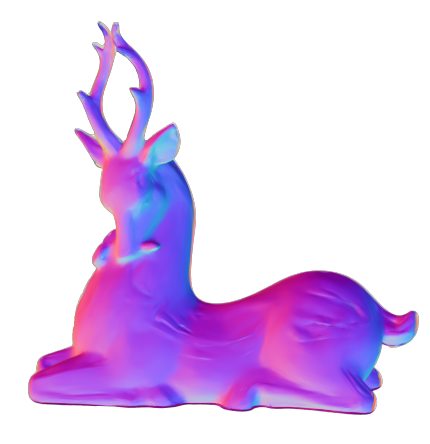}&
    \includegraphics[width=1in]{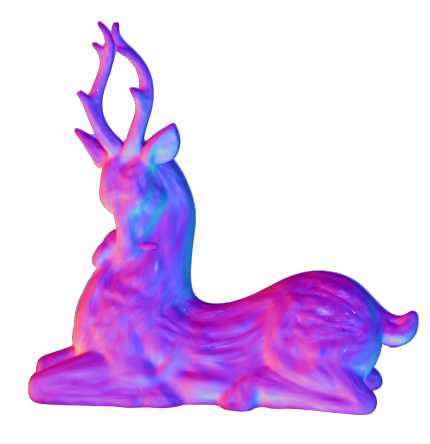} &
        \includegraphics[width=1in]{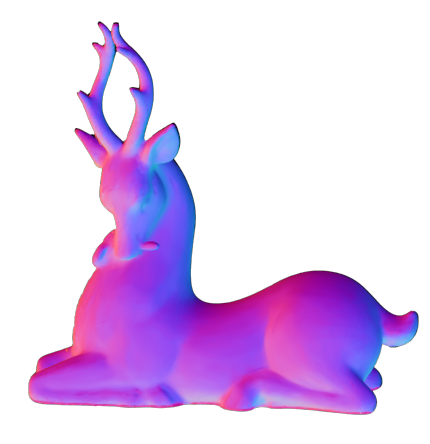} &
    \includegraphics[width=1in]{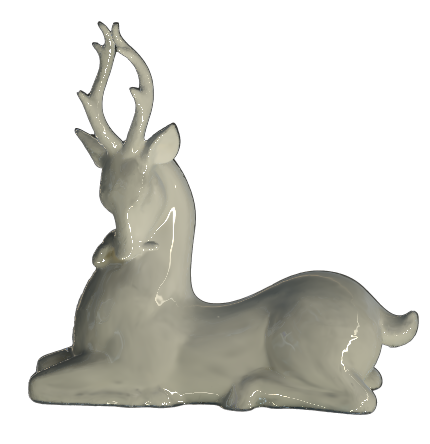} \\
    CeramicElk & & & & & & \\
    
    \end{tabular}
\end{center}
\caption{Comparison of normal reconstruction and relighting results among our method, NMF~\cite{mai2023neural}, NeRO \cite{liu2023nero}, Gshader \cite{jiang2023gaussianshader} and GSIR \cite{liang2023gs} on the NEILF-HDR \cite{zhang2023neilf++} dataset (the first and second rows) and real captured scenes (the third and forth rows).}
\label{fig:neilf}
\end{figure*}

\subsection{Ablation studies}
In this subsection, we evaluate several design choices in our pipeline by conducting ablation studies on them. 

    


\begin{figure}[htbp]
\begin{center}
    \addtolength{\tabcolsep}{-4pt}
    \begin{tabular}{ccccc}
    
    w/o prior & w/o prefiltering & full color & full normal & GT color  \\
    \includegraphics[height=0.6in]{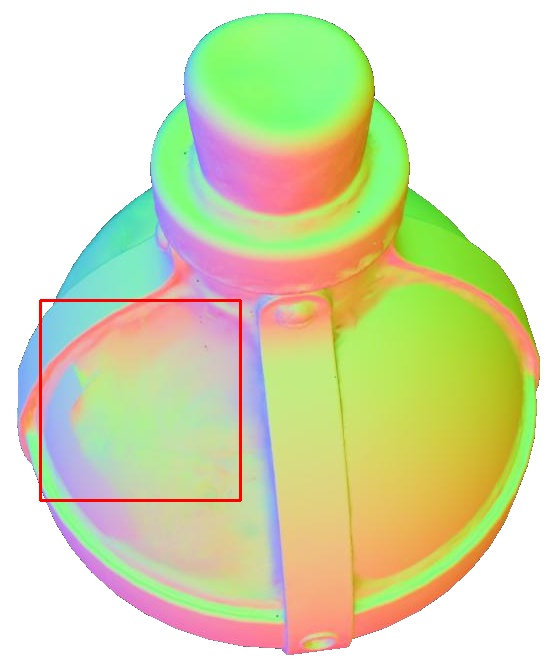} &
    \includegraphics[height=0.6in]{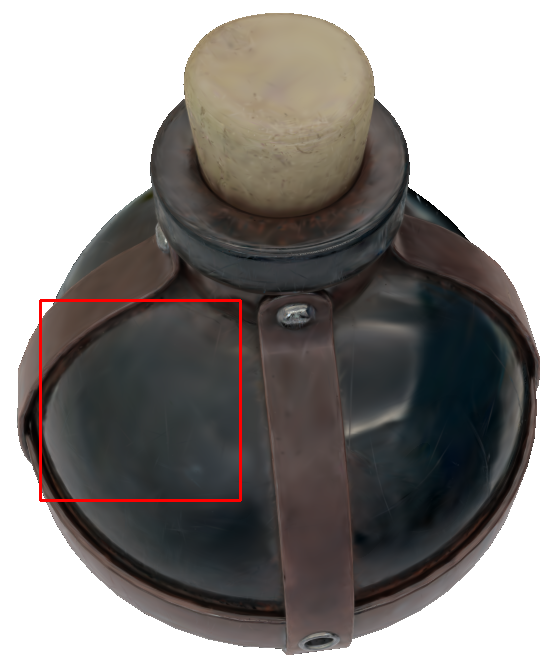} &
    \includegraphics[height=0.6in]{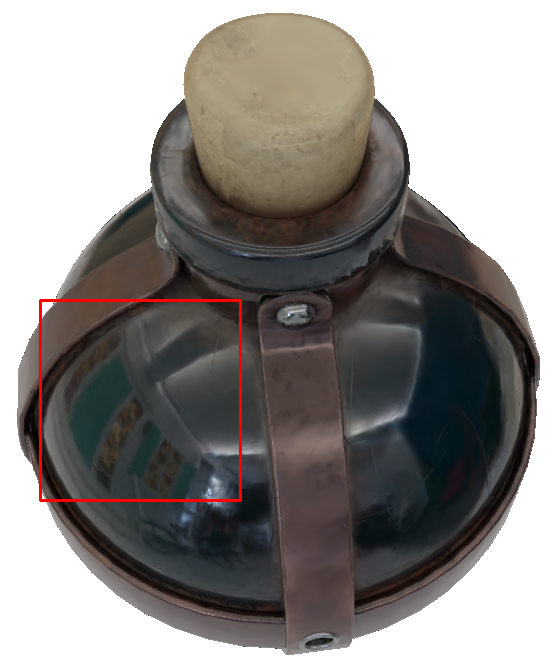} &
    \includegraphics[height=0.6in]{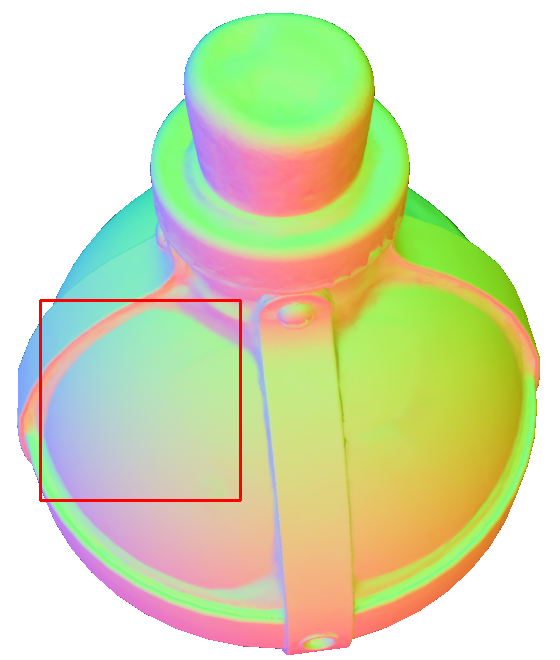} &
    \includegraphics[height=0.6in]{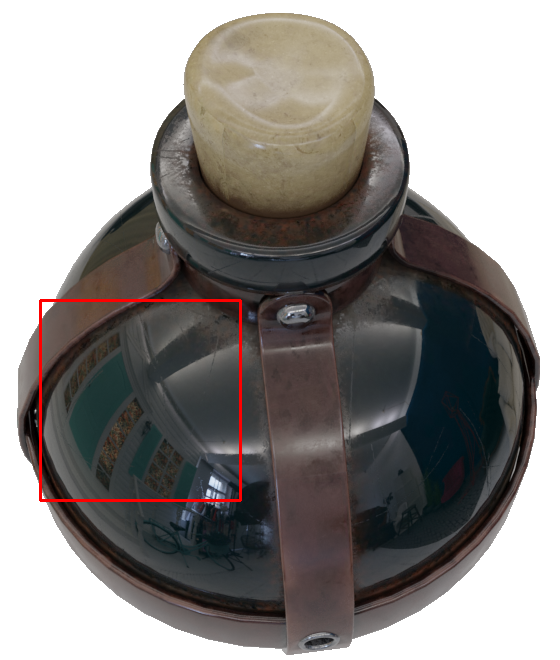}\\
    
    \includegraphics[height=0.6in]{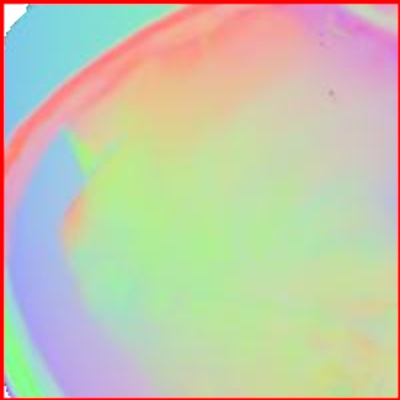} &
    \includegraphics[height=0.6in]{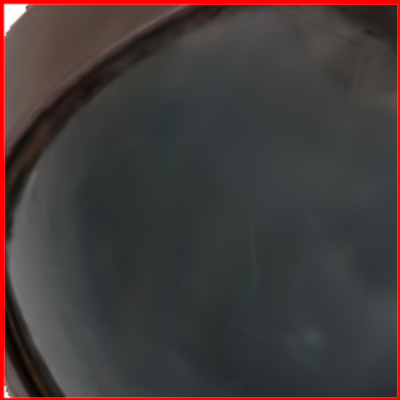} &
    \includegraphics[height=0.6in]{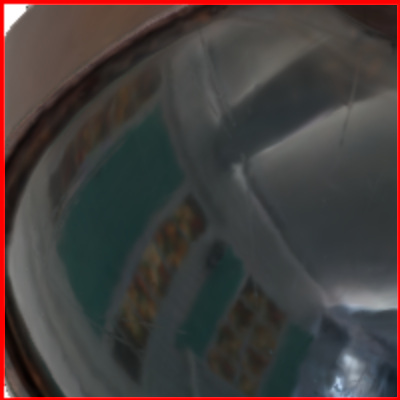} &
    \includegraphics[height=0.6in]{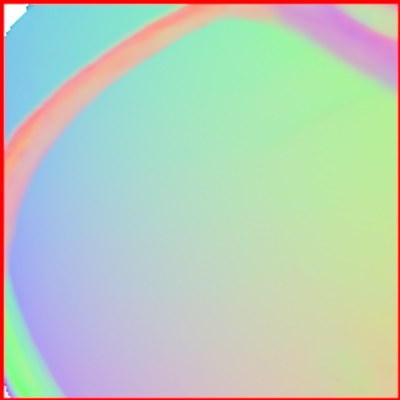} &
    \includegraphics[height=0.6in]{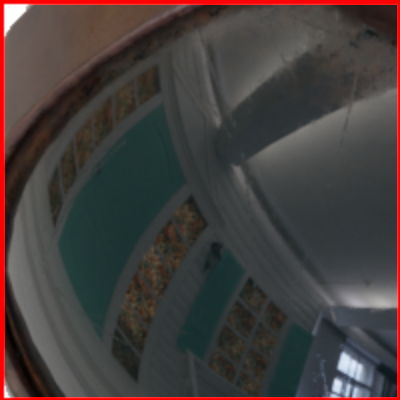}\\

    \includegraphics[height=0.6in]{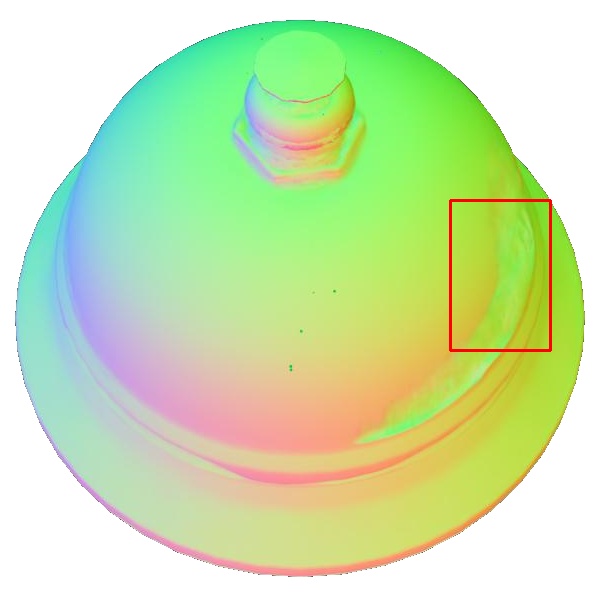} &
    \includegraphics[height=0.6in]{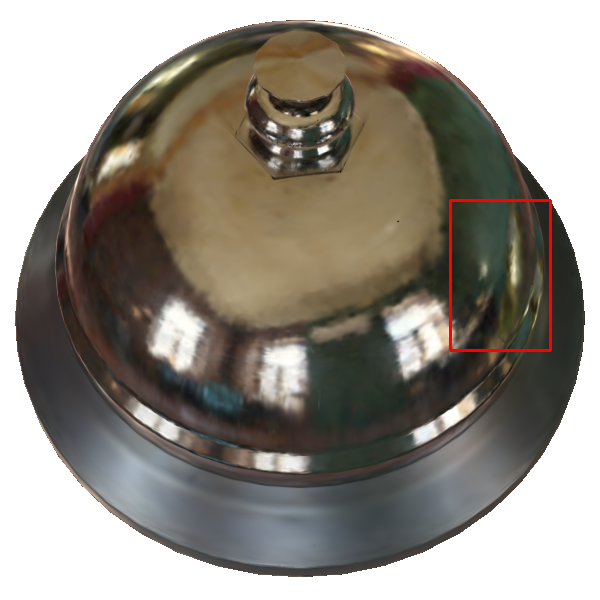} &
    \includegraphics[height=0.6in]{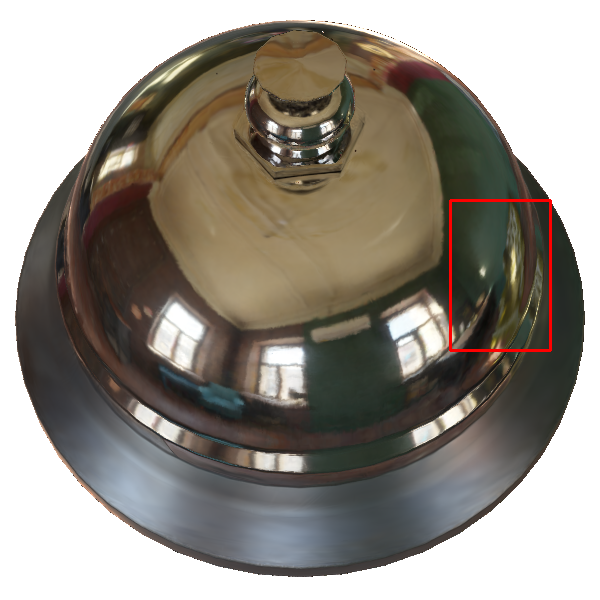} &
    \includegraphics[height=0.6in]{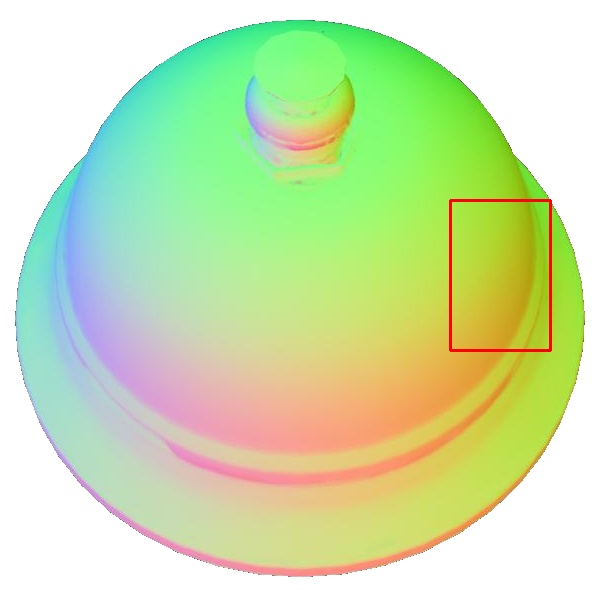} &
    \includegraphics[height=0.6in]{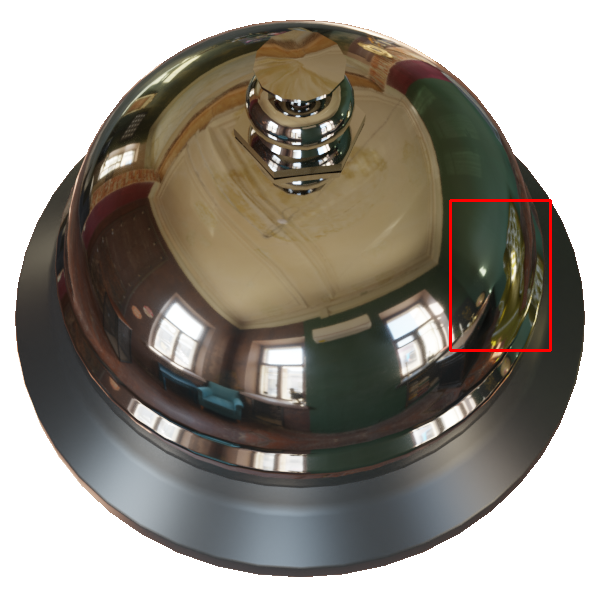}\\

    \includegraphics[height=0.6in]{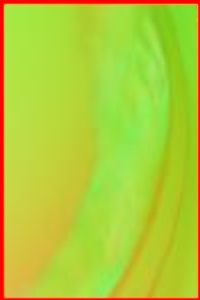} &
    \includegraphics[height=0.6in]{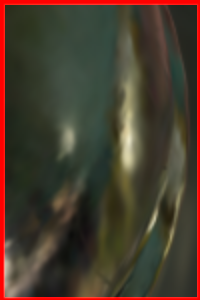} &
    \includegraphics[height=0.6in]{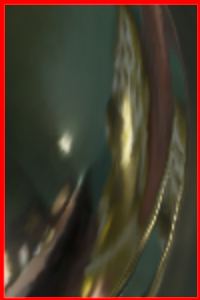} &
    \includegraphics[height=0.6in]{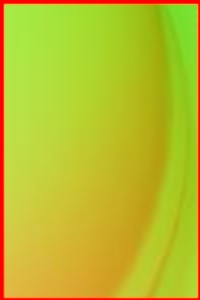} &
    \includegraphics[height=0.6in]{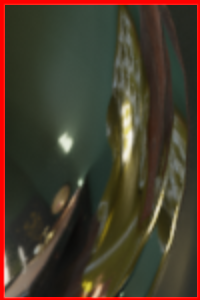}\\
    \end{tabular}
\end{center}
\caption{Validating the effect of the micro-facet geometry segmentation prior and the normal map prefiltering in training.}
\label{fig:ablation}
\end{figure}

\begin{table}[]
  \small
  \centering
\caption{\label{tab:ablation}Ablation studies on the relighting results using objects from the Glossy Synthetic dataset \cite{liu2023nero}. \textbf{Bold} means the best performance.}
\begin{tabular}{c|ccc}
\hline
& PSNR$\uparrow$  & SSIM$\uparrow$  & LPIPS$\downarrow$ \\
\hline
w/o prior                                                          & 23.51 & 0.898 & 0.098 \\
w/o normal map prefiltering & 24.61 & 0.922 & 0.072 \\
w/o hybrid & 23.52 & 0.902 & 0.082 \\
Ours         		                                                      & \textbf{25.47} & \textbf{0.928} & \textbf{0.068}\\
\hline
\end{tabular}
\end{table}

\begin{figure}[htbp]
\begin{center}
    \addtolength{\tabcolsep}{-4pt}
    \begin{tabular}{ccc}
    
   w/o hybird & w/ hybrid & GT  \\
    \includegraphics[width=1.1in]{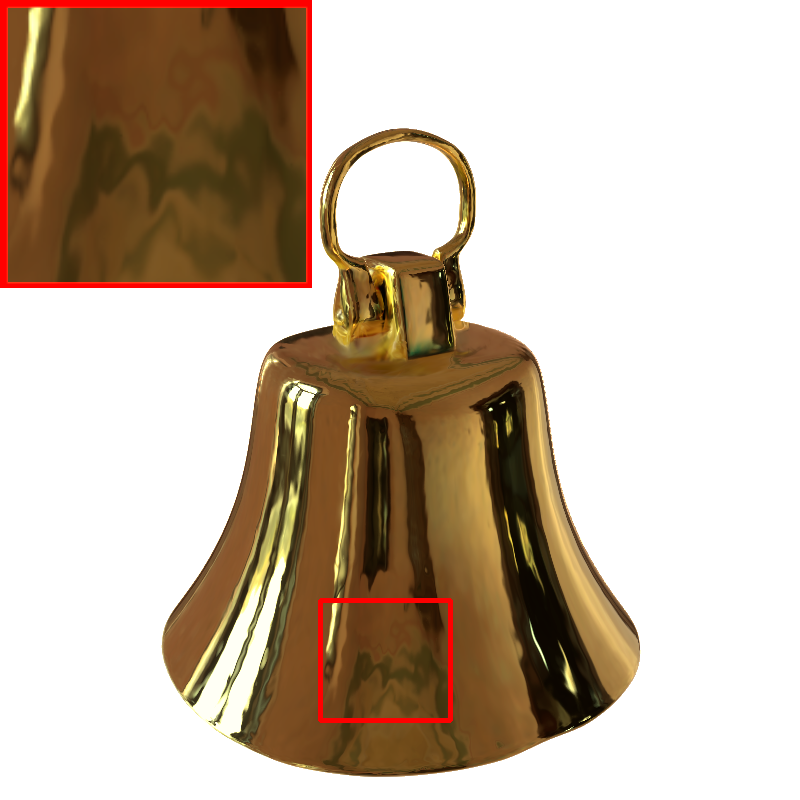} &
    \includegraphics[width=1.1in]{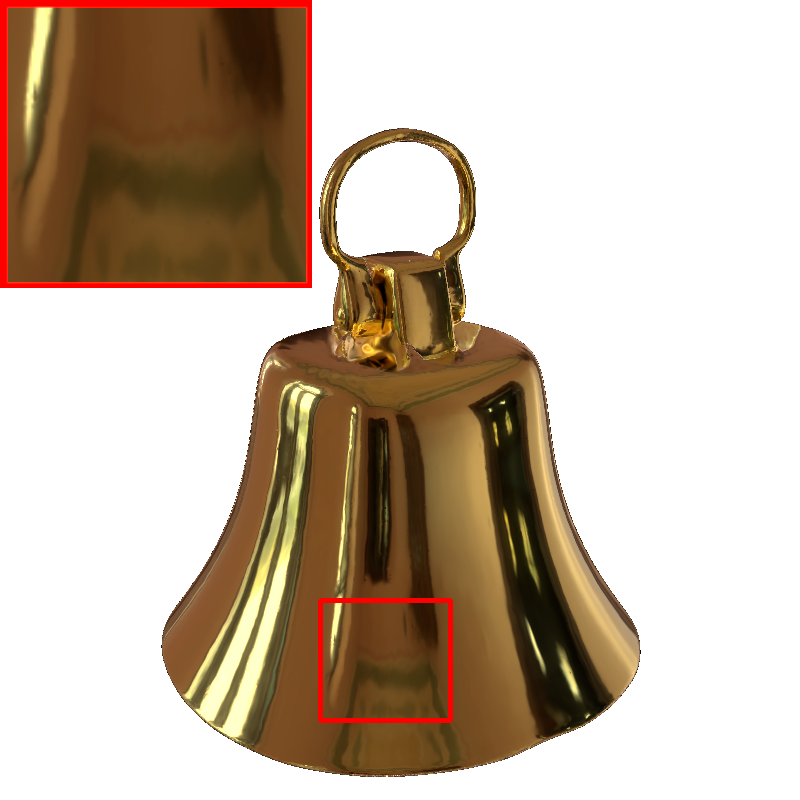} &
    \includegraphics[width=1.1in]{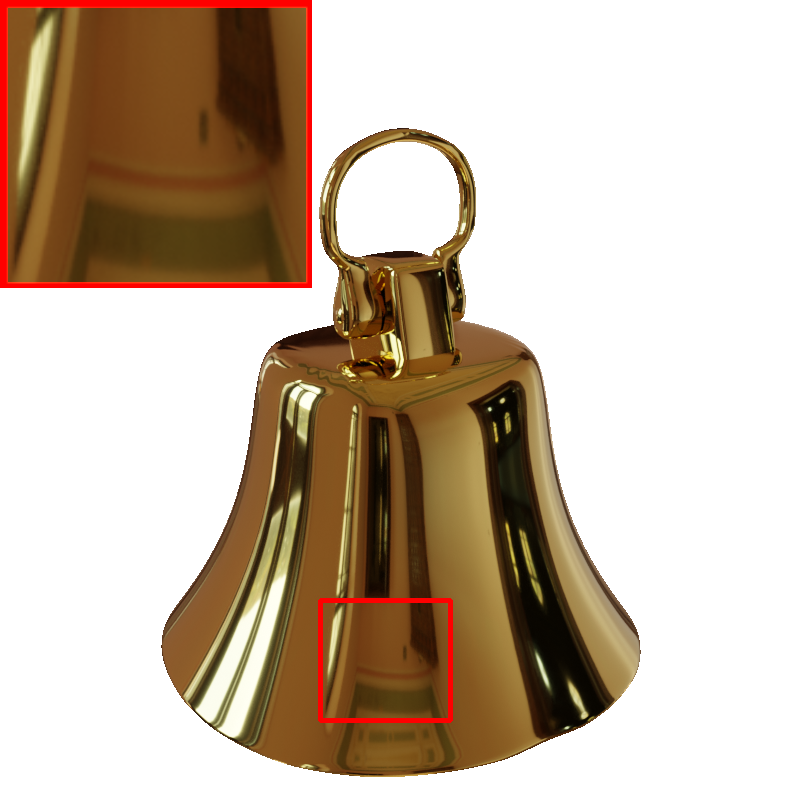}\\
    \end{tabular}
\end{center}
\caption{Validating the effect of the hybrid geometry and material representation.}
\label{fig:ablation_hybrid}
\end{figure}

\paragraph{Micro-facet Geometry Segmentation Prior} We introduce the micro-facet geometry segmentation prior to enhance the decomposition of geometry and material properties. To assess the effectiveness of the prior, we conduct a comparison with a baseline model trained without the prior. As depicted in Fig. \ref{fig:ablation}, the baseline model incorrectly separates micro-scale roughness and macro-scale normal, resulting in inaccurate geometry. Utilizing prior knowledge, our method successfully generates the correct normals. The data presented in Tab. \ref{tab:ablation} further confirm that the integration of the prior aids the model in generating reliable materials.

\paragraph{Normal Map Prefiltering} To assess the effectiveness of the strategy, we conduct an experiment by implementing the training without normal prefiltering. The comparative analysis between the relighting results is displayed in Fig. \ref{fig:ablation}. In the absence of normal prefiltering, the relighting results tend to produce blurred visual appearance. However, our method is capable of clearly mapping environmental lighting, achieving results that are close to the ground truth.

\paragraph{Hybrid Geometry and Material Representation} We utilize a hybrid explicit-implicit geometry representation to generate neural Gaussians for representing glossy objects. To validate the effectiveness of this strategy for inverse rendering, we conducted an experiment using the explicit Gaussian design in the same manner as Gshader \cite{jiang2023gaussianshader}. As shown in Fig. \ref{fig:ablation_hybrid}, without hybrid geometry and material representation, the explicit Gaussian fails to produce realistic relighting results. Quantitative metrics in Tab. \ref{tab:ablation} further validate this.

\section{Conclusion}
We have presented a new framework aimed at the efficient reconstruction of glossy objects with 3D-GS.
The key challenge of reconstructing glossy objects is the inherent ambiguities associated with inverse rendering. To tackle this problem, we propose the micro-facet geometry segmentation prior to significantly reduces the ambiguities and enhances the decomposition of both geometry and material properties. Additionally, we introduce a normal map prefiltering strategy that enhances the GlossyGS's ability to more accurately simulate the normal distribution of glossy surfaces. Finally, we introduce a hybrid geometry and material representation tailored for glossy objects. Extensive experiments on both synthetic and real datasets demonstrate consistently superior performance of the proposed method.

\begin{figure}
\begin{center}
\addtolength{\tabcolsep}{-4pt}
	\begin{tabular}{cc|cc}
        \includegraphics[height=0.8in]{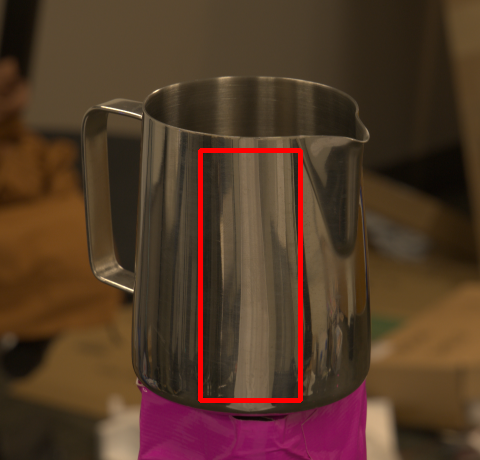} &
        \includegraphics[height=0.8in] {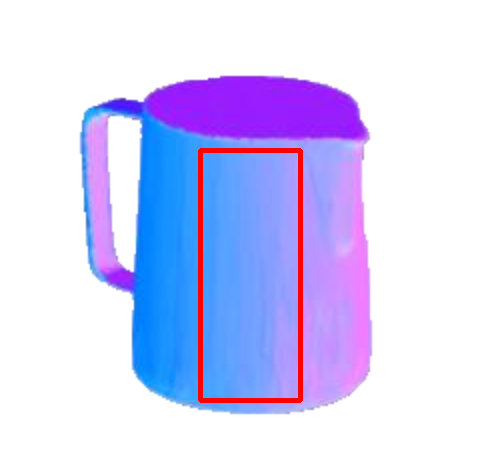}&
        \includegraphics[height=0.8in]{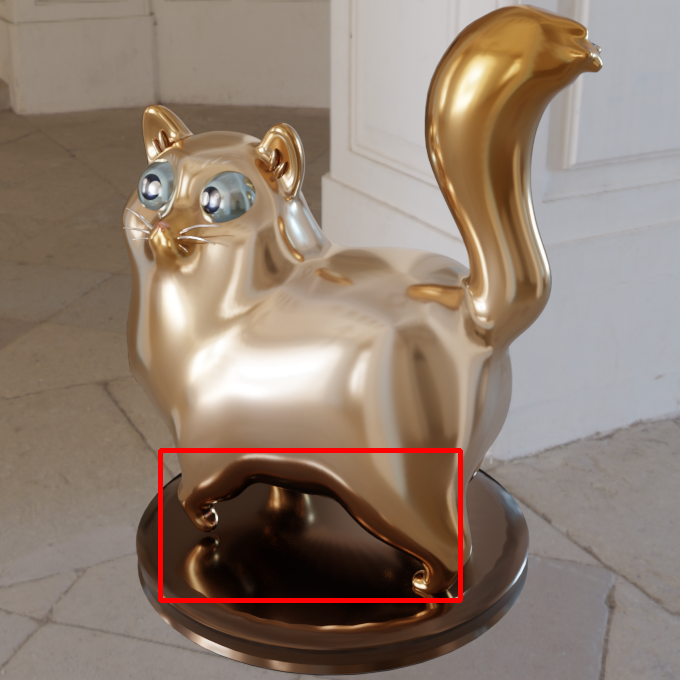} &
        \includegraphics[height=0.8in]{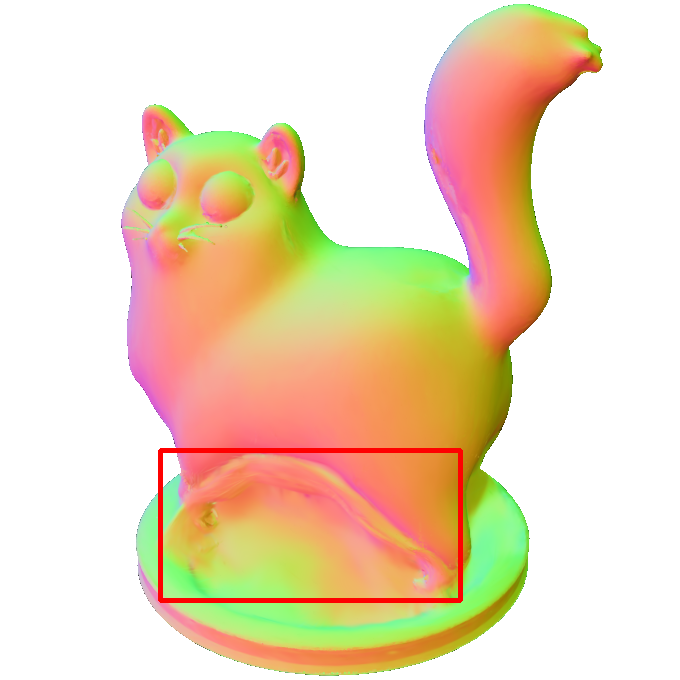} 
		\\
         \footnotesize Image  & {\footnotesize Our normal} & \footnotesize Image  & {\footnotesize Our normal}  \\
	\end{tabular}
\end{center}
  \caption{Limitations. Our method struggles to remove interreflections, resulting in incorrect geometries.}
  \label{fig:limitation}
\end{figure}


\textbf{Limitations and future work.}
Although GlossyGS can accurately reconstruct the geometries and materials of glossy objects without sacrificing performance, it still has some limitations. One problem is that our method assumes distant illumination, which means near-field lighting effects are not considered. In reality, glossy objects often capture reflections from the photographer, which change with camera angles, as shown in Fig. \ref{fig:limitation}. Additionally, our method does not effectively handle interreflections due to our rendering scheme. In such scenarios, while our carefully designed prior and prefiltering strategy can minimize the impact of these reflections on geometry, it still struggles to produce accurate geometries. One potential solution is to model dynamic illumination similarly to NeRO~\cite{liu2023nero}. Furthermore, our method struggles with concave surfaces due to viewpoint limitations, a common challenge in geometry reconstruction. We plan to address this issue in future work.

\bibliographystyle{IEEEtran}
\bibliography{svbrdf}

\end{document}